\newcommand{\myAbstractText}{
    The geometric design of fully-actuated and omnidirectional $N$-rotor aerial vehicles is conventionally formulated as a parametric optimization problem, seeking a single optimal set of $N$ orientations within a fixed architectural family. This work departs from that paradigm to investigate the intrinsic topological structure of the optimization landscape itself. We formulate the design problem on the product manifold of Projective Lines $(\RPtwo)^N$, fixing the rotor positions to the vertices of polyhedral chassis while varying their lines of action. By minimizing a scaling/rotation-invariant Log-Volume isotropy metric, we reveal that the topology of the global optima is governed strictly by the symmetry of the chassis.
For generic (irregular) vertex arrangements, the solutions appear as a discrete set of isolated points. However, as the chassis geometry approaches regularity, the solution space undergoes a critical phase transition, collapsing onto an $N$-dimensional Torus of the lines tangent at the vertices to the circumscribing sphere, and subsequently reducing to continuous 1-dimensional curves driven by ``Affine Phase Locking.''
\secondrev{We synthesize these observations into the "N-5 Scaling Law": an empirical relationship demonstrating that for low complexity chassis ($N \le 7$), the space of optimal configurations collapses into exactly $K=N-5$ strictly disconnected 1D topological closed branches, whereas for higher-dimensional architectures ($N \ge 8$), these 1D loops serve as the fundamental topological backbones embedded within broader multidimensional optimal hypersurfaces.}
For polygonal chassis, we empirically discover that these locking patterns correspond to the sequences of all Star Polygons $\{N/q\}$ with $2<q<N-2$, allowing exact prediction for arbitrary $N$.
Crucially, this topology reveals a \emph{design redundancy} that enables \textit{optimality-preserving morphing}: the vehicle can continuously reconfigure along these branches while preserving optimal isotropic control authority.
}

\renewcommand{\myAbstractText}{
\secondrev{We investigate the topological structure of the optimal actuation landscape for fully-actuated $N$-rotor aerial vehicles. By formulating the design problem on the $2N$-dimensional product manifold of projective lines $(\RPtwo)^N$ and minimizing a rotation-invariant Log-Volume isotropy metric, we map how optimal rotor orientations evolve across diverse polyhedral chassis. 
The results establish that global optimality is strictly bounded by geometric symmetry. While irregular chassis yield discrete, isolated optimal configurations, regular geometries induce a structural phase transition: the optimal space initially collapses onto an $N$-dimensional tangent torus, then systematically reduces to continuous configurations governed by affine phase coordination. 
These collapses define the ``$N-5$ Scaling Law.'' For $N \le 7$, the optimal landscape fundamentally forms exactly $K=N-5$ disconnected 1D closed loops. For $N \ge 8$, these 1D trajectories expand into core backbones embedded within multi-dimensional flat optimal hypersurfaces. Furthermore, for regular planar geometries, we theoretically unify these trajectories by demonstrating a strict geometric isomorphism to star polygons $\{N/q\}$ ($2<q<N-2$), yielding exact analytical predictions of the $N-5$-branched optimal manifolds for arbitrary $N$. 
By shifting the classical framework from single and static optimal points to continuous optimal manifolds, this work establishes the mathematical foundations for optimality-preserving morphing. The identified topological branches provide continuous null spaces, introducing to the field the concept of \emph{design redundancy}, and allowing symmetric multi-rotor systems to dynamically reconfigure their thrust vectors without degrading their perfectly isotropic control authority.}
}

\documentclass[Afour,sageh,times]{sagej}

\usepackage{moreverb,url}

\usepackage[colorlinks,bookmarksopen,bookmarksnumbered,citecolor=red,urlcolor=red]{hyperref}

\usepackage[utf8]{inputenc}
\usepackage[english]{babel}
\usepackage[autostyle]{csquotes}
\MakeOuterQuote{"}

\usepackage{amsmath, amssymb, amsfonts, amsthm}
\usepackage{graphicx}
\usepackage{booktabs}
\usepackage{algorithm}
\usepackage{algpseudocode}
\usepackage{bm}

\usepackage{subcaption} 

\usepackage[table]{xcolor} 
\usepackage{multirow}
\usepackage{tabularx}
\usepackage{array}     
\usepackage{makecell}  
\usepackage[export]{adjustbox} 

\usepackage[title]{appendix}

\renewcommand{\tabularxcolumn}[1]{m{#1}}

\newcommand{\Graff}{\mathcal{G}(1,3)}   
\newcommand{\RPtwo}{\mathbb{R}P^2}        
\newcommand{\RPone}{\mathbb{R}P^1}        
\newcommand{\SphereTwo}{S^2}        

\newcommand{\Torus}{\mathbb{T}}        
\newcommand{\tangentTorus}{{^\mathcal{C}\Torus}^N_\perp}        
\newcommand{\Manifold}{\mathcal{M}}    
\newcommand{\Grasp}{\mathbf{A}}        
\newcommand{\PosSet}{\mathcal{P}}      

\newcommand{\dvec}{\mathbf{d}}

\newcommand{\pvec}{\mathbf{p}}

\newif\ifblind
\blindfalse  

\newtheorem{empirical}{Empirical Law}
\newtheorem{definition}{Definition}
\newtheorem{remark}{Remark}

\usepackage{hyperref}
\hypersetup{
    colorlinks=true,  
    linkcolor=black,  
    citecolor=black,  
    filecolor=black,  
    urlcolor=black    
}

\newcommand{\firstrev}[1]{#1}                  

\newcommand{\secondrev}[1]{#1}      

\begin{document}


\runninghead{Franchi} 
\title{The N$-$5 Scaling Law: Topological Dimensionality Reduction in the Optimal Design of Fully-actuated Multirotors}
\author{Antonio Franchi\affilnum{1,2}}
\affiliation{\affilnum{1}Robotics and Mechatronics Department, Electrical Engineering,  Mathematics, and Computer Science (EEMCS) Faculty, University of Twente, 7500 AE Enschede, The Netherlands.\\
\affilnum{2}Department of Computer, Control and Management Engineering, Sapienza University of Rome, 00185 Rome, Italy.}
\corrauth{Antonio Franchi, University of Twente, The Netherlands and Sapienza University of Rome, Italy.}
\email{schol@r-franchi.eu}

\begin{abstract}
    \myAbstractText  
\end{abstract}

\maketitle


\section{Introduction}

In recent years, the domain of aerial robotics has expanded beyond visual inspection to encompass active physical interaction with the environment. Multi-Rotor Aerial Vehicles (MRAVs) have emerged as the dominant platform for these applications, offering rapid deployment to high-altitude and unstructured workspaces, thereby significantly reducing risk and cost in industrial maintenance scenarios,~\cite{AerialRobotic}. However, the transition from free-flight observation to contact-based manipulation imposes stringent requirements on stability and the capability for precise wrench generation, ~\cite{Franchi2019Interaction, PhyintReview}.

The ubiquitous quadrotor architecture features collinear propellers with parallel thrust vectors. While this design offers mechanical simplicity, it is inherently underactuated. The platform possesses only four control inputs for six Degrees of Freedom (DoF), creating an inescapable coupling between translational and rotational dynamics. To exert a lateral force, a quadrotor must tilt its body, thereby changing its orientation. This dynamic coupling fundamentally limits dexterity in interaction tasks, despite the development of robust control strategies like IDA-PBC designed to stabilize such underactuated contact, see, e.g.,~\cite{Yuksel2019Aerial}.

\begin{table*}[th!]
\caption{\textbf{Research Dossier: Executive Summary of Context, Methods, and Findings.}}
\label{tab:research_dossier}
\centering
\small
\renewcommand{\arraystretch}{1.25}
\begin{tabularx}{\textwidth}{|l|X|}
\hline
\rowcolor[gray]{0.9}
\textbf{Attribute} & \textbf{Description \& Specification} \\ \hline

\multicolumn{2}{|c|}{\textbf{I. CONTEXT \& SCOPE}} \\ \hline
\textbf{Problem Domain} & \textbf{Fully Actuated Multirotors (FAM).} \firstrev{Fully-actuated and redundant} platforms with $N \ge 6$ fixed-pitch and \firstrev{either} bidirectional \firstrev{or unidirectional} rotors. Control inputs are rotor speeds only. \\ \hline
\textbf{Chassis Model} & A rigid spatial arrangement of $N$ vertices (chassis geometry $\mathcal{C}$), where rotor centers coincide with vertices. \\ \hline
\textbf{Core Question} & Given a chassis $\mathcal{C}$, what is the set of all rotor orientations $\mathcal{L}(\mathcal{C})$ (the ``solution landscape'') that maximizes the \textbf{isotropy} of force and moment generation (i.e., the control authority)? Is $\mathcal{L}(\mathcal{C})$ a discrete set, a continuous manifold, or scattered points with little structure, \ldots? \\ \hline

\multicolumn{2}{|c|}{\textbf{II. FORMULATION}} \\ \hline
\textbf{Search Space} & The product of $N$ projective planes: $\mathcal{M} = (\RPtwo)^N$. Dimensionality: $2N$. \\ \hline
\textbf{Objective Function} & \textbf{Log-Volume of the Wrench Polytope, Eq.~\eqref{eq:cost_logvolume}.}
\emph{Why?} Superior to Condition Number or Minimum Singular Value because it acts as a smooth barrier function, is differentiable, and strongly penalizes rank deficiency while rewarding overall control authority. \\ \hline
\textbf{Assumptions} & 1. \textbf{Rigid Body Dynamics and Geometric Idealization:} Standard Euler-Lagrange formalism and point-source thrust at vertices. (Established  approach in robotic multirotor literature, proven to work very well in practice). \\
& 2. \textbf{Simplified Aerodynamics:} Thrust $\propto \omega^2$, Moment $\propto$ Thrust. Drag moments and dynamic effects (e.g., blade flapping) are omitted. \\
& \emph{Justification:} This study optimizes the intrinsic geometric wrench capacity. Unmodeled aerodynamic forces do not alter the optimality of the static grasp map and are treated as extrinsic disturbances to be rejected by the controller. This is a standard approach in the literature, proven to work excellently well in practice. \\
& 3. \textbf{Bidirectional Thrust:} ESCs allow full force polytope coverage. (Standard hardware nowadays). \secondrev{However, the bidirectional formulation establishes a geometric upper bound that strongly informs the design of unidirectional thrust ESCs as well.}
\\\hline

\multicolumn{2}{|c|}{\textbf{III. KEY DISCOVERIES}} \\ \hline
\textbf{Design Redundancy} & \textbf{Manifold Collapse.} For symmetric chassis (Type IV.B/C in Table~\ref{tab:unified_classification}, e.g., Regular Polygons and Platonic Solids), the global optimum is not a unique point but a continuous \emph{1D manifold}. This proves the existence of a ``Design Nullspace,'' allowing structural reconfiguration while maintaining maximal control authority. \\ \hline
\textbf{The ``$N-5$'' Law} 
& 
\secondrev{\textbf{Topological Scaling.} For regular chassis, the solution manifold is geometrically quantized. For $N \le 7$, it consists of exactly \emph{$N-5$ strictly distinct}, disconnected 1D loops (\emph{isomers}) with different phase offsets, which then evolve into $N-5$ embedded core trajectories within higher-dimensional flat spaces for $N \ge 8$.}
For polygons, the phase offsets for each branch correspond to the vertex sequences of an $\{N/q\}$ \emph{star polygon} with $2 < q < N-2$. This discrete structure suggests a topological basis for fault-tolerant modes.\\ \hline
\textbf{Tangent Heuristic} & \textbf{Dimensionality Reduction.} For regular chassis, the optimal directions on the curve do not span the full space $\Manifold$. Global optima are asymptotically confined to the \emph{Tangent Torus} $\tangentTorus$, where thrust vectors are perpendicular to position vectors. This reduces the landscape dimensionality by half ($2N \to N$). \\ \hline

\multicolumn{2}{|c|}{\textbf{IV. METHODOLOGY}} \\ \hline
\textbf{Approach} & \textbf{Experimental Mathematics} \& 
Mathematical Phenomenology. \\
& \emph{Note on Rigor:} We employ abductive reasoning to transition from numerical data to general laws. We formulate a general \emph{geometric isomorphism} with star polygons $\{N/q\}$ that analytically predicts the optimal phase patterns for arbitrary $N$. \\ \hline
\textbf{Pipeline} & 1. \textbf{Global Optimization:} SQP with massive multi-start ($10^3$ or more seeds) to map the landscape $\mathcal{L}(\mathcal{C})$. \\
& 2. \textbf{Topological Fitting:} PCA and manifold learning to identify the algebraic structure of solution clusters. \\
& 3. \textbf{Theoretical Unification:} Derivation of the \emph{N-5 Empirical Law} and the  \emph{Star Polygon Isomorphism}, providing a closed-form generative sequence for the optimal phases. \\ \hline
\textbf{Scope Limitation} & This is a geometric study (Mathematical Phenomenology) of widely adopted robotic multirotor models, with strong foundational implications for future practical MRAV design. Performance benchmarking via hardware experiments or simulations is \textbf{explicitly out of scope}. \\ \hline
\multicolumn{2}{|c|}{\textbf{V. IMPACT \& IMPLICATIONS}} \\ \hline
\textbf{Target Audience} & Robotic MRAV designers and engineers (for optimal geometry synthesis), applied control roboticists (allocation matrix properties), roboticists interested in optimal kinematics, applied mathematicians. \\ \hline
\textbf{Synthesis Strategy} & Transforms design from a ``black-box'' search into a predictable geometric selection. Introduces the \emph{Tangent Prior} to exponentially accelerate optimization algorithms. \\ \hline
\textbf{Morphing Robotics} & Provides a systematic theoretical foundation for pushing further the technology of \emph{Morphing MRAVs}. Demonstrates that MRAVs can continuously reshape (e.g., to adapt to other real-world external constraints) by traversing the solution manifold while maintaining optimal authority. \\ \hline
\end{tabularx}
\end{table*}

To overcome these physical limitations, the robotics community has explored a vast landscape of fully-actuated designs, a diversity recently formalized in extensive reviews and taxonomies~\citep{Rashad2020Review, Hamandi2021Taxonomy}. Early efforts to achieve full actuation involved redundant variable-tilt quadrotor configurations~\citep{Ryll2015Holocopter}
which were then later refined in  implementations like the Voliro~\citep{Kamel2018Voliro}. More recent innovations include the addition of actuated gimbal mechanisms to redirect thrust vectors~\citep{Nigro2021}, and  morphing airframes that adapt their geometry for energy efficiency~\citep{Aboudorra2024}.
However, the most prevalent solution for robust full actuation remains the Non-Collinear Fixedly Tilted Propeller (NCFTP) architecture. Pioneered by~\cite{Toratani2012} 
and~\cite{Jiang2013Dexterous},  these platforms decouple force generation from attitude. This decoupling is crucial for the ``flying end-effector'' paradigm, where the vehicle must exert wrenches in any direction ~\citep{Ryll2019Paradigm}. Specific fully actuated designs that are able to sustain their own weight in any orientation, often termed \textit{Omnidirectional} MRAVs, have been extensively analyzed to maximize the feasible wrench polytope~\citep{Park2016ODAR, Brescianini2016ICRA}.

The operational utility of these platforms is now well-established, evidenced by a surge of literature validating their performance in complex scenarios. Applications range from manipulating soft-robotic arms~\citep{Szasz2022} and executing robust physical interaction~\citep{Bodie2024Book, Brunner2022Tank,Morgado-2023}, to novel maneuvers like inclined docking~\citep{Frankenberg2018}, agile tracking~\citep{Lee2025Omnirotor}, and \secondrev{close-proximity operations~\citep{Lin-2025}}. The aerospace sector has also adopted these platforms as terrestrial emulators for satellite dynamics and formation flying~\citep{McCarthy2024Space}. \secondrev{Concurrently, the fundamental aerodynamic and kinematic properties of these tilted-rotor architectures continue to be rigorously characterized, including the modeling of complex aerodynamic ground effects~\citep{Garofano-2024} and the explicit analysis of maneuverability dictated by specific tilt and cant angles~\citep{Perin-2024}}.

Despite this maturity in application, the geometric design of the platform itself remains a complex challenge. The prevailing design paradigm relies on parametric optimization: defining a cost function and utilizing numerical solvers to find a specific instance that satisfies it.
Following the early pioneers, researchers focused on optimizing tilt angles for specific engineering criteria:~\cite{Rajappa2015Modeling} targeted power consumption;~\cite{Nikou2015Mechanical} optimized relative to desired force and moment capabilities; and~\cite{Lei2017Aerodynamic} focused on minimizing aerodynamic interference in non-planar pairs. Subsequent research refined these objectives, utilizing metrics like the condition number of the allocation matrix in~\cite{Tognon2018Omni}, or constructing specific optimized instances like the ``OmniOcta'' in~\cite{Hamandi2024OmniOcta}.

Recent literature reflects a surge in complexity for these optimization strategies.~\cite{AlZubaidi2025Comparison} highlighted the difficulty of finding a single ``optimal'' design across conflicting requirements like payload versus agility. Similarly, task-specific approaches have gained traction:~\cite{Nikitas2024Thesis} developed generalized procedures for ballast tank inspections, while~\cite{Arza2025Performance} introduced a sophisticated co-optimization framework using Reinforcement Learning (RL) to tailor airframes to specific navigation tasks.

However, these approaches share a common limitation: they treat the optimization landscape as a ``black box.'' Whether via gradient descent or modern evolutionary strategies, these methods effectively perform a blind search to extract a \emph{single} optimal design point. They do not reveal the global topological structure of the solution space.
A canonical example is the seminal ``Omnicopter'' by 
\cite{Brescianini2018}. By leveraging discrete symmetries, they derived \emph{a single specific static} configuration that maximizes isotropy for regular configurations with $N=6,8,12,20$ propellers. While physically effective, \emph{our topological framework reveals that their solution for, e.g., $N=8$, is merely a single coordinate on a continuous, 3-branch, and 1-dimensional manifold of equally optimal designs}. By restricting the search to static parameters, previous methodologies were blind to the existence of this continuous ``valley'' of optimality.

This work departs from the parametric search paradigm to conduct a systematic topological investigation. We ask: \textit{Given a fixed arrangement of rotor positions, what is the global topology of the manifold of optimal orientations?} 

The mathematical complexity of this problem is significant, requiring the topological analysis of critical points for a singular-value-based function defined over the Cartesian product of $N$ Projective Lines. Given that a classical deductive derivation of the global optima for generic $N$ is analytically intractable, we adopt an approach of \emph{systematic computational exploration} and \emph{Experimental Mathematics}. By exploring the design manifold via high-density stochastic global optimization, we uncover a striking and counter-intuitive geometric order. Crucially, this data-driven rigor allows us to transcend isolated numerical solutions and formulate a general \emph{Generative Law} based on Star Polygons, which analytically predicts the optimal topological structure of the landscape of optimal designs for arbitrary $N$ in polygonal chassis.

Our primary contributions are:
\begin{enumerate}
    \item \textbf{Manifold Formulation:} We reframe the optimization problem on the global design manifold $\Manifold = (\RPtwo)^N$, eliminating redundancies associated with vector directionality and focusing strictly on the lines of action.
    \item \textbf{Topological Phase Transition:} We identify a structural transition in the optimization landscape. For generic (irregular) chassis, the global optima are isolated points. However, as the chassis approaches regularity, the solution space collapses onto a  \emph{newly introduced} $N$-dimensional Tangent Torus $\tangentTorus$, a subspace where all thrust vectors are orthogonal to their structural radii.
    \secondrev{\item \textbf{The N-5 Scaling Law:} Based on extensive topological data analysis of regular geometries up to $N=20$, we formulate the \emph{N-5 Scaling Law}. We demonstrate that for configurations up to $N \le 7$, the optimal points synchronize via ``Affine Phase Locking'' to form exactly $K = N-5$ strictly disconnected 1D loops. For higher-$N$ systems, we reveal that these isolated loops become the topological core trajectories embedded within expanded multi-dimensional optimal hypersurfaces.}
    \item \textbf{The Star Polygon Isomorphism:} We reveal a fundamental geometric isomorphism between the set of global optima and the family of admissible Star Polygons $\{N/q\}$. We derive a closed-form \emph{Generative Law} that analytically predicts the exact optimal phase offsets for arbitrary $N$ without the need for numerical optimization, hinting that the ``Affine Phase Locking'' is likely a manifestation of harmonic geometric resonance.    
\end{enumerate}

To understand the physical significance of this topological collapse, one may draw an analogy to a mechanical transmission. In a "Phase Locked" system, such as a train of gears, rotating one gear forces a precise, deterministic rotation in all others; the system possesses a single degree of freedom (1-DOF) despite having many components. Similarly, we find that as the multirotor chassis approaches geometric regularity, the optimal lines of action "lock" into a 1-DOF continuous manifold. The design space effectively acts as a virtual mechanism, allowing the designer to vary parameters continuously without breaking the optimality condition.

To provide an immediate overview of the problem scope, assumptions, and key findings, we present a structured research dossier in Table~\ref{tab:research_dossier}.

\secondrev{The remainder of this paper is organized as follows. Section~\ref{sec:methodology} outlines the overall synthesis framework and the research paradigm. Section~\ref{sec:geometric_characterization} establishes the geometric characterization of the actuation manifold and the relevant configuration space. Section~\ref{sec:optimization} details the Log-Volume potential metric and the global search strategy over $(\RPtwo)^N$. Section~\ref{sec:topology} presents the topological analysis of the optimal orientation space, describing the initial collapse of the solutions onto the Tangent Torus. Section~\ref{sec:scaling_law} details the secondary dimensionality reduction to a 1D curve and introduces the $N-5$ scaling law. Section~\ref{sec:high_n_analysis} investigates the evolution of the solution landscape for increasing $N$, mapping the expansion of these core trajectories into multi-dimensional hypersurfaces. Section~\ref{sec:empirical_law} consolidates the empirical $N-5$ scaling law alongside a unified chassis classification. Section~\ref{sec:star_polygons} provides a theoretical unification of these results, establishing a geometric isomorphism between the global optima and star polygons. Section~\ref{sec:implications} discusses the practical implications for design redundancy and synthesis strategies, while Section~\ref{sec:dynamic_reconfiguration} illustrates an application scenario focused on dynamic reconfiguration and aerodynamic footprinting. Finally, Section~\ref{sec:reproducibility} describes the open-source software package released to ensure reproducibility, and Section~\ref{sec:conclusions} summarizes the contributions and outlines directions for future research.}

\section{Synthesis Framework for Fully-Actuated and Omnidirectional Aerial Systems}
\label{sec:methodology}

This study adopts a research paradigm best described as \textit{Experimental Mathematics} combined with \textit{Mathematical Phenomenology}. Unlike traditional numerical design approaches that seek a single local minimum for a specific engineering instance, we utilize computation as a ``telescope'' to observe the global topological behavior of the design space.

Our workflow treats the optimization problem not merely as a search for a single solution, but as an exploration of the \textit{Solution Landscape}. The cost function $J$, defined in~\eqref{eq:cost_logvolume} by the singular values of the Grasp Matrix (via the Log-Volume metric), induces a scalar field over the configuration space $\mathcal{M} = (\RPtwo)^N$. By systematically varying the input parameters--specifically the symmetry of the chassis (e.g., irregular, semi-regular, and regular shapes)--we treat the resulting optimal configurations as distinct ``species'' of solutions. This allows us to classify the solution morphology based on the underlying symmetries of the input geometry.

The investigation proceeds through three primary discovery phases, anchored by an analytical step that bridges the gap between numerical observation and geometric theory:

\subsection{Phase I: Generative Exploration via Multi-Start Optimization}
To map the global topology of the cost function, we employ a \textit{High-Density Multi-Start Optimization} strategy. Since the global structure of the manifold is unknown a priori, we initialize the local optimizer from a massive set of randomized starting configurations uniformly distributed over the design space $(\RPtwo)^N$.

For each sample, the solver descends to \firstrev{a near local minimum}. While a single run yields only a static point, the aggregation of extensive trials reveals, if present, the hidden connectivity of the solution space. In the case of generic (irregular) chassis, the solver consistently converges to a discrete set of isolated coordinates. However, for regular chassis, a fundamental \textit{Topological Phase Transition} is observed: the solver halts exclusively at configurations where every thrust vector is orthogonal to the line connecting the center of the chassis with the corresponding vertex where the force is applied. This effectively collapses the candidate solution space from the generic $2N$-dimensional $(\RPtwo)^N$ onto the $N$-dimensional Tangent Torus $\tangentTorus$ (see Definition~\ref{def:tangent_torus}), providing the first geometric constraint of the optimal manifold.

\subsection{Phase II: Abductive Reasoning (Topological Extraction)}
The raw data generated by the Multi-Start process--now confined to the Tangent Torus--are processed by a custom \textit{Topological Mode Extraction} algorithm. This phase constitutes the ``abductive'' step of the investigation, utilizing dimensionality reduction to identify the hidden algebraic structure within the toroidal data:
\begin{itemize}
    \item \textbf{Intrinsic Dimensionality Reduction:} We transition from the extrinsic representation of thrust vectors to an intrinsic parameterization of the Tangent Torus. 
    \secondrev{By projecting the data onto the local phase coordinates, we reveal a \emph{secondary topological collapse}: for low-to-medium configurations ($N \le 7$), the solutions are confined to a union of isolated 1-dimensional closed curves, whereas for higher-order systems ($N \ge 8$), these pathways form the core trajectories within expanded multi-dimensional optimal hypersurfaces.}
    
    \item \textbf{Affine Phase Locking:} We demonstrate that these 1D manifolds are governed by linear coordination laws. Through regression analysis of the phase angles, we extract the precise ``Affine Phase Locking'' parameters (unitary slopes and rational offsets), showing that every rotor is synchronously coupled to a single global driving parameter.
\end{itemize}

\secondrev{It is through this segmentation that the ``$N-5$ Scaling Law'' empirically emerges. For $N \le 7$, the law defines exactly $K=N-5$ disconnected 1D loops (isomers) governed by quantized phase shifts; for higher-order architectures ($N \ge 8$), these configurations generalize into $N-5$ distinct core trajectories embedded within broader, multi-dimensional optimal hypersurfaces.}

Finally, by synthesizing the results across the full library of analyzed chassis, we construct a \emph{Unified Taxonomy of Solution Manifolds} (Table~\ref{tab:unified_classification}). This classification maps the spectrum of topological behaviors: from the rigid \textit{Discrete Point Sets} (Type I) typical of irregular chassis, through intermediate \textit{Unstructured Clouds} (Type II/III), to the coherent \emph{Type IV} collapse. 
\secondrev{Within this final category, we distinguish between the isolated 1D isomers governed strictly by the $N-5$ Law at lower dimensions (Type IV-B) and the high-$N$ regime (Type IV-C), where the core trajectories expand into multi-dimensional flat hypersurfaces along additional null space directions, reducing the effective spatial separation between adjacent topological branches.}

\subsection{Phase III: Theoretical Unification (Star Polygon Law)}
In the final step, we move from empirical observation to geometric synthesis. By analyzing the affine slopes identified in Phase II for regular polygonal chassis, we identify an  isomorphism between the $N-5$ optimal phase patterns and the set of star polygons $\{N/q\}$ with $2<q<N-2$. This allows us to predict the optimal topologies analytically, explaining the ``Affine Phase Locking'' as a manifestation of harmonic geometric resonance (Section~\ref{sec:star_polygons}).

\subsection{Foundational Value of the Work}
This work lays the foundation for a new topological theory of robotic aerial vehicle design. Rather than offering a closed chapter, we provide a \emph{blueprint for a new field of investigation in robotic design}: moving beyond the search for static optimal points to the exploration of continuous optimal manifolds. 

A first step toward this future exploration is provided by the \emph{Generative Law} derived in this work, which demonstrates that the complex optimal landscapes of high-$N$ systems are not random, but are governed by elegant, predictable integer sequences rooted in the fundamental symmetries of the chassis.

\section{Geometric Characterization of the Actuation Manifold}
\label{sec:geometric_characterization}

\subsection{Dynamical Model of a Generic MRAV}
We consider a generic Multi-Rotor Aerial Vehicle (MRAV) consisting of a rigid chassis and $N$ propeller units. A body-fixed frame $\mathcal{F}_B$ is attached to the Center of Mass (CoM). The equations of motion are governed by the Newton-Euler formalism (see, e.g.,~\cite{Hamandi2021Taxonomy}):
\begin{equation}
\label{eq:dynamics}
\begin{bmatrix}
m \bm{\ddot{p}}_{B} \\
\bm{J}\bm{\dot{\omega}}_B
\end{bmatrix} = -\begin{bmatrix}
mg \bm{e_3} \\
\bm{\omega}_B \times \bm{J}\bm{\omega}_B
\end{bmatrix} + \begin{bmatrix}
^B\bm{f}^{\text{rot}}_B \\
^B\mathbf{m}^{\text{rot}}_B
\end{bmatrix} + \bm{w}_{ext}
\end{equation}
where $m$ and $\bm{J} \in \mathbb{R}^{3\times3}$ are the mass and inertia tensor, respectively; $\mathbf{p}_{B}$ and $\bm{\omega}_B$ denote the inertial position and body angular velocity. The term $\bm{w}_{ext}$ encompasses external disturbances and interaction wrenches.

The control input to the system is the vector of signed thrusts $\bm{u} = [u_1, \dots, u_N]^\top \in \mathbb{R}^N$, where the thrust is proportional to the square of the rotor speed ($u_i \propto \omega_{i} |\omega_{i}|$). 
\firstrev{For the development of this work} we assume the use of bidirectional Electronic Speed Controllers (ESCs)\footnote{Such hardware is now an industry standard~\cite{GONZALEZMORGADO2025276}, utilizing protocols like Bidirectional DShot, see~\cite{DShotStandard}, and mature firmware ecosystems such as BLHeli\_32, see~\cite{BLHeli32}, 
Bluejay, or AM32, which enable real-time RPM telemetry.},
allowing $u_i$ to take both positive and negative values. \secondrev{However, we demonstrate in Sec.~\ref{sec:unidirectional_extension} that the theoretical findings of this work strongly inform and establish the geometric bounds for the case of standard unidirectional propellers as well}.

The mapping from these inputs to the body wrench is linear:
\begin{equation}
\begin{bmatrix}
^B\bm{f}^{\text{rot}}_{B}\\
^B\mathbf{m}^{\text{rot}}_{B}
\end{bmatrix} = \bm{A} \bm{u}
\end{equation}
where $\bm{A} \in \mathbb{R}^{6\times N}$ is the \emph{Grasp Matrix} (or Allocation Matrix).

While control literature typically assumes $\bm{A}$ is given, in this work, $\bm{A}$ is the primary variable of interest. Specifically, we investigate the geometric structure of $\bm{A}$ when the \emph{rotor positions} are fixed parameters of the chassis, and the \emph{rotor directions} are the free design variables.

\begin{definition}[The Chassis]
A Chassis $\mathcal{C}$ is defined as the rigid geometrical structure whose vertices correspond to the points of application of the control forces generated by the propellers. The set of positions of the vertices of $\mathcal{C}$ relative to $\mathcal{F}_B$ is denoted as $\PosSet = \{\pvec_1, \dots, \pvec_N\}$ with $\pvec_i\in \mathbb{R}^3$.
\end{definition}

\begin{figure}[t]
    \centering
     \includegraphics[width=0.99\linewidth]{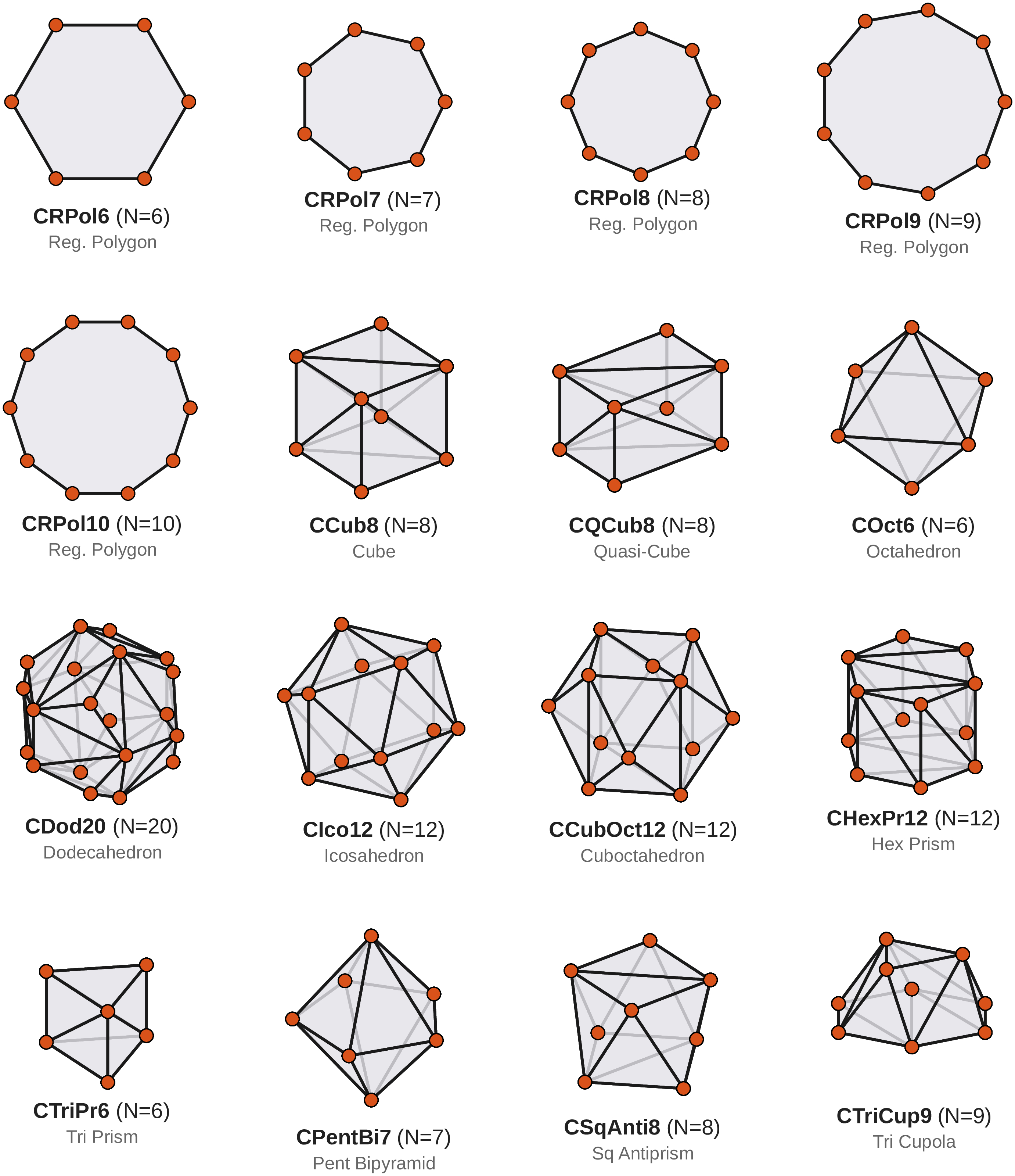}
    \caption{\textbf{Visual dictionary of representative chassis.}
    The designs range from planar regular polygons (Top Row) to complex 3D polyhedra (Bottom Rows).
    Each sub-figure displays the chassis geometry (grey faces and black edges) and the points of application of the propeller forces (orange vertices).
    The labels indicate the unique Identifier (ID) used throughout the paper and the number of rotors ($N$).
    See Table~\ref{tab:chassis_zoo} for full specifications.}
    \label{fig:geometric_zoo}
\end{figure}

A comprehensive list of the main chassis analyzed in this study is provided in Table~\ref{tab:chassis_zoo}, with corresponding visual representations in Figure~\ref{fig:geometric_zoo}.

\begin{table}[b]
\centering
\caption{Geometric definitions of the chassis analyzed, including regular, quasi-regular, and irregular variants.}
\label{tab:chassis_zoo}
\resizebox{\columnwidth}{!}{%
    \begin{tabular}{llc}
    \toprule
    \textbf{ID} & \textbf{Chassis Name (Type)} & \textbf{Rotors ($N$)} \\
    \midrule
    CRPol6 & Reg. Hexagon \textit{(Polygon)} & 6 \\
    CRPol7 & Reg. Heptagon \textit{(Polygon)} & 7 \\
    CRPol8 & Reg. Octagon \textit{(Polygon)} & 8 \\
    CRPol9 & Reg. Nonagon \textit{(Polygon)} & 9 \\
    CRPol10 & Reg. Decagon \textit{(Polygon)} & 10 \\
    \midrule
    CQRPol6 & Quasi-Reg. Hexagon \textit{(Polygon)} & 6 \\
    CQRPol7 & Quasi-Reg. Heptagon \textit{(Polygon)} & 7 \\
    CQRPol8 & Quasi-Reg. Octagon \textit{(Polygon)} & 8 \\
    CQRPol9 & Quasi-Reg. Nonagon \textit{(Polygon)} & 9 \\
    CQRPol10 & Quasi-Reg. Decagon \textit{(Polygon)} & 10 \\
    \midrule
    COct6 & Octahedron \textit{(Platonic Solid)} & 6 \\
    CCub8 & Cube \textit{(Platonic Solid)} & 8 \\
    CIco12 & Icosahedron \textit{(Platonic Solid)} & 12 \\
    CDod20 & Dodecahedron \textit{(Platonic Solid)} & 20 \\
    \midrule
    CTriPr6 & Tri. Prism \textit{(Prismatic)} & 6 \\
    CPentBi7 & Pent. Bipyramid \textit{(Johnson Solid)} & 7 \\
    CQCub8 & Quasi-Cube \textit{(Distorted Solid)} & 8 \\
    CSqAnti8 & Sq. Antiprism \textit{(Prismatic)} & 8 \\
    CTriCup9 & Tri. Cupola \textit{(Johnson Solid)} & 9 \\
    CCubOct12 & Cuboctahedron \textit{(Archimedean)} & 12 \\
    CHexPr12 & Hex. Prism \textit{(Prismatic)} & 12 \\
    \bottomrule
    \end{tabular}%
}
\end{table}

\subsection{The Affine Grassmannian Formulation}

To rigorously analyze the design space, we must move beyond simple Euclidean vectors. A single propeller force is physically characterized by its \emph{line of action} in space. While the magnitude is a control variable, the line itself is a geometric constant determined by the design.

The mathematical space of all lines in $\mathbb{R}^3$ is a 4-dimensional manifold called the \emph{Affine Grassmannian}, denoted as $\Graff$.
To perform algebraic optimization on $\Graff$, we parameterize the $i$-th line using normalized Plücker coordinates $\mathbf{h}_i = (\mathbf{d}_i, \mathbf{m}_i) \in \mathbb{R}^6$.
Here:
\begin{itemize}
    \item $\mathbf{d}_i \in \SphereTwo$ is the unit direction vector of the thrust force.
    \item $\mathbf{m}_i = \mathbf{p}_i \times \mathbf{d}_i + c_{\tau f}  \mathbf{d}_i$ is the total moment vector.
\end{itemize}
Note that for the geometric optimization of the \emph{primary} wrench capacity in fully actuated and omnidirectional  platforms, the 
combination of the effects of aerodynamic drag moments (the $c_{\tau f}  \mathbf{d}_i$'s) is typically negligible compared to the combination of the effects of the lever-arm moments (the $\mathbf{p}_i \times \mathbf{d}_i$'s). Supported by the literature, in this geometric study, we focus on the dominant term $\mathbf{m}_i \approx \mathbf{p}_i \times \mathbf{d}_i$ to find the structural optima\footnote{Neglecting the drag moment is standard in the literature (e.g., \cite{ Brescianini2018,Lee2025Omnirotor}). These works demonstrate that geometric optima derived from the dominant thrust term ($\mathbf{p}_i \times \mathbf{d}_i$) yield highly agile platforms, where second-order aerodynamic drag effects are effectively rejected by robust feedback control.}. This satisfies the classical Klein Quadric constraint $\mathbf{d}_i \cdot \mathbf{m}_i = 0$. Together with the unit norm constraint of $\mathbf{d}_i$, this restricts $\mathbf{h}_i$ to the 4-dimensional manifold of valid lines in $\Graff$.

The Grasp Matrix is thus constructed column-wise:
\begin{equation}
    \bm{A} = \begin{bmatrix} \mathbf{d}_1 & \dots & \mathbf{d}_N \\ \mathbf{m}_1 & \dots & \mathbf{m}_N \end{bmatrix}
\end{equation}
representing an element of the product manifold $\Graff^N$.

\emph{Remark (Aerodynamic Considerations):} 
This study restricts its scope to the \emph{geometric authority} of the platform, defined by the grasp matrix $\mathbf{A}$. While real-world rotor performance is coupled with aerodynamic effects (e.g., drag moments, blade flapping, inflow interference), we adopt the standard decoupling assumption found in the foundational and recent literature~\citep{Brescianini2018,Park2016ODAR,Lee2025Omnirotor}. 
These works demonstrate that designs optimized purely for geometric isotropy successfully reject aerodynamic residuals as state-dependent disturbances via robust feedback control.
The rationale is thus hierarchical: a configuration that is geometrically singular ($\det(\mathbf{A}\mathbf{A}^\top) \to 0$) cannot be recovered by aerodynamic nuances, whereas a topologically robust geometric optimum provides the maximum control authority margin to handle unmodeled dynamics. \secondrev{However, it is important to explicitly note that while unmodeled aerodynamic interactions (such as downwash interference and drag torque) do not alter the definition of the geometric structural optimum under the adopted static grasp-map model, they represent a critical secondary layer that will alter the actual energy efficiency and shift the true physical optimum during high-speed or high-load dynamic morphing.}

\subsection{The Projective Design Manifold}
\label{subsec:proj_design_manifold}

Given a chassis $\mathcal{C}$, the design freedom is restricted to the orientation of the line passing through each fixed position $\mathbf{p}_i$.
Conventionally, one might model this orientation as a point on the sphere $\SphereTwo$. However, since the actuators are bidirectional ($u_i \in [-u_{max}, u_{max}]$), a thrust vector $\mathbf{d}_i$ and its opposite $-\mathbf{d}_i$ generate the exact same \emph{line of action} and wrench capacity limits.
Therefore, the intrinsic configuration space for a single rotor is not the sphere, but the \emph{Real Projective Plane} $\RPtwo$, which is the sphere with antipodal points identified. Formally, for our purpose, the coordinates $\dvec_i$ and $-\dvec_i$ represent the same physical element $\mathcal{D}_i\in \RPtwo_i$.

Given a chassis $\mathcal{C}$, the total \emph{Design Manifold} is the Cartesian product:
\begin{equation}
    \Manifold = \prod_{i=1}^N \RPtwo_i \cong (\RPtwo)^N
\end{equation}
This manifold is compact, non-orientable, and has dimension $2N$. This formulation eliminates the redundancy of vector directionality, ensuring that each point in $\Manifold$ corresponds to a unique physical architecture. 

\begin{definition}[The Tangent Torus]
\label{def:tangent_torus}
Given a chassis $\mathcal{C}$, we identify a critical $N$-dimensional submanifold of the design space $\Manifold$, denoted as the \textbf{Tangent Torus} $\tangentTorus \subset \Manifold$. This is the set of all thrust directions (lines of force) in $\Manifold$ that are orthogonal to the position vector of their respective vertices:
\begin{equation}
    \tangentTorus = \{ (\mathcal{D}_1,\ldots,\mathcal{D}_N) \in \Manifold \mid \dvec_i \cdot \pvec_i = 0, \; \forall i \}
\end{equation}
Geometrically, this restricts each rotor line of force  to the projective line  $\RPone$ corresponding  to the plane tangent to the sphere at $\pvec_i$.
\end{definition}

\subsection{Visualization Methodology: The Flattened Manifold}
\label{sec:visualization}

\begin{figure}[t]
    \centering
    \includegraphics[width=0.9\linewidth]{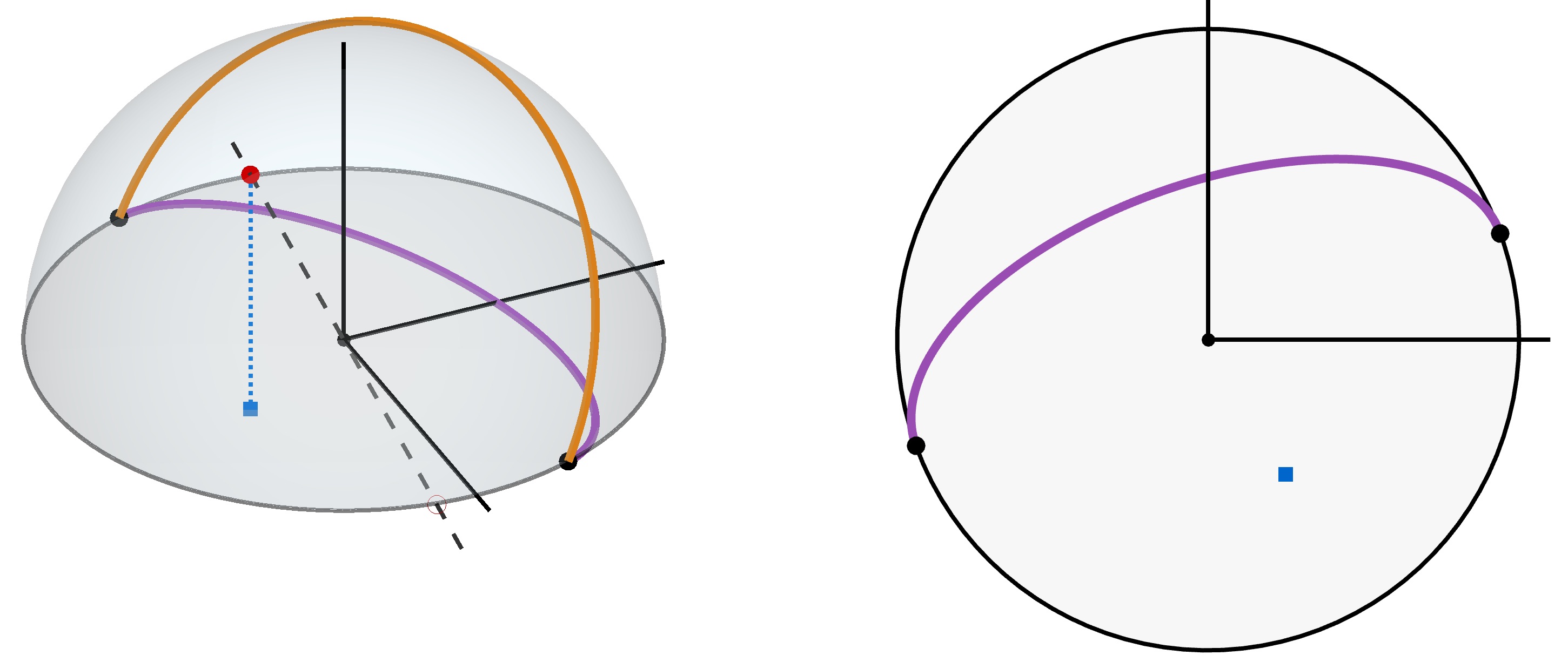}
    \caption{\textbf{Geometric construction of the $\RPtwo$ Disc Model.} 
    \textbf{(Left)} A 3D view of the upper hemisphere of $\SphereTwo$ and the equatorial disc. A rotor's line of action (dashed line) passes through the origin, intersects the hemisphere at a {red dot}, and projects orthographically to a {blue square} on the disc. 
    \textbf{(Right)} The resulting 2D disc representation.
    \textbf{Topological Loops:} We also visualize a full rotation of a line about a fixed axis (effectively a plane of lines passing through the origin). On the hemisphere, this traces an {orange great semi-circle} connecting two antipodal black dots. When projected onto the disc, this forms a {purple curve}. Geometrically, this curve is a semi-ellipse (degenerating to a straight diameter or a circular arc depending on the axis tilt). Because the antipodal boundary points are topologically identified, this purple curve connects back to itself, forming a continuous closed loop in $\RPtwo$.}
    \label{fig:disc_concept}
\end{figure}

To analyze the global solution space $\Manifold$, we require a visualization method that respects the non-trivial topology of the Projective Plane while remaining intuitive for design. Visualizing $N$ lines in 3D space is typically cluttered and obscures subtle symmetries. 
We introduce a \emph{Moduli Disc Representation}.
Since $\RPtwo$ is topologically equivalent to a closed unit disc with antipodal boundary points identified, we establish a canonical mapping for visualization.
From the pair of antipodal unit vectors representing the same physical element $\mathcal{D}_i \in \RPtwo$, we select a representative $\mathbf{d}_i = [x_i, y_i, z_i]^\top$ that resides in the upper hemisphere (i.e., satisfying $z_i \ge 0$).
We then project this vector onto the 2D equatorial plane using an orthographic projection, effectively flattening the manifold into a readable domain.

The geometric construction of this mapping is detailed in Fig.~\ref{fig:disc_concept}. 
On the left, we observe the 3D setup: a generic line of action (dashed line) passes through the origin and intersects the upper hemisphere at a single point (indicated by a red dot). This point is projected orthographically onto the equatorial disc, resulting in a unique 2D coordinate (indicated by a blue square dot). On the right, we see the resulting 2D representation used for analysis.

Crucially, this model allows us to visualize continuous rotations of the thrust vector. A full rotation of a line about a fixed axis corresponds to a plane of lines slicing through the origin. On the hemisphere (Fig.~\ref{fig:disc_concept}, left), this intersection forms a great semi-circle (highlighted in orange) that starts and ends at two antipodal black dots on the equator. 
When projected onto the disc, this arc forms a purple curve connecting the boundary points. While this curve appears to be an open semi-ellipse in Euclidean 2D space, the topological identification of antipodal boundary points ($x \sim -x$) means that the two black endpoints are, in fact, the same point in $\RPtwo$. Consequently, this purple curve represents a single, seamless \emph{closed loop} in the configuration space of rotor directions.

\begin{figure}[t]
    \centering
    \includegraphics[width=0.48\linewidth]{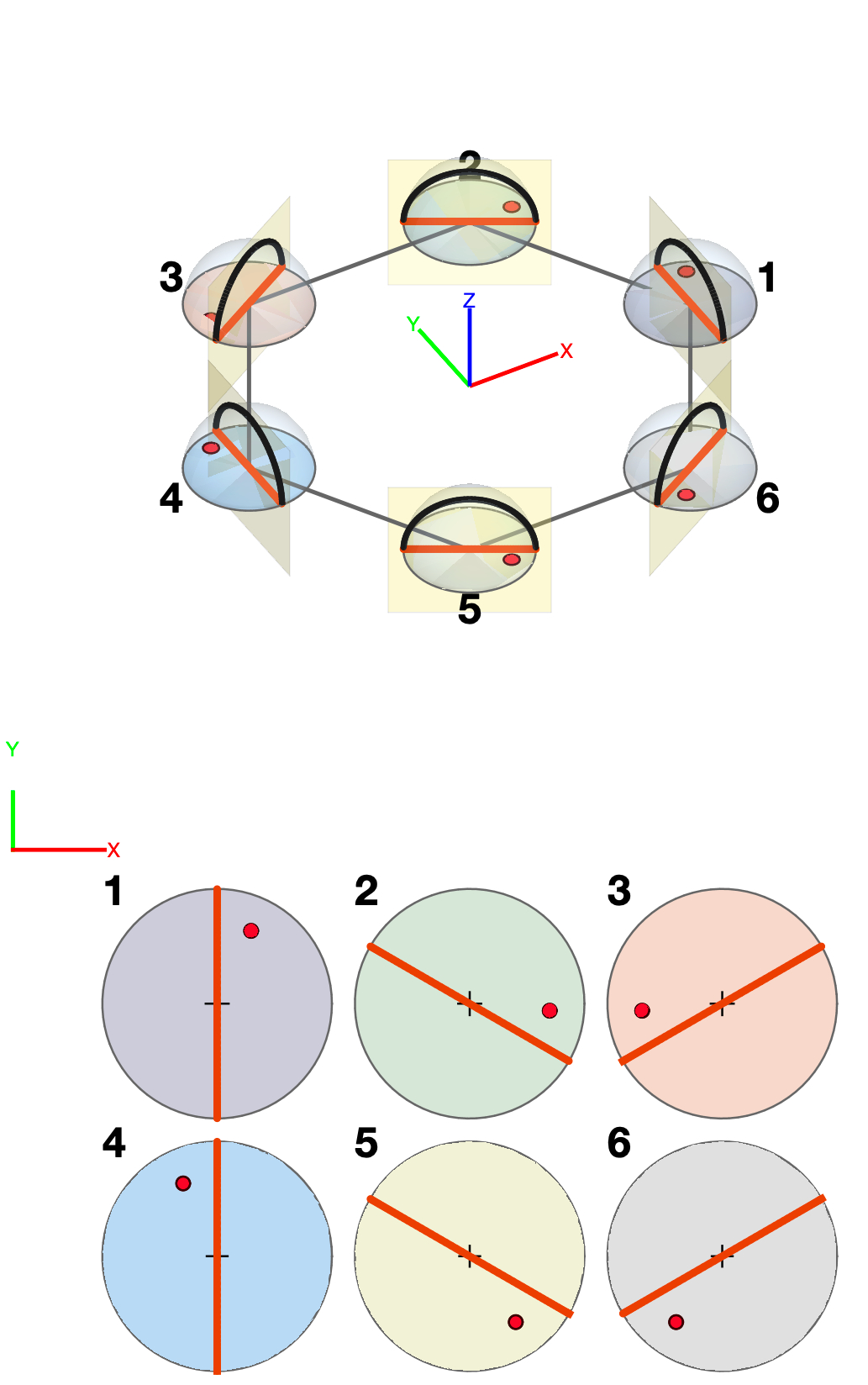}
    \hfill\includegraphics[width=0.48\linewidth]{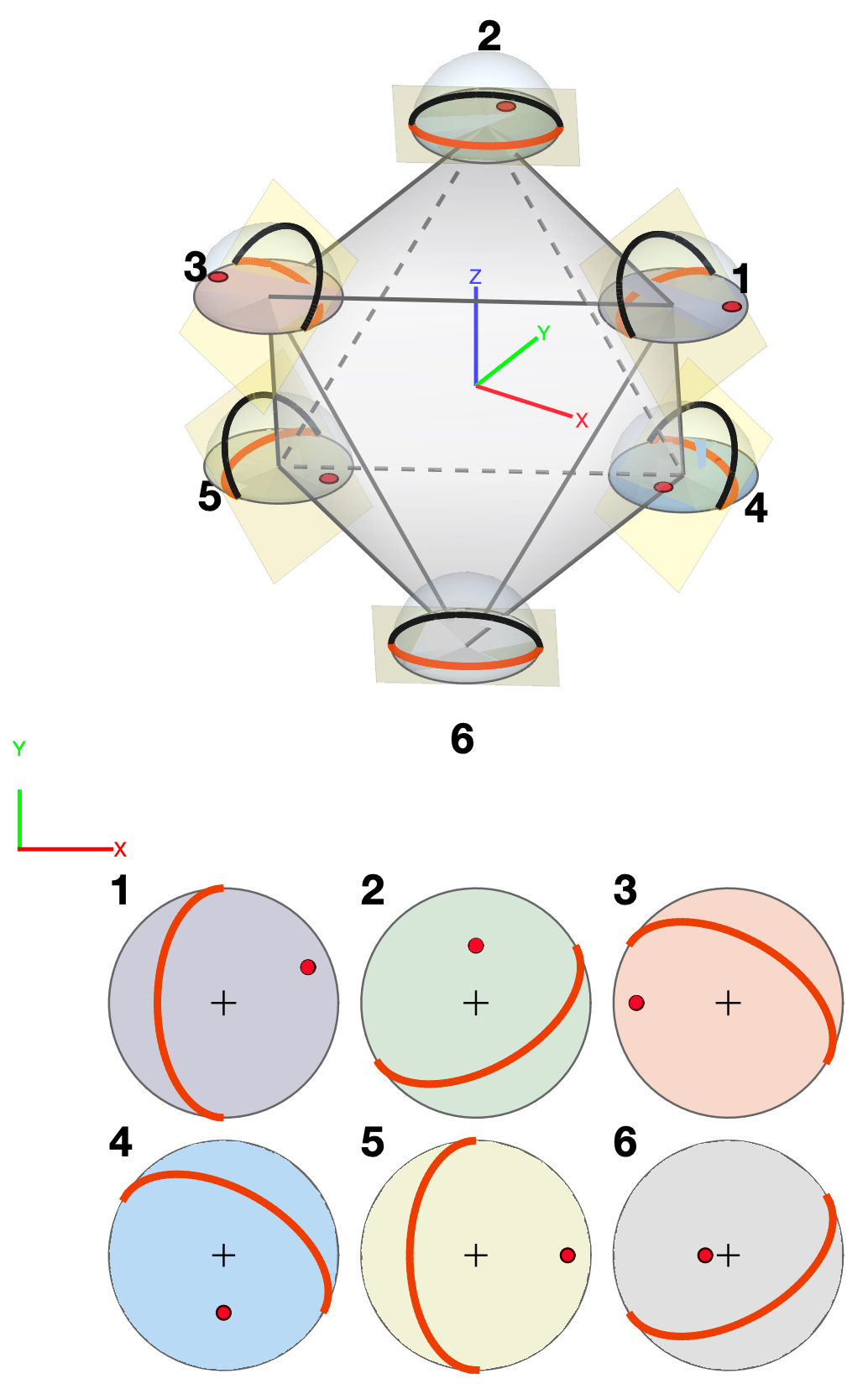}
    \caption{\textbf{The disc dashboard view of $\Manifold=(\RPtwo)^N$:} a graphical model of the manifold of all force directions for two representative chassis.
    Configuration spaces for a regular hexagon (Top-Left) and an octahedron (Top-Right), where $N=6$.
    Each chassis vertex is equipped with a local copy of the $\RPtwo$ disc model defined in Fig.~\ref{fig:disc_concept}, representing the manifold of possible force directions at that specific rotor location.
    \textbf{Generic Configuration:} The red markers (\textbf{1--6}) represent the projections of a generic set of rotor directions (analogous to the {blue square} in Fig.~\ref{fig:disc_concept}). The bottom sub-figures provide a \emph{dashboard view} of these six discs.
    \textbf{The Tangent Torus:} To visualize the subspace of tangential forces lines represented by $\tangentTorus$, we illustrate the intersection of the local upper hemispheres with planes tangent to the sphere centered at the chassis origin and passing through the vertices of the chassis. These intersections project as semi-ellipses (orange curves) on the discs, geometrically identical to the {purple loops} described in Fig.~\ref{fig:disc_concept}. A configuration on $\Manifold$, e.g., a solution of~\eqref{eq:optimization},  lies on the Tangent Torus if all the $N$ markers in each disc lie on their respective orange curves.}
    \label{fig:RP2N}
\end{figure}

We extend this local representation to the full chassis in Fig.~\ref{fig:RP2N}. Here, we assign a local $\RPtwo$ disc to each of the $N$ vertices of the platform. This provides a complete \emph{dashboard view} of the manifold $\Manifold = (\RPtwo)^N$, allowing us to identify geometric patterns--such as the synchronization of rotor tilts along specific curves--that would be hard to visually process for the human eye in a standard 3D Cartesian view.  We make extensive use of this dashboard view throughout the work to compactly visualize the solution landscape.

\subsection{Dimensional Homogeneity and Characteristic Length}
A critical theoretical hurdle in optimizing $\bm{A}$ is unit inhomogeneity. The top three rows represent Force (N), while the bottom three represent Moment (Nm). 
Standard metrics like the Condition Number ($\kappa := \sigma_{\max}/\sigma_{\min}$), where $\sigma_{\max}$ and $\sigma_{\min}$ denote the maximum and minimum singular values, are not scale-invariant w.r.t. to the size of the chassis; changing the length unit alters the singular values of $\bm{A}$ non-uniformly, thereby affecting the metric.

To rigorously define isotropy across the 6-DOF wrench space, we introduce a \emph{characteristic length} $L_c$ and define the \emph{Dimensionless Grasp Matrix} $\bar{\bm{A}}$:
\begin{equation}
    \bar{\bm{A}} = \begin{bmatrix} \mathbf{d}_1 & \dots & \mathbf{d}_N \\ \frac{1}{L_c}\mathbf{m}_1 & \dots & \frac{1}{L_c}\mathbf{m}_N \end{bmatrix}
\end{equation}

We adopt the principle of \emph{Natural Scaling}: for a vertex set $\PosSet$ distributed on a sphere of radius $R$, the condition number of $\bar{\bm{A}}$ reaches its minimum when $L_c = R$. This choice balances the numerical magnitude of the force and moment subspaces, ensuring that the lever arms are normalized relative to the structural scale. Consequently, throughout this work, we fix $L_c = R_{geom}$, where $R_{geom}$ is the radius of the circumscribing sphere of the chassis. The validity of this choice is empirically verified in Section~\ref{sec:sensitivity}, where we demonstrate that the optimization cost function consistently achieves its global minimum at this specific scaling ratio.

\section{Optimization on $(\RPtwo)^N$: Log-Volume Potential and Global Search}
\label{sec:optimization}

\firstrev{When operational requirements demand high-dexterity physical interaction or arbitrary attitude tracking, a primary objective} in the design of a fully-actuated or omnidirectional multirotor aerial platform is the maximization of \textit{isotropy}: the ability of the vehicle to generate force and moment vectors with equal authority in any direction~\citep{Brescianini2018, Park2016ODAR, Tognon2018Omni,Hamandi2021Taxonomy}. 
This prioritization stems from the inherent trade-off of full actuation: the added mechanical complexity, weight, and reduced aerodynamic efficiency are justified only if the platform achieves superior, decoupled control authority that is unattainable by standard underactuated quadrotors.

\firstrev{It is crucial to formally bound the definition of ``optimality'' used within this geometric framework. We define an optimal configuration strictly as one that maximizes the hypervolume of the 6-DOF control authority polytope. Consequently, this topological search specifically isolates highly isotropic designs. Platforms optimized for pure payload capacity or unidirectional efficiency--such as the standard planar hexarotor, which maximizes the thrust-to-weight ratio along the $Z$-axis--possess a 6-DOF control hypervolume of exactly zero. By utilizing the Log-Volume potential, our formulation deliberately excludes strictly collinear (underactuated), or highly non-isotropic architectures from the optimal solution space, treating them as operationally distinct from the class of fully actuated and omnidirectional vehicles studied here.}

Geometrically, this corresponds to transforming the $N$-dimensional hypersphere of normalized motor inputs into a 6-dimensional wrench ellipsoid that is as spherical as possible. Quantitatively, a design is perfectly isotropic when all singular values of the grasp matrix are identical, resulting in a condition number of unity.

To achieve this goal by selecting optimal rotor orientations for a given chassis, we require a robust numerical framework capable of mapping the global landscape of the design manifold $\mathcal{M}$. This section details an objective function designed to prioritize volumetric authority over raw isotropy--thereby avoiding numerical singularities--alongside the computational strategy employed to exhaustively identify the multimodal landscape of solutions.

\subsection{The Degeneracy of Pure Isotropy}

The standard definition of kinematic isotropy is a Condition Number of unity, $\kappa(\bar{\bm{A}}) = 1$. While minimizing $J = \kappa(\bar{\bm{A}})$ is the conventional approach in robotic design, our preliminary topological analysis revealed two critical flaws in this metric:

\begin{itemize}
    \item \textbf{Isotropic Collapse:} The condition number is scale-invariant. The optimizer can theoretically achieve $\kappa=1$ by driving all singular values to zero ($\sigma_i \to 0$). This results in a ``perfectly isotropic'' platform that possesses zero wrench authority---a trivial and physically useless singularity.
    
    \item \textbf{Gradient Instability:} The Condition Number is a non-smooth function; its gradient is undefined at configurations where singular values coincide (i.e., at isotropic points). 
    This non-differentiability creates discontinuities in the gradient field, causing standard gradient-based solvers (like SQP) to exhibit exhibit oscillation or failure to converge~\citep{lewis1996eigenvalue,nocedal2006optimization}.
\end{itemize}

\subsection{The Log-Volume Potential Field}
To resolve these issues, we construct a cost function that acts as a smooth potential field over $\Manifold$. We seek to maximize the volume of the feasible wrench ellipsoid (proportional to the product of singular values) while simultaneously penalizing anisotropy.
We utilize the negative sum of the logarithm of the singular values of the dimensionless grasp matrix:
\begin{equation}
    J_{\text{vol}}(\mathcal{D},\mathcal{C}) = - \sum_{k=1}^6 \ln(\sigma_k(\bar{\bm{A}}(\mathcal{D},\mathcal{C})) + \epsilon)
    \label{eq:cost_logvolume}
\end{equation}
where $\epsilon$ is a small regularization term ($10^{-9}$) to prevent numerical blow-up at rank-deficient points.

We justify this formulation by observing that the sum of logarithms is equivalent to the logarithm of the product. Neglecting $\epsilon$ for the sake of geometric interpretation, Eq.~\eqref{eq:cost_logvolume} can be rewritten as:
\begin{equation}
    J_{\text{vol}} \approx - \ln\left( \textstyle\prod_{k=1}^6 \sigma_k \right) = - \ln\left( \sqrt{\det(\bar{\bm{A}}\bar{\bm{A}}^T)} \right)
\end{equation}
Since $\sqrt{\det(\bar{\bm{A}}\bar{\bm{A}}^T)}$ represents the volume of the manipulability ellipsoid, minimizing $J_{\text{vol}}$ is directly equivalent to maximizing the log-determinant of the Grasp Gramian. 

Eq.~\eqref{eq:cost_logvolume} provides a computationally stable method for maximizing the wrench volume.
In fact, this function offers a few distinct advantages for topological exploration:
\begin{itemize}
    \item \textbf{Smoothness at Optima:} Unlike the Condition Number, the sum of logarithms remains smooth and differentiable even when singular values coincide. This provides the SQP solver with valid gradient information all the way to the isotropic optimum.
    \item \textbf{The Barrier Effect:} As any $\sigma_k \to 0$, $J_{\text{vol}} \to \infty$. This naturally repels the optimizer from kinematic singularities, ensuring that all discovered local minima represent valid, high-authority designs.
    \item \textbf{Projective Well-posedness:} The cost function depends on $\mathbf{d}_i \mathbf{d}_i^\top$, making it invariant to the sign of the thrust vectors. Thus, $J_{\text{vol}}$ is a well-defined scalar field on the projective space $\RPtwo$, respecting the intrinsic topology of the actuator.
\end{itemize}

Given a fixed chassis geometry defined by the set of rotor positions $\PosSet$, the search for the optimal control authority is formulated as a constrained minimization problem over the product manifold:
\begin{equation}
    \label{eq:optimization}
    \begin{aligned}
        \min_{\mathcal{D}} \quad & J_{\text{vol}}(\mathcal{D},\PosSet) \\
        \text{s.t.} \quad & \mathcal{D} \in \mathcal{M} = (\RPtwo)^N
    \end{aligned}
\end{equation}
This formulation ensures that the resulting orientation set $\mathcal{D}^*$ corresponds to a configuration that maximizes the volumetric capability of the wrench space while remaining within the geometrically feasible bounds of the rotor orientations.

\begin{algorithm}[t]
\caption{Global Manifold Exhaustion Strategy}
\label{algo:optimization}
\begin{algorithmic}[1]
\Require Vertex Set $\mathcal{P} = \{\mathbf{p}_1, \dots, \mathbf{p}_N\}$, Characteristic Length $L_c$, Tolerance $\epsilon_{tol}$.
\Ensure Set of Global Solutions $\mathcal{S}^* = \{ \mathcal{D}^*_1, \dots, \mathcal{D}^*_M \}$.

\State \textit{Step 1: Initialization}
\State $\mathcal{S}_{raw} \leftarrow \emptyset$.

\State \textit{Step 2: Sampling \& Optimization Loop}
\For{$k = 1$ to $M$ (e.g., $10^3$ Samples)}
    \State \emph{Random Guess:} Sample $\mathbf{d}_i^{(0)} \sim \text{Unif}(\mathbb{S}^2)$ for $i=1\dots N$.
    \State \emph{Local Refinement (SQP):}
    \begin{align}
         \mathcal{D}^*_{trial} = \arg\min_{\mathcal{D}} \quad & J_{vol}(\mathcal{D}) \notag\\
         \text{s.t.} \quad & \|\mathbf{d}_i\| = 1 \quad \forall i=1\dots N \notag
     \end{align}
    \State \emph{Canonical Mapping:} Map $\mathcal{D}^*_{trial}$ to upper hemisphere (account for $\mathbb{R}P^2$ symmetry):
    \For{$i = 1$ to $N$}
        \If{$\mathbf{d}_i \cdot \mathbf{e}_z < 0$} $\mathbf{d}_i \leftarrow -\mathbf{d}_i$ \EndIf
    \EndFor
    \State Add $(\mathcal{D}^*_{trial}, J_{trial})$ to $\mathcal{S}_{raw}$.
\EndFor

\State \textit{Step 3: Global Identification \& Pruning}
\State $J_{min} \leftarrow \min \{J \mid (\mathcal{D}, J) \in \mathcal{S}_{raw} \}$.
\State $\mathcal{S}^* \leftarrow \{ \mathcal{D} \in \mathcal{S}_{raw} \mid J(\mathcal{D}) \le J_{min} + \epsilon_{tol} \}$.

\State \textbf{Return} $\mathcal{S}^*$.
\end{algorithmic}
\end{algorithm}

\subsection{Computational Optimization Strategy}

The configuration space $\mathcal{M}$ is inherently non-convex and high-dimensional. Consequently, local gradient-based methods are susceptible to entrapment in suboptimal basins. To address this, we implement a Global Manifold Exhaustion strategy, detailed in Algorithm~\ref{algo:optimization}.
The approach hinges on a decoupled two-stage process: a massive stochastic initialization followed by deterministic local refinement. This allows us to map the multimodal landscape of the manifold rather than merely finding a single point solution.

As outlined in Algorithm~\ref{algo:optimization}, we employ a multi-start Sequential Quadratic Programming (SQP) approach. Each execution begins with a Monte Carlo sampling of $M \geq 10^3$ starting configurations scattered uniformly and pseudo-randomly across $\mathcal{M}$. Following the local convergence of each sample, we apply a canonical mapping to the $\RPtwo$ disc model to resolve antipodal symmetries (ensuring $z_i \geq 0$).

Crucially, to distinguish true global optima from high-quality local minima, we apply a \emph{pruning} step. We define the global minimum cost $J_{min}$ found across the entire batch and retain only the subset of solutions $\mathcal{S}^*$ whose cost lies within a numerical tolerance $\epsilon_{tol}$ of $J_{min}$. This pruned dense point cloud forms the basis for the phenomenological analysis in Section~\ref{sec:topology}, where distinct solution families are identified through visual inspection of the projected solution density on the disc dashboard for each chassis.

\begin{figure}[t]
    \centering
     \includegraphics[width=0.99\linewidth]{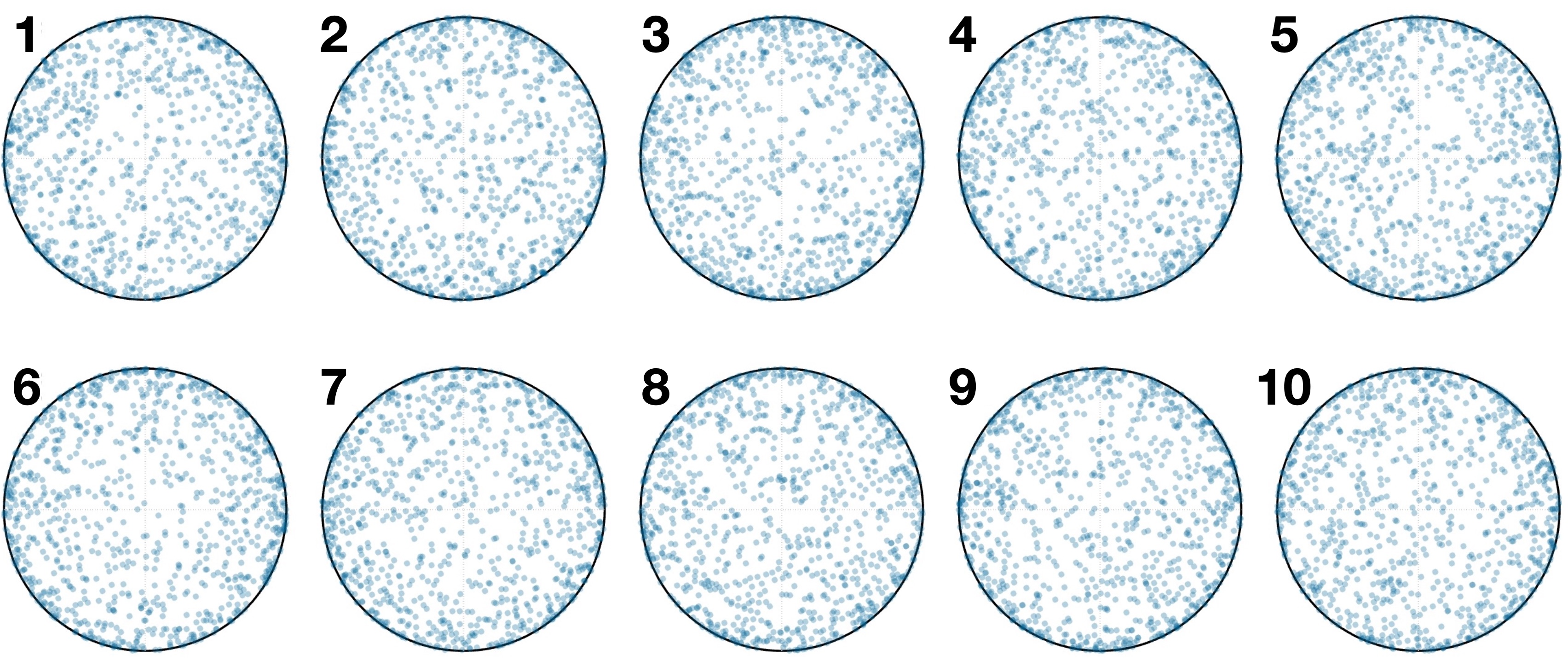}
    \caption{\textbf{Stochastic initialization of the search space for an $N=10$ chassis.}
    Each disc of the dashboard represents the projection on the local orientation manifold $\RPtwo_i$ of the $i$-th rotor.
    The scatter plots depict the $2$-dimensional projections on the discs of $10^3$ random starting $2N$-dimensional configurations sampled uniformly across the product manifold $\mathcal{M} = (\RPtwo)^N$.   
    The uniform coverage confirms a state of maximal configuration entropy, ensuring an unbiased initialization of the global optimization landscape.}
    \label{fig:monte_carlo}
\end{figure}

\subsection{Sensitivity Analysis and Characteristic Length}
\label{sec:sensitivity}
As established in Section~\ref{sec:geometric_characterization}, defining the Grasp Matrix requires a characteristic length $L_c$ to homogenize the units of force and moment. To verify the proposition that $L_c = R_{geom}$ is the natural scale for these systems, we performed a sensitivity analysis of the optimized metrics across the chassis database.

\begin{figure}[t]
    \centering
\includegraphics[width=0.99\linewidth]{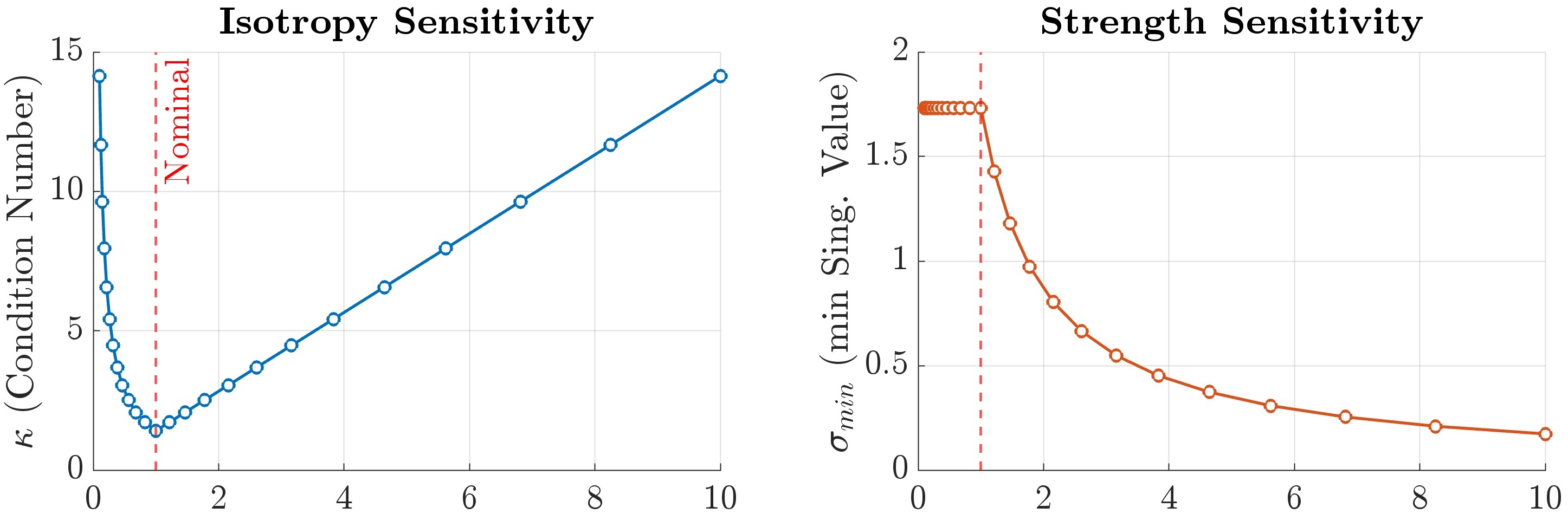}\\\smallskip
      \includegraphics[width=0.99\linewidth]{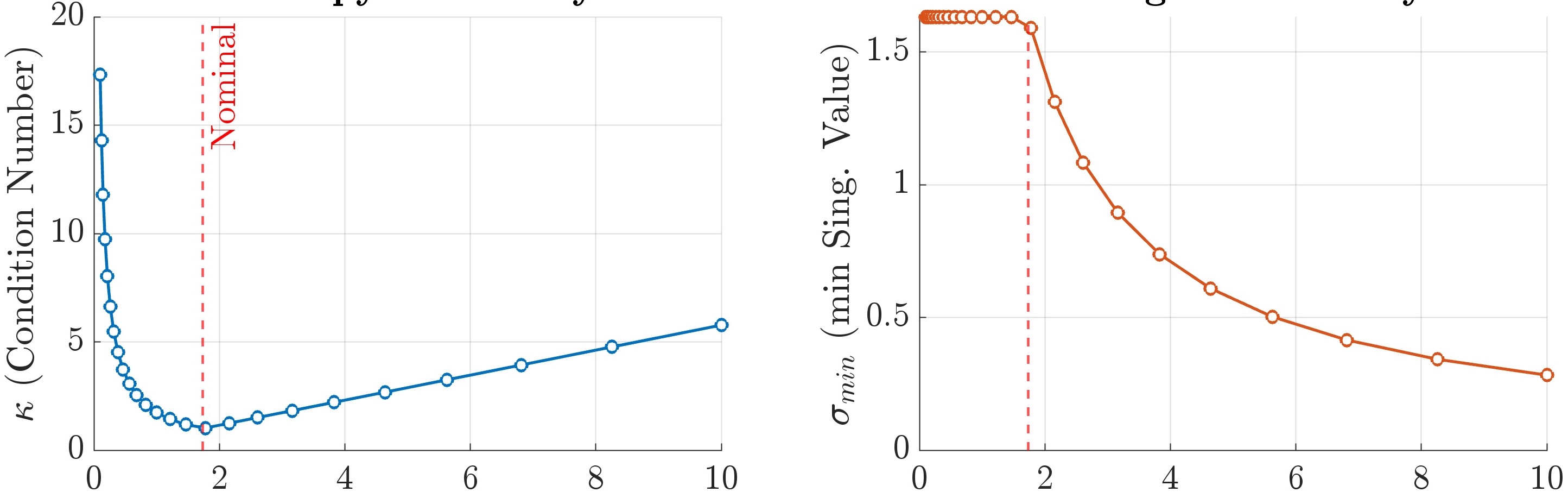}\\\smallskip
    \caption{\textbf{Sensitivity Analysis of Isotropy ($\kappa$) and Strength ($\sigma_{min}$) with respect to the characteristic length ratio $L_c / R$.}
    Results are shown for two representative chassis: a regular polygon ($N=12$, top) and a dodecahedron ($N=20$, bottom).
    \textbf{Left:} The Condition Number is minimized exactly at the unit ratio ($L_c = R$), indicating maximal balance between force and moment subspaces.
    \textbf{Right:} The Minimum Singular Value is maximized and stable for $L_c \le R$, but degrades for $L_c > R$ as the system becomes moment-limited. The vertical dashed line indicates the nominal design choice $L_c=R$.}
    \label{fig:sensitivity}
\end{figure}

Figure~\ref{fig:sensitivity} presents the resulting Condition Number ($\kappa$) and Minimum Singular Value ($\sigma_{min}$) as a function of the scaling ratio $L_c / R_{geom}$ for two exemplificative chassis. The results are similar for all the tested chassis and reveal two distinct physical regimes:
\begin{enumerate}
    \item \textbf{Isotropy Sensitivity (Left):} In all cases, the condition number $\kappa$ follows a convex trajectory with a global minimum at precisely $L_c \approx R_{geom}$. Values of $L_c \ll R$ exaggerate moment terms, while $L_c \gg R$ exaggerates force terms; the unit ratio represents the point of maximum balance between translational and rotational authority.
    \item \textbf{Strength Sensitivity (Right):} The minimum singular value $\sigma_{min}$ (representing the worst-case wrench capability) exhibits a saturation behavior. For $L_c \le R$, $\sigma_{min}$ remains constant, indicating a \textit{force-limited} regime where lever-arm scaling does not improve the bottleneck. For $L_c > R$, $\sigma_{min}$ decays rapidly, indicating a \textit{moment-limited} regime.
\end{enumerate}
The coincidence of the isotropy minimum and the strength saturation ``knee'' at $L_c = R_{geom}$ empirically confirms that the structural radius is the intrinsic metric for optimizing these systems.

\subsection{Geometric Remarks and Invariances}
A chassis $\mathcal{C}$ is formally defined by its vertex positions $\PosSet = \{\pvec_1, \dots, \pvec_N\}$. However, since the Log-Volume cost function utilizes the dimensionless grasp matrix $\bar{\Grasp}$ and is invariant to rigid body rotations, the solution landscape is invariant under the Similarity Group $Sim(3)$. Thus, a ``chassis'' in this context implicitly refers to the equivalence class of all vertex arrangements related by uniform scaling and rigid rotation.

\subsection{Computational Performance and Solution Quality}
To assess the practical viability of the Global Manifold Exhaustion strategy, we evaluate both the numerical efficiency and the geometric consistency of the convergence. The multi-start SQP framework was implemented in MATLAB and executed on an \emph{Apple M1 Max} workstation (10-core ARM64 architecture running at $\approx 3.22$ GHz) equipped with 64 GB of unified memory.

\subsubsection{Efficiency and Convergence Scaling}
The computational cost scales with the number of rotors $N$ and the geometric complexity of the chassis. Table~\ref{tab:comp_time} summarizes the average execution time for a single stochastic sample (averaged over $M=100$ runs) across various chassis.

The proposed Log-Volume potential ($J_{\text{vol}}$) demonstrates a significant performance advantage, achieving an average speedup of 2.3x compared to Condition Number minimization ($\kappa$). Average convergence times for a single run from the multi-start seed for $J_{\text{vol}}$ range from $8.3$ ms for simple hexagonal symmetries to roughly $76$ ms for the complex $N=20$ dodecahedron. This efficiency confirms that the Log-Volume gradient field remains sufficiently smooth and locally convex to ensure rapid convergence, even as the dimensionality of the search space scales as $2N$.

\begin{table}[b]
\centering
\caption{Computational Performance Comparison\\ (Average convergence time per seed)}
\resizebox{\columnwidth}{!}{%
\begin{tabular}{lcccc}
\toprule
Chassis Geometry & $N$ & Time ($J_{\text{vol}}$) & Time ($\kappa$) & Speedup \\
\midrule
Reg. Hexagon        & 6  & 8.3 ms  & 30.2 ms  & \textbf{3.7x} \\
Quasi-Hexagon     & 6  & 12.3 ms & 32.4 ms  & \textbf{2.6x} \\
Irreg. Pent. Pyr.   & 6  & 14.7 ms & 32.5 ms  & \textbf{2.2x} \\
Cube                & 8  & 13.9 ms & 43.8 ms  & \textbf{3.2x} \\
Icosahedron ($N_{12}$) & 12 & 46.7 ms & 72.4 ms & \textbf{1.6x} \\
Dodecahedron ($N_{20}$) & 20 & 76.4 ms & 140.1 ms & \textbf{1.8x} \\
\bottomrule
\end{tabular}%
}
\label{tab:comp_time}
\end{table}

\subsubsection{Ablation Study: Geometric Consistency vs. Condition Number}
While the efficiency gains are notable, the primary advantage of the Log-Volume potential $J_{\text{vol}}$ lies in the topological superiority of the resulting solution sets. We compared our approach directly against the standard minimization of the condition number $\kappa$.

The numerical results highlight two fundamental mathematical limitations of $\kappa$ that degrade optimization performance:
\begin{enumerate}
    \item \textbf{Gradient Instability:} The condition number is a non-smooth function; its gradient is undefined at configurations where singular values coincide (i.e., at the very isotropic optima it seeks to find). This non-differentiability creates discontinuities in the gradient field, causing standard gradient-based solvers (like SQP) to exhibit oscillation or convergence failure~\citep{lewis1996eigenvalue,nocedal2006optimization}.
    \item \textbf{Scale Invariance:} Because $\kappa$ is a ratio, it is invariant to the magnitude of the wrench capability. This induces a theoretically flat optimization landscape populated by numerous weak local minima, rather than a well-defined basin of attraction.
\end{enumerate}

\begin{figure}[t]
\centering
\footnotesize
\begin{tabularx}{\columnwidth}{@{\hspace{0pt}} >{\centering\arraybackslash}m{0.05\linewidth} @{\hspace{1pt}} | @{\hspace{0pt}} X @{\hspace{0pt}}}
\hline
\textbf{ID} & \multicolumn{1}{>{\centering\arraybackslash}X@{\hspace{0pt}}}{\textbf{Comparative Solution Topologies ($J_{\text{vol}}$ vs. $\kappa$)}} \\
\hline
\rowcolor[gray]{0.95} \multicolumn{2}{l}{\textit{Blue: Log-Volume (Coherent) \quad Orange: Condition Num. (Scattered)}} \\
\hline
\raisebox{-2.5em}{\rotatebox[origin=c]{90}{\textbf{CRPol10}}} & 
\includegraphics[width=0.95\linewidth, valign=m]{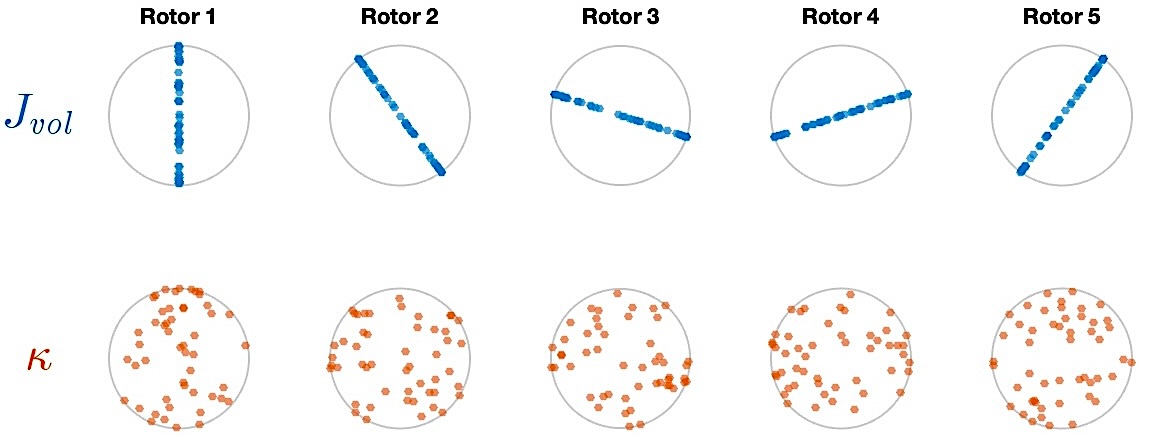} \\
\hline
\raisebox{-2.5em}{\rotatebox[origin=c]{90}{\textbf{CCub8}}} & 
\includegraphics[width=0.80\linewidth, valign=m]{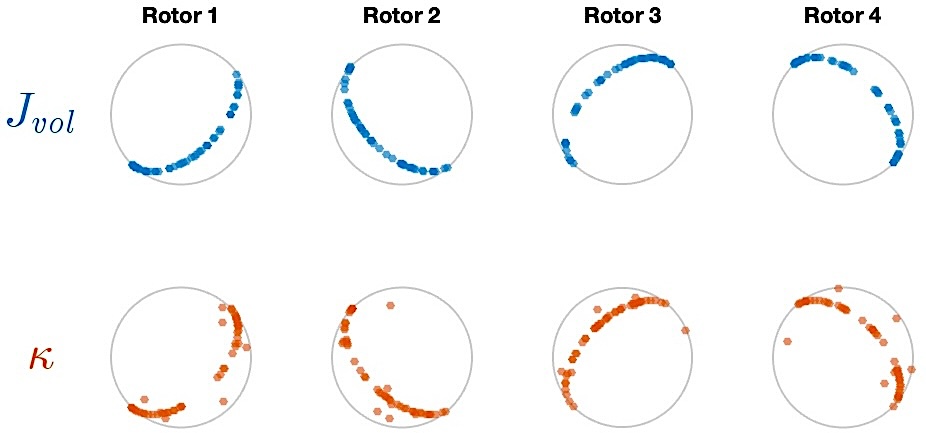} \\
\hline
\raisebox{-2.5em}{\rotatebox[origin=c]{90}{\textbf{CQRPol8}}} & 
\includegraphics[width=0.75\linewidth, valign=m]{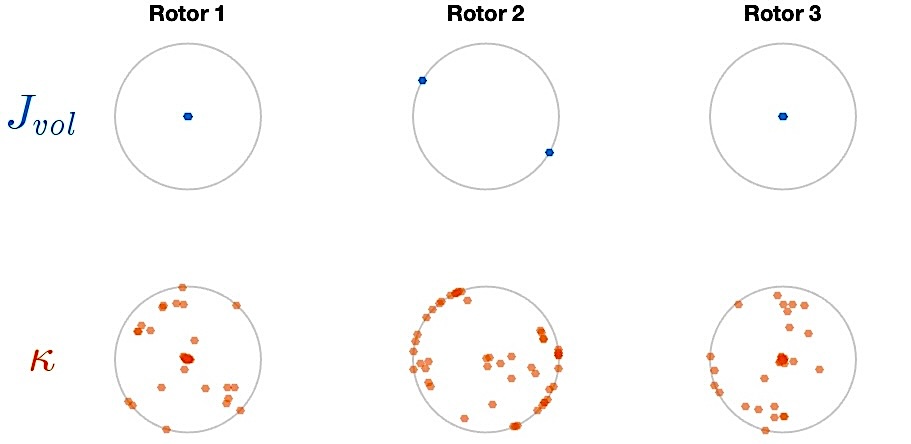} \\
\hline
\end{tabularx}
\caption{\textbf{Ablation of Objective Functions.} Side-by-side comparison of solution spaces (half of the discs shown;  the remaining half follow a similar trend). A reduced sample size of $M=50$ starting points is used here to clearly illustrate the distribution. The Log-Volume potential ($J_{\text{vol}}$, Blue) produces coherent solution sets (manifesting as tightly clustered points or smooth curve segments), whereas Condition Number optimization ($\kappa$, Orange) yields scattered, noisy minima despite identical initialization. This phenomenon is consistent across all tested chassis.}
\label{fig:ablation_comparison}
\end{figure}

Figure~\ref{fig:ablation_comparison} presents three representative cases selected from the broader dataset. Visual inspection reveals a stark contrast in solution topology driven by these mathematical properties. While $\kappa$-minimization (orange) results in unstructured, scattered point clouds due to gradient instability, the proposed Log-Volume method (blue) consistently converges to distinct, coherent sets of solutions. The topological structure of these sets is analyzed in detail in Sec.~\ref{sec:topology}. This structural consistency is critical for rigorous scientific analysis, providing a predictable set of valid solutions rather than the chaotic scatter resulting from an ill-posed cost function.

\section{Topological Analysis of the Optimal Orientation Space}
\label{sec:topology}

Having executed the global solving algorithm (Algorithm~\ref{algo:optimization}) for the chassis library listed in Table~\ref{tab:chassis_zoo} and illustrated in Fig.~\ref{fig:geometric_zoo}, we now analyze the resulting solution landscapes.
Rather than focusing solely on scalar optimality values--which are identical for all successful runs for a given geometry--we examine the spatial distribution of the converged solution set $\mathcal{S}^* = \{\mathcal{D}^*_1, \dots, \mathcal{D}^*_{M^*}\} \subset \Manifold$, where $M^* = |\mathcal{S}^*|$ denotes the number of distinct global optima identified.
For clarity, we distinguish between $\mathcal{S}^*$ (the set of all discovered solutions), $\mathcal{D}^*_m$ (the $m$-th distinct solution in that set), and $\mathcal{D}^*$ (a generic optimal configuration). 
Note that while $\mathcal{S}^*$ refers to a specific chassis $\mathcal{C}$, we omit this explicit dependency to simplify the notation.
This analysis reveals fundamental differences in the structure of the optimization landscape governed by the geometric properties of the chassis.

\subsection{Data Analysis and Phenomenological Classification}

To visualize the high-dimensional solution set, we display the optimal thrust direction lines of $\mathcal{S}^*$ on the disc dashboard representation introduced in Section~\ref{sec:geometric_characterization}. Based on the observed behaviors across the full chassis library, we classify the chassis into four distinct topological categories:

\begin{itemize}

    \item \textbf{Type I: Discrete Point Convergence.} For specific low-complexity, asymmetric chassis, the optimizer consistently converges to a very small set of discrete points (typically fewer than 10 unique configurations). Here, the solution manifold is effectively 0-dimensional, offering no continuous range of valid parameters for reconfiguration.

    \item \textbf{Type II: Unstructured Scattering.} For highly irregular chassis, the solutions in $\mathcal{S}^*$ appear generically scattered across the discs. There is no discernible pattern; the solutions form nebulous clouds covering large portions of the available area ($> 10$ distinguishable clusters or continuous noise). This indicates a flat, noisy optimization landscape with a vast number of local minima and no global coherence.

    \item \textbf{Type III: Hybrid Structure.} In intermediate cases, we observe a mixed behavior. For certain rotors, the solutions aggregate clearly around specific shapes (semi-elliptic-like arcs on the moduli discs), while for other rotors in the same chassis, the solutions remain scattered or form loose clusters ($>50$ isolated points). This suggests a partial, but incomplete, collapse of dimensionality.

    \item \textbf{Type IV: Coherent Manifold Collapse.} This is uniquely observed for Regular Polygons ($N \ge 6$), Platonic Solids, and a few quasi-regular chassis. For \emph{every} rotor in these systems, the $M^*$ independent solutions collapse perfectly onto precise, continuous curves within the 2D disc. Although the optimizer was free to explore the full $2N$-dimensional manifold $(\RPtwo)^N$, the global optima spontaneously confined themselves to a specific continuous $N$-dimensional submanifold.

\end{itemize}

\secondrev{The topological boundary between these phenomenological classes is governed by the degree of geometric asymmetry in the chassis. We quantify this using a normalized geometric variance metric, $\epsilon_{geom}$, defined as the maximum relative radial or angular deviation from a regular polygon. Based on our perturbation analysis, the transition from a ``quasi-regular'' condensing state (Type III) to a scattered ``irregular'' state (Types I and II) occurs at a critical geometric threshold of $\epsilon_{geom} \approx 0.15$. 

This geometric boundary in the physical chassis is directly linked to the algorithmic failure thresholds of the branch-extraction process. When the physical perturbation reaches $\epsilon_{geom} \ge 0.15$, the resulting optimal solutions scatter sufficiently in the phase space to break the continuity of the manifold. Statistically, this scattering causes the internal cluster cohesion to drop below the algorithm's detection limit (mean Silhouette score $S < 0.35$), and the spatial separation between branches to fall below the minimum resolution constraint ($\Delta\theta_{min} = 15^\circ$). At this point, the algorithm ceases to resolve continuous submanifolds on the Tangent Torus $\tangentTorus$ and registers the landscape as a set of discrete, disconnected local minima.}

This classification is empirically supported by the simulation data presented in Figures~\ref{fig:irregular_results} and~\ref{fig:regular_results}. In detail, we analyze these two distinct regimes below.

\subsubsection{Irregular Chassis: Discrete and Fragmented Solutions}

\begin{figure}[t]
\centering
\footnotesize
\begin{tabularx}{\columnwidth}{@{\hspace{0pt}} >{\centering\arraybackslash}m{0.18\linewidth} @{\hspace{1pt}} | @{\hspace{0pt}} X @{\hspace{0pt}}}
\hline
\textbf{ID} & \multicolumn{1}{>{\centering\arraybackslash}X@{\hspace{0pt}}}{\makecell{\textbf{Types I, II and III:} \\ \textbf{Global solution landscapes (Discrete/Fragmented)}}} \\
\hline
\rowcolor[gray]{0.95} \multicolumn{2}{l}{\textit{Type I: Sparse / Isolated Solutions (Few discrete points)}} \\
\hline
\textbf{CTriPr6} & \includegraphics[width=0.99\linewidth, valign=m]{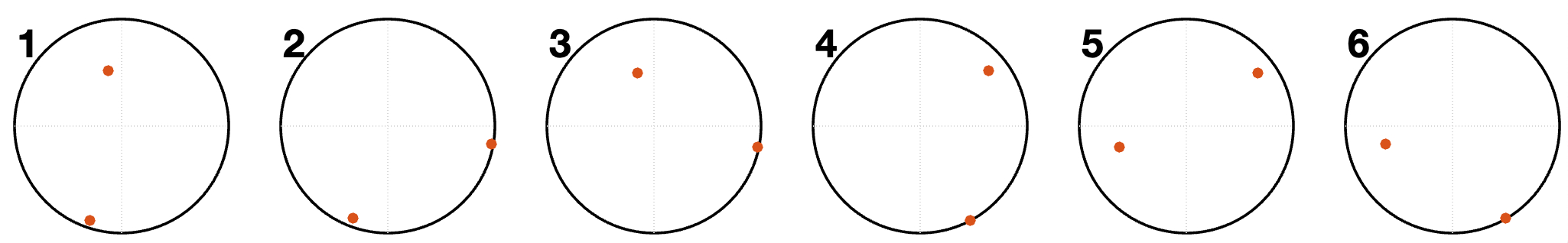} \\
\hline
\textbf{CQRPol6} & \includegraphics[width=0.99\linewidth, valign=m]{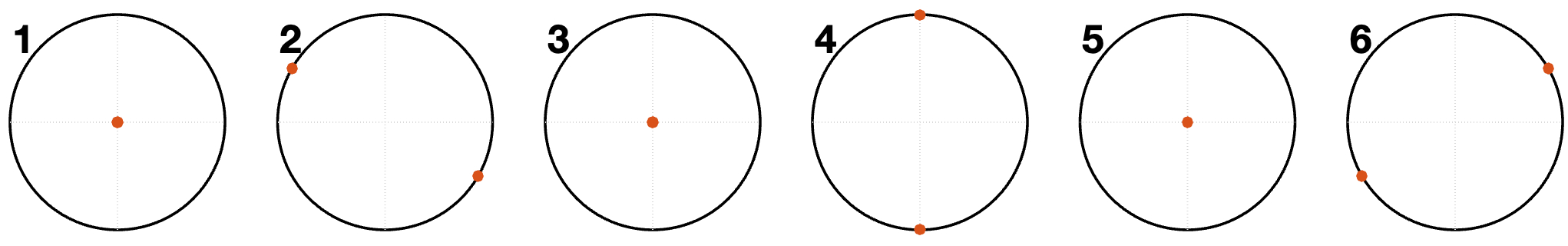} \\
\hline
\textbf{CQRPol7} & \includegraphics[width=0.99\linewidth, valign=m]{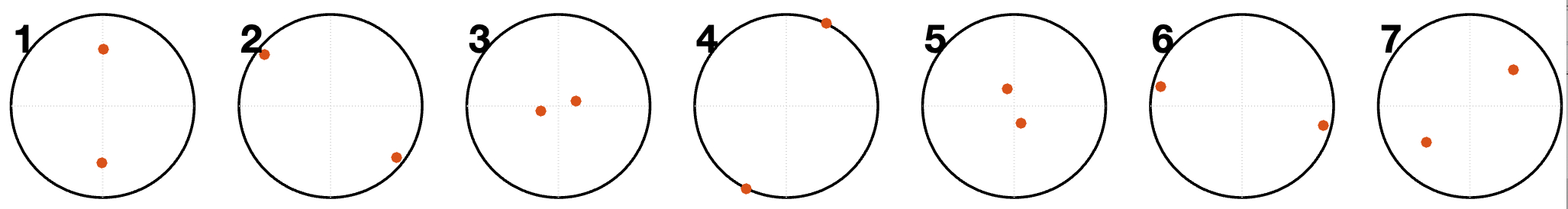} \\
\hline
\textbf{CQRPol9} & \includegraphics[width=0.99\linewidth, valign=m]{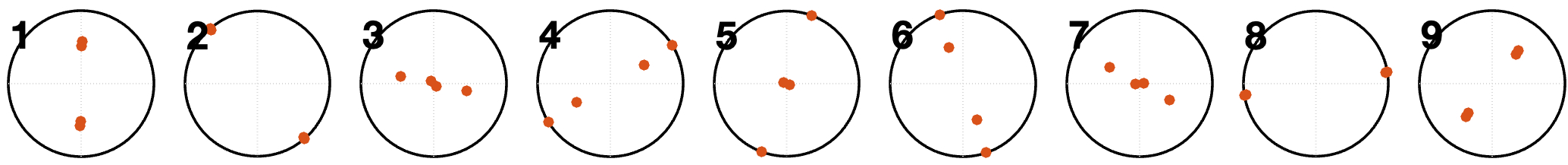} \\
\Xhline{2pt}
\rowcolor[gray]{0.95} \multicolumn{2}{l}{\textit{Type II: Clustered Solutions (Larger, denser solution islands)}} \\
\hline
\textbf{CPentBi7} & \includegraphics[width=0.99\linewidth, valign=m]{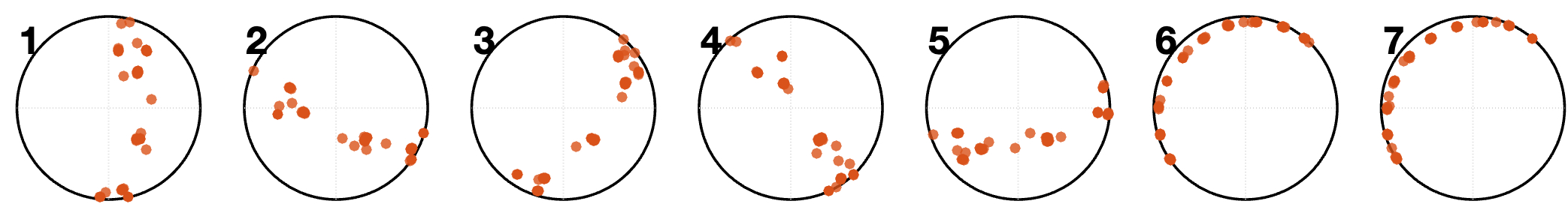} \\
\hline
\textbf{CTriCup9} & \includegraphics[width=0.99\linewidth, valign=m]{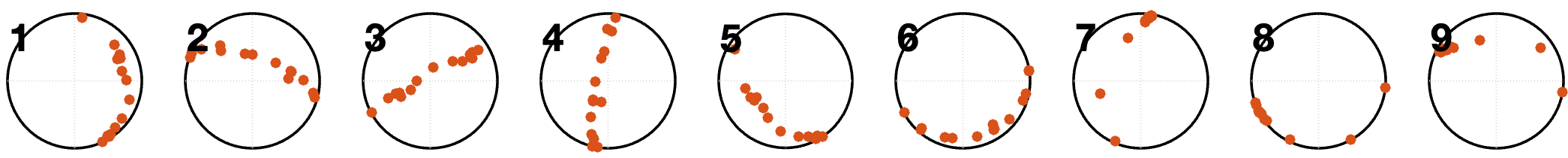} \\
\Xhline{2pt}
\rowcolor[gray]{0.95} \multicolumn{2}{l}{\textit{Type III: Condensing Solutions (Scattered but forming around $\tangentTorus$)}} \\
\hline
\textbf{CQRPol8} & \includegraphics[width=0.99\linewidth, valign=m]{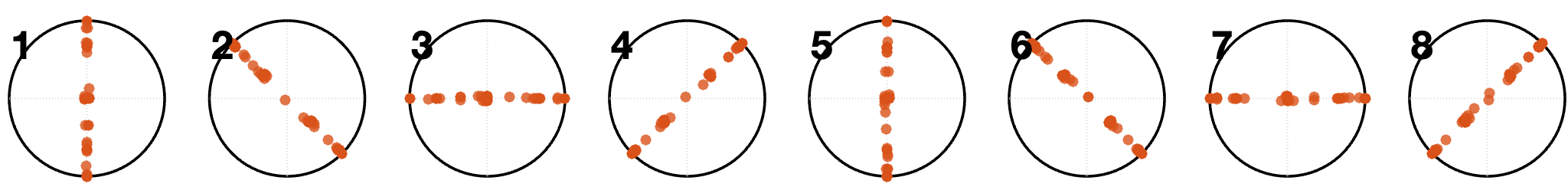} \\
\hline
\textbf{CSqAnti8} & \includegraphics[width=0.99\linewidth, valign=m]{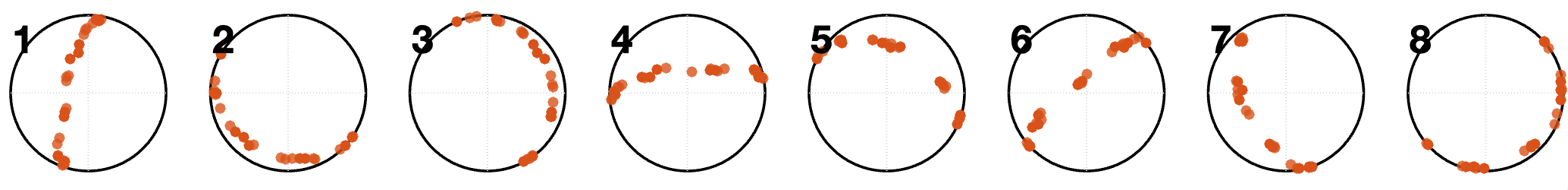} \\
\hline
\textbf{CQRPol10} & \includegraphics[width=0.99\linewidth, valign=m]{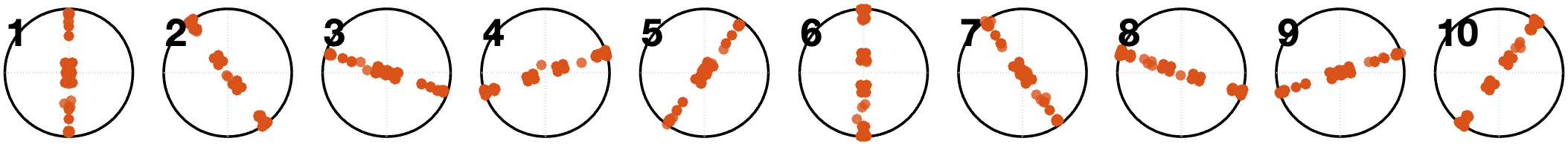} \\
\hline
\end{tabularx}
\caption{\textbf{Global solution landscapes for irregular and quasi-regular chassis.}
    These chassis exhibit discrete or fragmented solution sets that do not fully resolve onto the Tangent Torus $\tangentTorus$. The results are categorized by the density and distribution of their minima:
    \textbf{Type I: Sparse/Isolated (CTriPr6, CQRPol6, CQRPol7, CQRPol9):} The optimizer converges to a very small number of distinct, isolated points, indicating a rigid solution space with no continuous degrees of freedom.
    \textbf{Type II: Clustered (CPentBi7, CTriCup9):} Solutions appear as larger, denser ``islands'' or local clusters, suggesting a higher density of near-optimal configurations but no true connectivity.
    \textbf{Type III: Condensing (CQRPol8, CSqAnti8, CQRPol10):} The solutions remain scattered but show a clear tendency to ``condense'' or align towards the manifold structure of $\tangentTorus$, though they fail to form a continuous smooth curve, possibly due to the lack of sufficient geometric symmetry.}
\label{fig:irregular_results}
\end{figure}

Figure~\ref{fig:irregular_results} presents the results for a variety of irregular and quasi-regular chassis ranging from $N=6$ to $N=10$. The visualization reveals that none of these examples demonstrate a clean topological collapse onto a lower-dimensional manifold.

In extreme cases of asymmetry, the continuous solution space completely vanishes, leaving only a handful of rigid, isolated optimal configurations. This \textbf{Type I (Discrete Point Convergence)} behavior is starkly evident in the \emph{Triangular Prism ($N=6$)} and \emph{Quasi-Polygon ($N=7$)}, where the optimizer consistently lands on only $2$ distinct points per disc across all $M^*$ pruned runs. Similarly, the \emph{Quasi-Polygon ($N=9$)} converges to only $4$ isolated solutions. This collapse to a 0-dimensional solution set implies that the lack of geometric symmetry breaks the continuous degeneracy of the problem, locking the rotors into a finite set of fixed optimal orientations.

In other cases, the pruned $M^*$ multistart solutions are distributed in scattered clusters, often separated by large ``holes'' in the configuration space, characteristic of \textbf{Type II (Unstructured Scattering)}. This indicates an optimization landscape characterized by numerous discrete local minima rather than a continuous valley of optimal solutions.

Other chassis, such as the \emph{Pentagonal Bipyramid ($N=7$)} and \emph{Square Antiprism ($N=8$)}, exhibit an intermediate \textbf{Type III (Hybrid Structure)} behavior. While the solutions are not continuous, they show a tendency to align loosely along specific geometrical paths, though they remain fragmented with significant gaps. This suggests a ``pre-collapse'' state where the optimization landscape is influenced by an underlying near-symmetry but lacks the perfect regularity required for a continuous manifold to emerge.

\subsubsection{Regular Chassis: Continuous Manifold Collapse}

\begin{figure}[t]
\centering
\footnotesize
\begin{tabularx}{\columnwidth}{@{\hspace{0pt}} >{\centering\arraybackslash}m{0.18\linewidth} @{\hspace{1pt}} | @{\hspace{0pt}} X @{\hspace{0pt}}}
\hline
\textbf{ID} & \multicolumn{1}{>{\centering\arraybackslash}X@{\hspace{0pt}}}{\textbf{Type IV: Global solution landscapes (collapsed on $\tangentTorus$)}} \\
\hline
\rowcolor[gray]{0.95} \multicolumn{2}{l}{\textit{Near-Regular / Quasi-Regular (Scattered or Thickened Collapse)}} \\
\hline
\textbf{CQCub8} & \includegraphics[width=0.99\linewidth, valign=m]{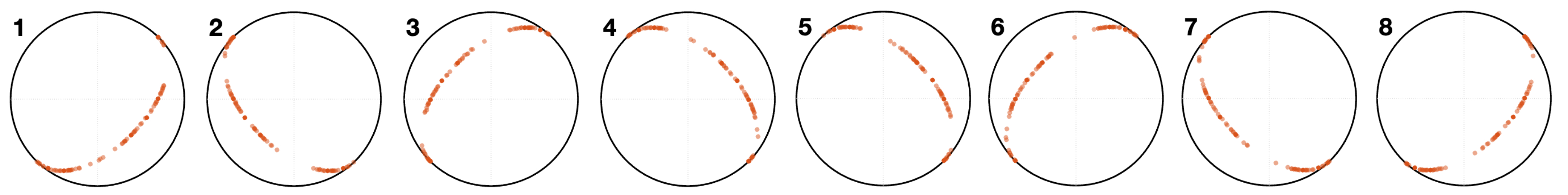} \\
\hline
\textbf{CCubOct12} & \includegraphics[width=0.99\linewidth, valign=m]{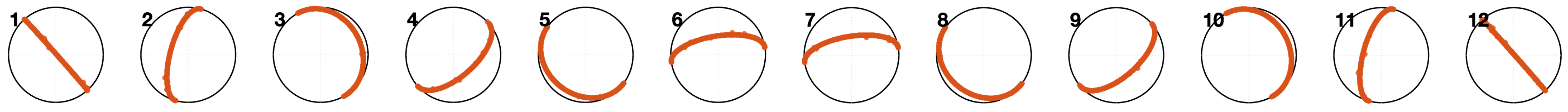} \\
\hline
\textbf{CHexPr12} & \includegraphics[width=0.99\linewidth, valign=m]{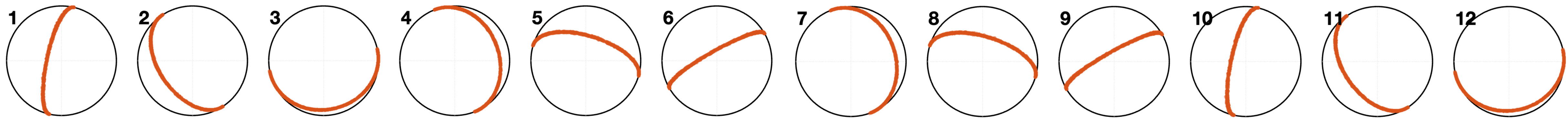} \\
\Xhline{2pt}
\rowcolor[gray]{0.95} \multicolumn{2}{l}{\textit{Regular Polygons (Perfect and Evenly Distributed Collapse)}} \\
\hline
\textbf{CRPol6} & \includegraphics[width=0.99\linewidth, valign=m]{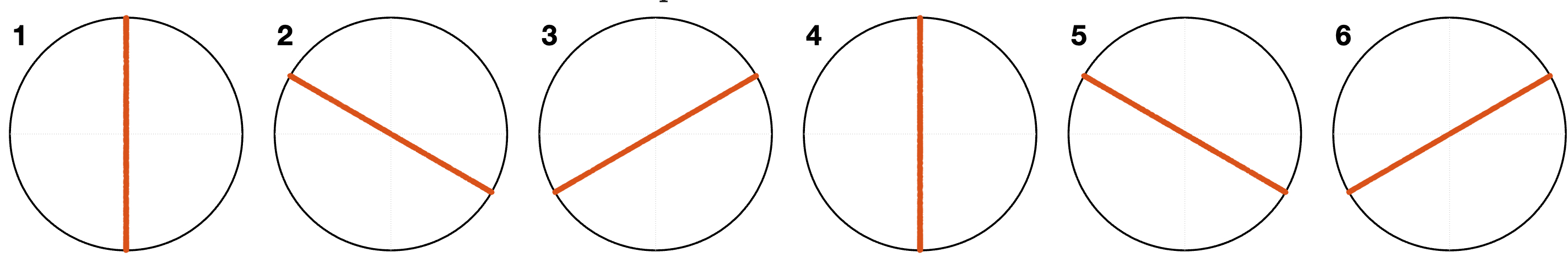} \\
\hline
\textbf{CRPol7} & \includegraphics[width=0.99\linewidth, valign=m]{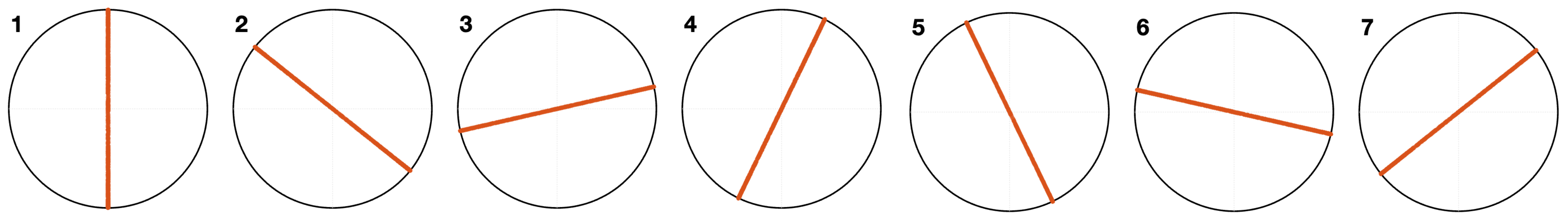} \\
\hline
\textbf{CRPol8} & \includegraphics[width=0.99\linewidth, valign=m]{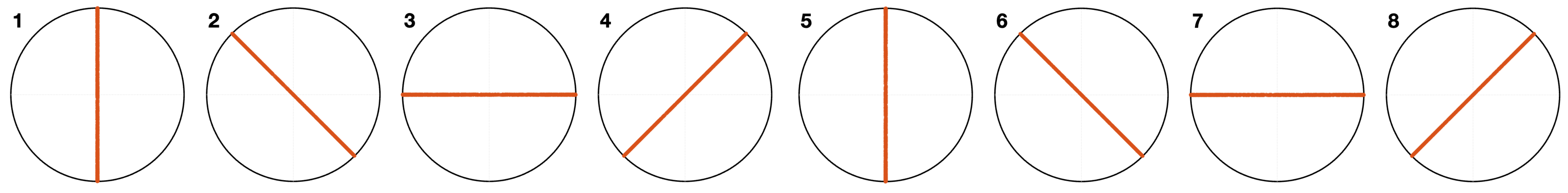} \\
\hline
\textbf{CRPol9} & \includegraphics[width=0.99\linewidth, valign=m]{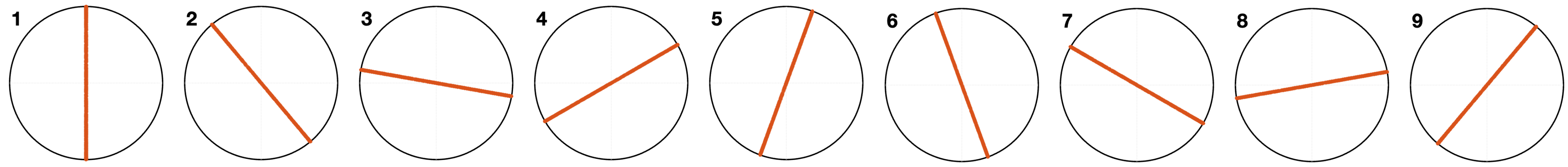} \\
\hline
\textbf{CRPol10} & \includegraphics[width=0.99\linewidth, valign=m]{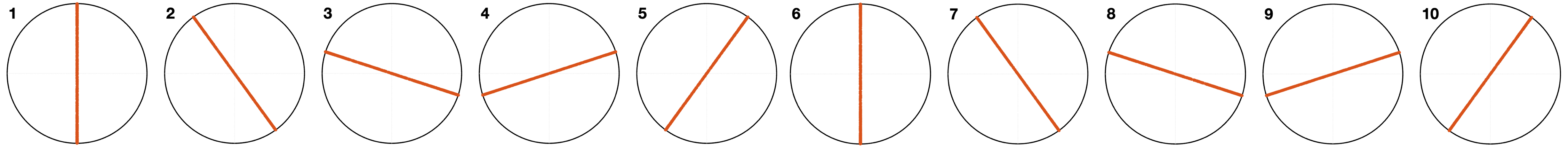} \\
\hline
\rowcolor[gray]{0.95} \multicolumn{2}{l}{\textit{Platonic Solids (Perfect and Evenly Distributed Collapse)}} \\
\hline
\textbf{COct6} & \includegraphics[width=0.99\linewidth, valign=m]{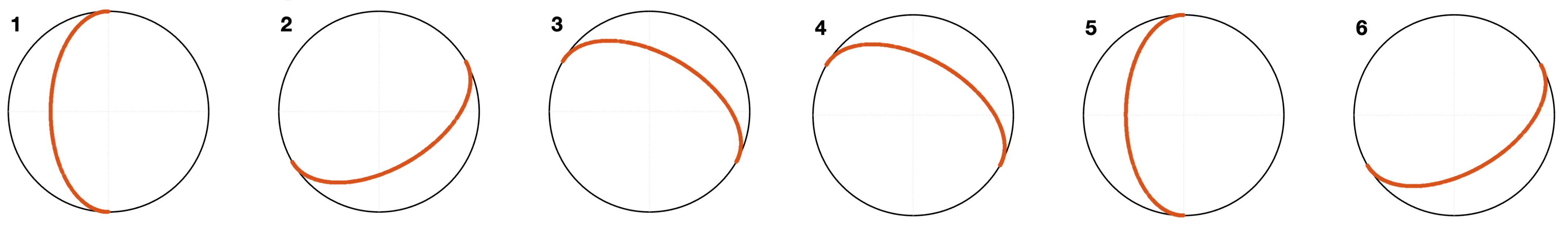} \\
\hline
\textbf{CCub8} & \includegraphics[width=0.99\linewidth, valign=m]{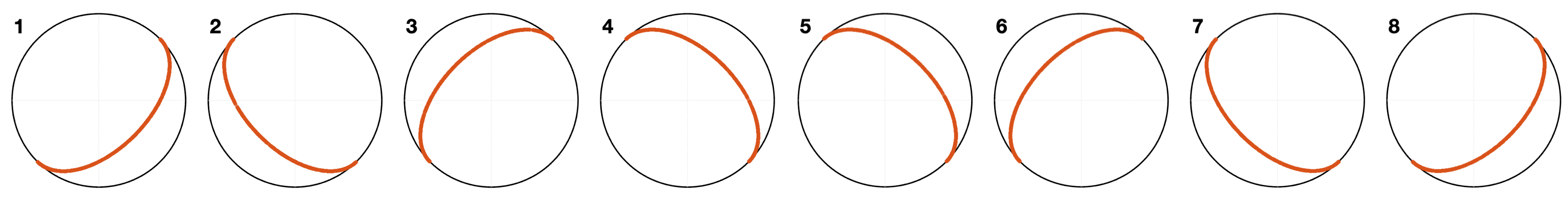} \\
\hline
\textbf{CIco12} & \includegraphics[width=0.99\linewidth, valign=m]{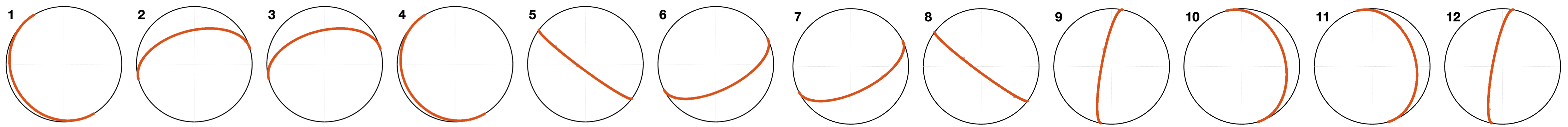} \\
\hline
\textbf{CDod20} & \includegraphics[width=0.99\linewidth, valign=m]{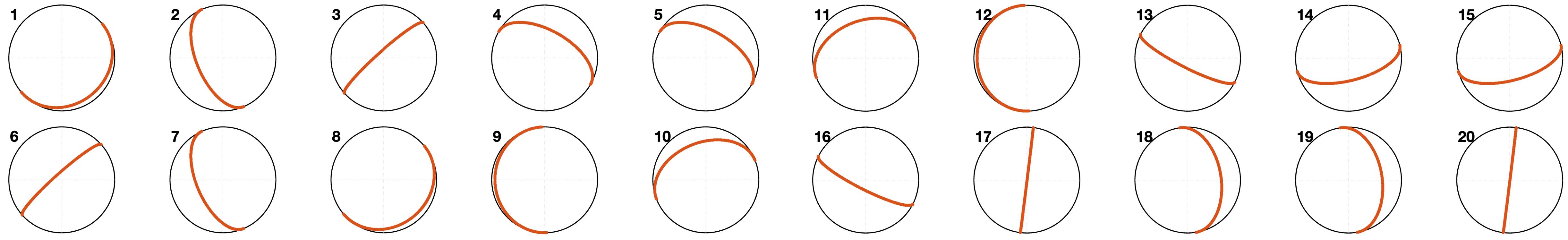} \\
\hline
\end{tabularx}
\caption{\textbf{Type IV: Global solution landscapes for symmetric chassis categorized by the nature of their manifold collapse.} The results demonstrate that for symmetric geometries, the optimizer converges toward the Tangent Torus $\tangentTorus$, but the quality of this collapse varies.
    \textbf{Near-regular geometries (CQCub8, CCubOct12, CHexPr12)} exhibit an imperfect collapse: for the Quasi-Cube (\textbf{CQCub8}), the solutions form thin lines but are non-uniformly distributed (clusters with empty gaps), while for the Cuboctahedron (\textbf{CCubOct12}) and Hexagonal Prism (\textbf{CHexPr12}), the solutions are evenly distributed but the manifold retains a residual thickness.
    In contrast, \textbf{Regular Polygons (CRPol6--10)} and \textbf{Platonic Solids (COct6, CCub8, CIco12, CDod20)} exhibit a ``perfect'' collapse, where the global minima form a vanishingly thin and perfectly evenly distributed manifold within $\tangentTorus$. Similar behavior is observed for regular polygons with $N\leq 20$. This suggests that high geometric regularity is a prerequisite for a smooth, continuous $N$-dimensional valley of optimal solutions contained in $\tangentTorus$, inside the $2N$-dimensional $\Manifold$ of the optimization variables.}
\label{fig:regular_results}
\end{figure}

In stark contrast to the irregular cases, every regular chassis (and certain highly symmetric quasi-regular ones) exhibits a phenomenon we term \textbf{Type IV: Coherent Manifold Collapse}. As shown in Figure~\ref{fig:regular_results}, the solution landscape for the $i$-th rotor does not scatter across the disc dashboard, but is strictly confined to a 1D submanifold of each of the $N$ discs in the dashboard, and therefore all the solutions in $\mathcal{S}^*$ are confined to an $N$-dimensional submanifold of $\Manifold$.

Visually, the orthographic projections of the solution points collapse perfectly onto smooth curves. Remarkably, the $M^*$ independent solution points are so evenly distributed along these paths that they appear indistinguishable from continuous lines, highlighting the stability and smoothness of the underlying solution manifold.

This behavior is universal across all tested regular polyhedra and polygons.
\begin{itemize}
    \item \textbf{Platonic Solids:} In the case of the \emph{Octahedron ($N=6$)}, \emph{Cube ($N=8$)}, \emph{Icosahedron ($N=12$)}, and \emph{Dodecahedron ($N=20$)}, the data points form clean, closed loops on the projection discs with no deviations into the unconstrained space.
    \item \textbf{Regular Polygons:} The phenomenon is equally distinct in the planar family. The \emph{Hexagon ($N=6$)} through the \emph{Decagon ($N=10$)} all display the same continuous distribution of solutions along strictly defined paths. This structure persists for all higher-order polygons tested ($10 < N \leq 20$), even those not explicitly displayed in the figure.
\end{itemize}

This collapse also appears in near-regular geometries (CQCub8, CCubOct12, CHexPr12), which, however, exhibit a very close but imperfect collapse when examined more closely (see caption of Figure~\ref{fig:regular_results}).

\subsection{Dimensionality Reduction to the Tangent Torus}
\label{subsec:dim_reduction}

We now focus exclusively on the Type IV datasets, which offer the most significant theoretical insight. The continuous curves observed in Figure~\ref{fig:regular_results} are not arbitrary artifacts of the optimization; they possess a precise geometric structure.

To rigorously quantify the observation that the projections of the solutions on the discs for regular chassis lie on 1D submanifolds, we developed the geometric reconstruction technique outlined in Algorithm~\ref{algo:ellipse_fitting}. 
The fundamental premise of this analysis is that a ``great circle'' (geodesic) on the unit sphere $\SphereTwo$, when projected onto the moduli disc representation of $\RPtwo$, manifests as a semi-ellipse (see Sec.~\ref{sec:visualization} and Fig.~\ref{fig:disc_concept}). Consequently, the shape and orientation of this ellipse are directly determined by the normal vector of the plane defining that geodesic.

\begin{algorithm}[t]
\caption{Geodesic Manifold Reconstruction via Semi-Elliptical Fitting}
\label{algo:ellipse_fitting}
\begin{algorithmic}[1]
\Require Set of converged global optima $\mathcal{S}^* = \{ \mathcal{D}^*_1, \dots, \mathcal{D}^*_{M^*} \}$ from Algo.~\ref{algo:optimization}.
\Ensure Manifold parameters $\Phi = \{(\psi_i, \eta_i)\}_{i=1}^N$.

\For{rotor $i = 1$ to $N$}
    \State \textit{Step 1: Data Extraction \& Projection}
    \State Let $\mathcal{X}_i = \{ \mathbf{x}_{i,1}, \dots, \mathbf{x}_{i,{M^*}} \}$ be the set of 2D coordinates for rotor $i$, obtained by projecting each solution $\mathcal{D}^*_m \in \mathcal{S}^*$ onto the moduli space (disc $D^2$).

    \State \textit{Step 2: Initialization (PCA)}
    \State Compute covariance $\Sigma_i = \text{Cov}(\mathcal{X}_i)$ and eigenvectors $\mathbf{v}_1, \mathbf{v}_2$.
    \State Init rotation $\hat{\psi} \leftarrow \operatorname{atan2}(v_{1,y}, v_{1,x})$ and minor axis $\hat{b} \leftarrow 0.5$.

    \State \textit{Step 3: Non-linear Least Squares Optimization}
    \State Define semi-ellipse $\mathcal{E}(\psi, b)$ and distance metric $d(\mathbf{p}, \mathcal{E})$.
    \State Solve:
    \begin{equation*}
        (\psi^*, b^*) \leftarrow \arg\min_{\psi, b} \sum_{m=1}^{{M^*}} \left( d(\mathbf{x}_{i,m}, \mathcal{E}(\psi, b)) \right)^2
    \end{equation*}

    \State \textit{Step 4: Orientation Correction}
    \State Define normal $\mathbf{n} = [\sin(\psi^*), -\cos(\psi^*)]^\top$ and centroid $\bar{\mathbf{x}}$.
    \If{$\bar{\mathbf{x}} \cdot \mathbf{n} \le 0$}
        \State $\psi^* \leftarrow \psi^* + \pi$. \Comment{Flip to match $\mathbb{R}P^2$ sector}
    \EndIf

    \State \textit{Step 5: Parameter Recovery}
    \State Store $\psi_i \leftarrow \psi^*$ and $\eta_i \leftarrow \arccos(b^*)$ in $\Phi$.
\EndFor

\State \textbf{Return} $\Phi$.
\end{algorithmic}
\end{algorithm}

The algorithm accepts as input the raw ensemble of ${M^*}$ converged global optima $\mathcal{S}^* = \{ \mathcal{D}^*_1, \dots, \mathcal{D}^*_{M^*} \}$. Each solution $\mathcal{D}^*_m$ is projected onto the 2D moduli space, yielding a set of planar coordinates $\mathbf{x}_m = (\mathbf{x}_{1,m},\ldots, \mathbf{x}_{N,m}) \in (D^2)^N$, where $\mathbf{x}_{i,m}$ represents the state of the $i$-th rotor in the $m$-th solution. 

The algorithm's objective is to extract the structured manifold parameters $\Phi = \{(\psi_i, \eta_i)\}_{i=1}^N$. Here, $\psi_i$ and $\eta_i$ represent, respectively, the \emph{azimuth} (rotation) and the \emph{elevation} (inclination) of the normal vector defining the great circle whose projection matches the semi-ellipse observed in the data. These $2N$ parameters mathematically define the specific $N$-dimensional submanifold that best approximates the empirical solution set $\mathcal{S}^*$.

The reconstruction proceeds by treating the projected solution set for each rotor $i$ as a noisy point cloud in $D^2$. First, we employ Principal Component Analysis (PCA) to determine the principal axis of the distribution, providing a robust initialization for the rotation angle. Subsequently, a non-linear least-squares optimization refines the parameters of a semi-elliptical model to minimize the Euclidean distance to the projected data points. 
Crucially, the algorithm recovers the physical elevation angle $\eta_i$ of the constraint plane via the optimized minor axis parameter $b^*$, through the relation $\eta_i = \arccos(b^*)$. The final output $\Phi$ thus provides the precise azimuthal orientation $\psi_i$ and elevation $\eta_i$ required to reconstruct the $N$-dimensional submanifold.

\begin{figure}[t]
    \centering
    \framebox{\includegraphics[width=0.75\linewidth]{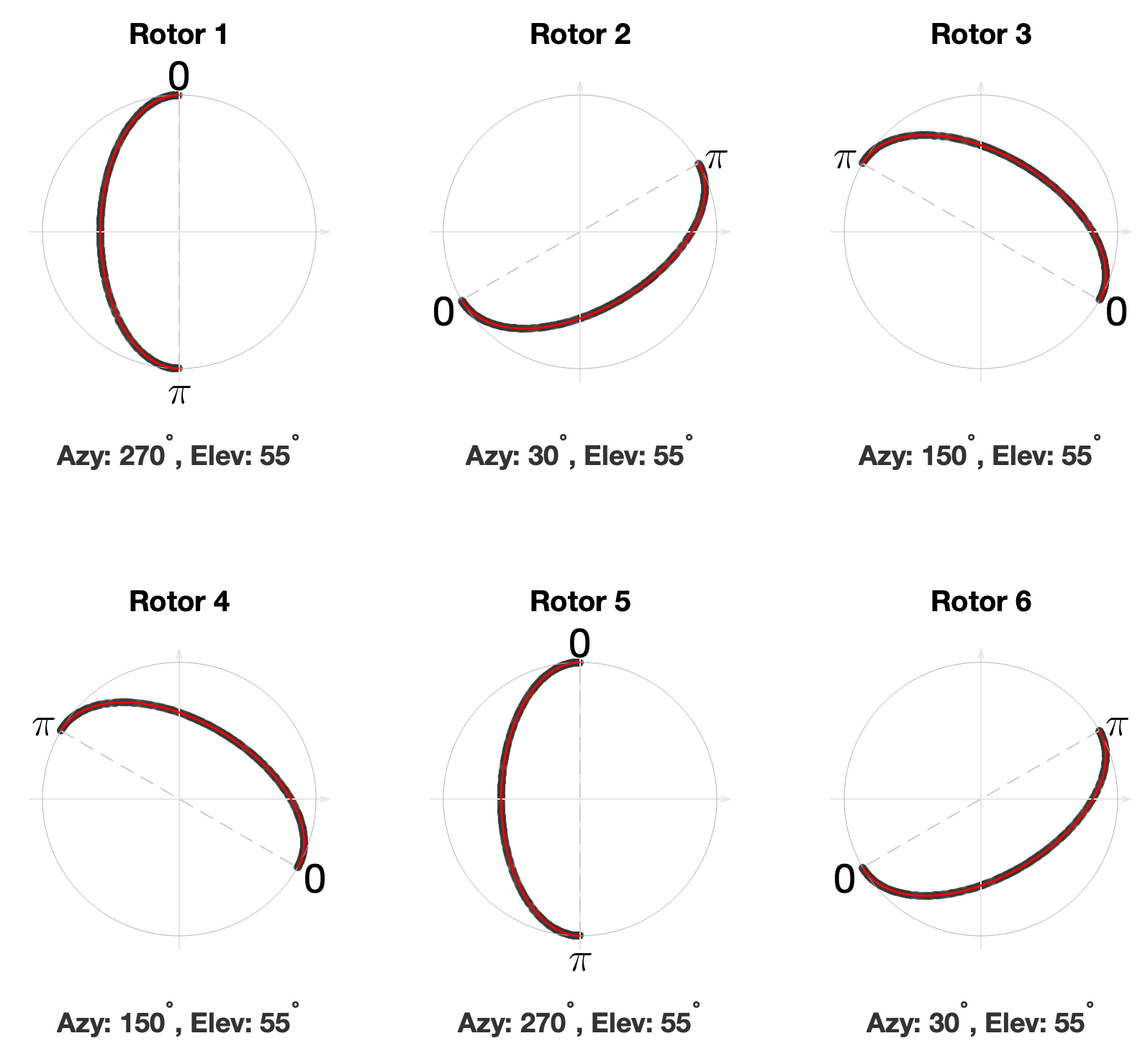}}
    \caption{Representation of the solution landscape data points (${M^*}$ points) in $\Manifold$ for the chassis: \textbf{Octahedron ($N=6$)}. The gray dots represent the two coordinates of these points on each disc $\RPtwo_i$. The data is perfectly fitted by the red semi-ellipses, whose parameters (rotation and elevation) are shown in each subplot. The parameters recovered by the fit reveal that the combination of all semi-elliptical curves corresponds exactly to the $\tangentTorus$ of the Octahedron ($N=6$), see Fig.~\ref{fig:RP2N} bottom right.}
    \label{fig:fit_octahedron}
\end{figure}

As visualized for exemplification in Fig.~\ref{fig:fit_octahedron} for the octahedron chassis (and in the  Figures~\ref{fig:fit_cube}, \ref{fig:fit_icosahedron}, \ref{fig:fit_dodecahedron}, \ref{fig:fit_poly6}, \ref{fig:fit_poly7}, \ref{fig:fit_poly8}, \ref{fig:fit_poly9}, \ref{fig:fit_poly10} of Appendix~\ref{app:additional_figs}), for all the other chassis the fit is numerically exact for all Type IV chassis. The analysis yields two fundamental discoveries:
\begin{enumerate}
    \item \textbf{Perfect Elliptical Fit:} The residual error of the fit is negligible, confirming that the solutions lie on ideal algebraic curves.
    \item \textbf{Geometric Identification with the Tangent Torus:} The recovered parameters of the $N$ fitted ellipses (specifically the rotation $\psi_i$ and elevation $\eta_i$) correspond precisely to the definition of the planes perpendicular to the structural radius vector ($\mathbf{d}_i \perp \mathbf{p}_i$).
\end{enumerate}

This second finding is critical. Geometrically, it implies that the identified submanifold is indeed the tangent torus $\tangentTorus$ defined in Sec.~\ref{subsec:proj_design_manifold}:
\begin{equation}
    \tangentTorus \cong \RPone \times \RPone \times \dots \times \RPone \quad (N \text{ times}).
\end{equation}
This represents a \emph{reduction of the  dimensionality of the landscape of solutions by half} ($2N \to N$). The problem of finding an isotropic design for regular chassis is   isomorphic to identifying specific configurations on this $N$-dimensional torus.

\section{Dimensionality Reduction to a 1D Curve and the $N-5$ Scaling Law}
\label{sec:scaling_law}

In Section~\ref{sec:topology}, we established that for regular chassis, the solution space collapses from the generic $2N$-dimensional manifold $\Manifold=(\RPtwo)^N$ to the $N$-dimensional Tangent Torus $\tangentTorus$. While this represents a significant reduction in dimensionality, the exact distribution of solutions \emph{within} this torus remains to be characterized. Specifically, we investigate whether the optimizer explores the full volume of $\tangentTorus$ or if a secondary collapse to an even lower-dimensional submanifold occurs.

\subsection{Intrinsic Parameterization of the Tangent Torus}
\label{subsec:tangent_basis_def}

To analyze the internal structure of the solution set, we must transition from the extrinsic 2D coordinates of the dashboard discs to an intrinsic parameterization of the tangent torus.
We define the Tangent Torus $\tangentTorus$ strictly via the chassis geometry. For each vertex $\mathcal{P}_i$ with position vector $\mathbf{p}_i$, we construct a local orthonormal tangent basis $\mathcal{F}_i = \{ \mathbf{u}_i, \mathbf{v}_i \}$ such that:
\begin{enumerate}
    \item The normal vector is radial: $\mathbf{n}_i = \mathbf{p}_i / \|\mathbf{p}_i\|$.
    \item The vector $\mathbf{u}_i$ denotes the ``horizontal'' tangent, defined by the cross product of the global vertical axis $\mathbf{e}_z$ and the normal: $\mathbf{u}_i = (\mathbf{e}_z \times \mathbf{n}_i) / \|\mathbf{e}_z \times \mathbf{n}_i\|$.
    \item The vector $\mathbf{v}_i$ completes the orthonormal set: $\mathbf{v}_i = \mathbf{n}_i \times \mathbf{u}_i$.
\end{enumerate}
This construction creates a consistent local frame where $\mathbf{u}_i$ lies parallel to the global $xy$-plane for all rotors.
Since the thrust vector $\mathbf{d}_i$ resides in the projective manifold $\RPtwo$, its restriction to the plane spanned by $\mathcal{F}_i$ is equivalent to the real projective line $\RPone$. Topologically, this space is a circle uniquely parameterized by a single projective phase angle $\theta_i \in [0, \pi)$. Any feasible rotor orientation on this manifold is expressed as:
\begin{equation}
    \mathbf{d}_i(\theta_i) = \cos(\theta_i)\mathbf{u}_i + \sin(\theta_i)\mathbf{v}_i, \quad \theta_i \in [0, \pi).
    \label{eq:tangent_param}
\end{equation}
The solution space is thus reduced from the $2N$-dimensional manifold $\Manifold$ (represented by 2D points on the dashboard discs) to the $N$-dimensional torus $\tangentTorus$, intrinsically represented by the product of $N$ projective lines: $\bm{\theta} = [\theta_1, \dots, \theta_N]^\top \in (\RPone)^N \cong \tangentTorus$.

\subsection{Exact Extraction of Phase Coordinates}
\label{subsec:tangent_manifold_mapping}

\begin{algorithm}[t]
\caption{Intrinsic Phase Extraction}
\label{algo:phase_extraction}
\begin{algorithmic}[1]
\Require Single Solution $\mathcal{D}^* = \{\mathbf{d}^*_1, \dots, \mathbf{d}^*_N\}$, Chassis Vertices $\mathcal{P}$.
\Ensure Intrinsic Phase Vector $\bm{\theta} \in [0, \pi)^N$.

\For{$i = 1$ to $N$}
    \State \textit{Step 1: Construct Tangent Basis}
    \State $\mathbf{n}_i \gets \mathbf{p}_i / \|\mathbf{p}_i\|$ \Comment{Radial normal}
    \State $\mathbf{u}_i \gets (\mathbf{e}_z \times \mathbf{n}_i) / \|\dots\|$ \Comment{Azimuthal tangent}
    \State $\mathbf{v}_i \gets \mathbf{n}_i \times \mathbf{u}_i$ \Comment{Polar tangent}
    
    \State \textit{Step 2: Project Solution}
    \State $x_{local} \gets \mathbf{d}^*_i \cdot \mathbf{u}_i$
    \State $y_{local} \gets \mathbf{d}^*_i \cdot \mathbf{v}_i$
    
    \State \textit{Step 3: Compute Intrinsic Angle}
    \State $\theta_i \gets \operatorname{atan2}(y_{local}, x_{local})$
    
    \State \textit{Step 4: Projective Mapping}
    \State \Comment{Enforce $\mathbb{R}P^1$ symmetry (periodicity $\pi$)}
    \If{$\theta_i < 0$} $\theta_i \gets \theta_i + \pi$ \EndIf
\EndFor

\State \textbf{Return} $\bm{\theta} = [\theta_1, \dots, \theta_N]^\top$.
\end{algorithmic}
\end{algorithm}

To determine if a deeper coordination pattern exists in the solution landscape--i.e., if the solutions $\mathcal{S}^*$ occupy a lower-dimensional subspace of $\tangentTorus$--we project each converged solution vector $\mathbf{d}^*_{i}$ of a configuration $\mathcal{D}^*$ onto the aforementioned basis $\mathcal{F}_i$ and extract its corresponding intrinsic phase $\theta_{i}$ by inverting the map in~\eqref{eq:tangent_param}.
The projection of the raw solution vector onto this basis acts as a geometric filter: it isolates the relevant tangent components while strictly rejecting any numerical noise in the radial direction $\mathbf{n}_i$.
The extraction procedure for a single solution set is detailed in Algorithm~\ref{algo:phase_extraction}.

\begin{figure*}[t]
\centering
\footnotesize
\begin{tabularx}{\textwidth}{@{\hspace{0pt}} >{\centering\arraybackslash}m{0.08\linewidth} @{\hspace{1pt}} | @{\hspace{0pt}} X @{\hspace{0pt}}}
\hline
\textbf{ID} & \multicolumn{1}{c}{\textbf{Pairwise Correlation Scatters}} \\ 
            & \multicolumn{1}{c}{($\theta_1$ vs. $\theta_i$ for $i=1 \dots N$)} \\
\hline

\rowcolor[gray]{0.95} \multicolumn{2}{l}{\textit{Regular Polygons}} \\
\hline
\textbf{CRPol6} & \includegraphics[width=0.99\linewidth, valign=m]{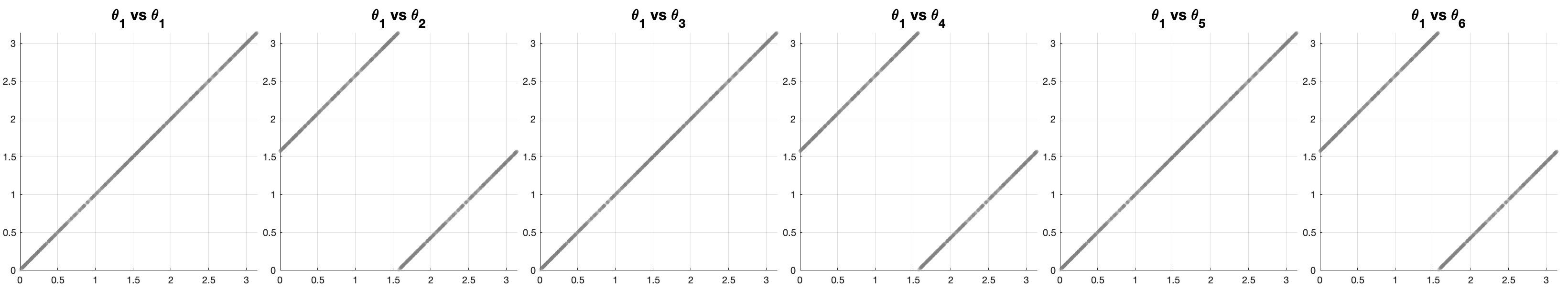} \\
\hline
\textbf{CRPol7} & \includegraphics[width=0.99\linewidth, valign=m]{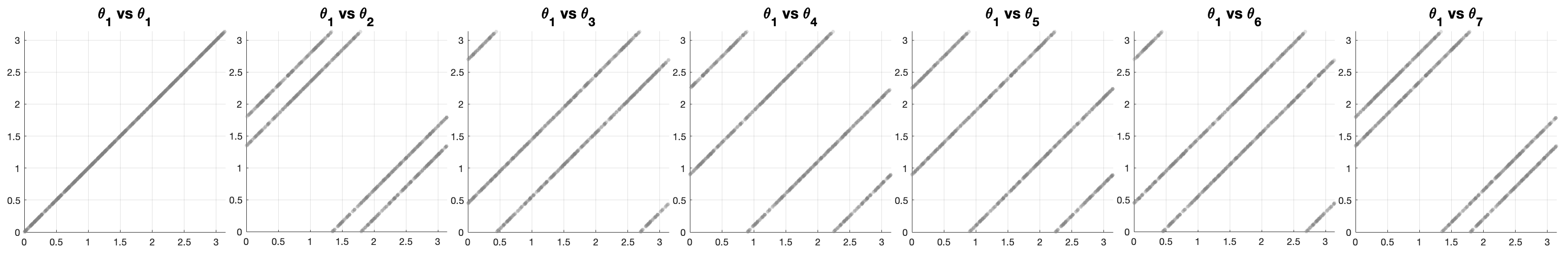} \\
\hline
\textbf{CRPol8} & \includegraphics[width=0.99\linewidth, valign=m]{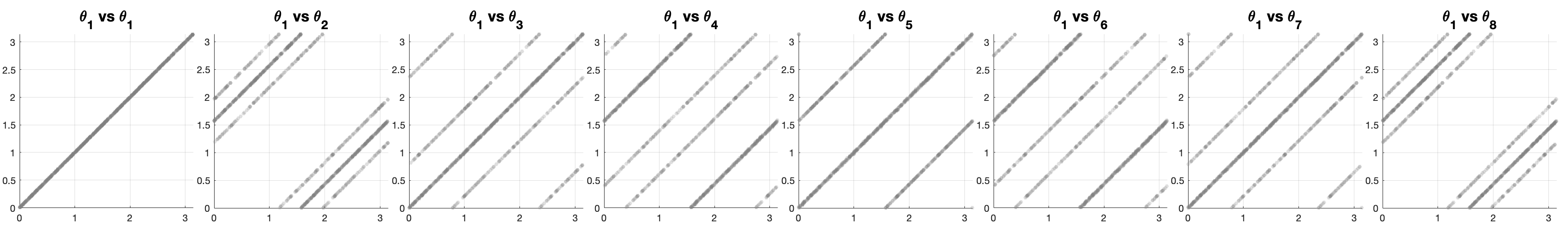} \\
\hline
\textbf{CRPol9} & \includegraphics[width=0.99\linewidth, valign=m]{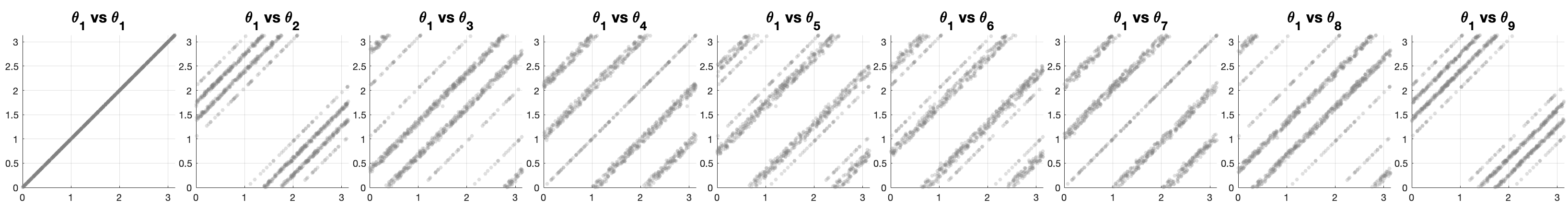} \\
\hline
\textbf{CRPol10} & \includegraphics[width=0.99\linewidth, valign=m]{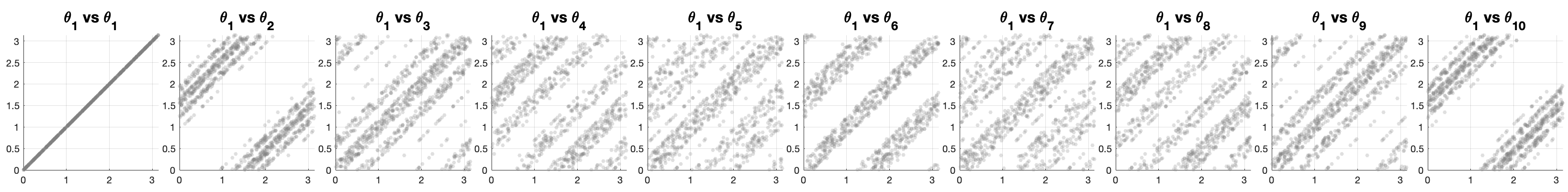} \\
\hline
\textbf{CRPol11} & \includegraphics[width=0.99\linewidth, valign=m]{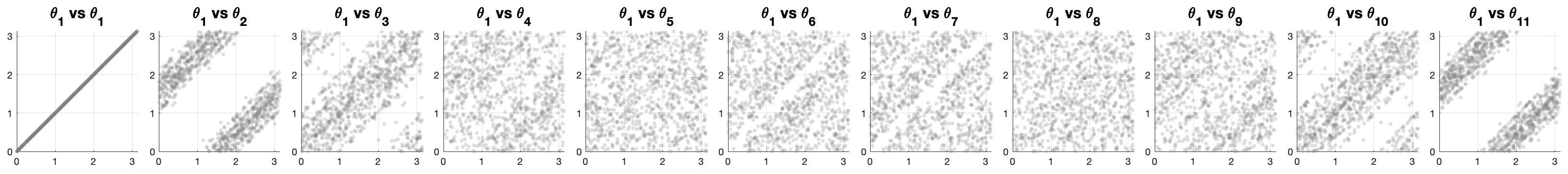} \\

\Xhline{2pt}
\rowcolor[gray]{0.95} \multicolumn{2}{l}{\textit{Platonic Solids}} \\
\hline
\textbf{COct6} & \includegraphics[width=0.99\linewidth, valign=m]{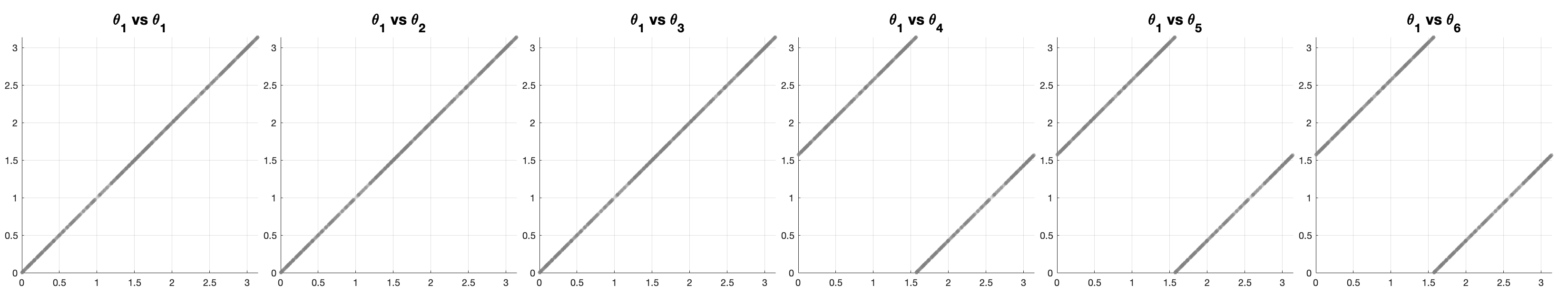} \\
\hline
\textbf{CCub8} & \includegraphics[width=0.99\linewidth, valign=m]{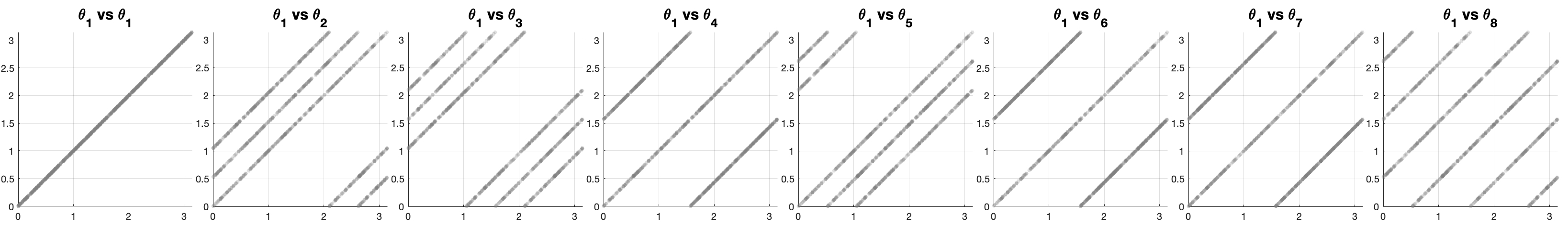} \\
\hline
\textbf{CIco12} & \includegraphics[width=0.99\linewidth, valign=m]{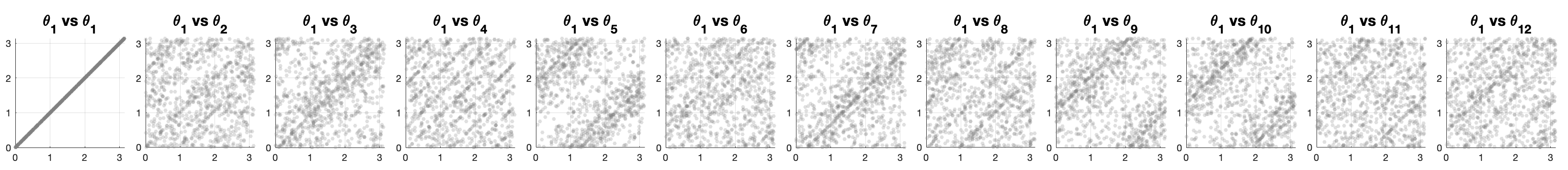} \\
\hline
\textbf{CDod20} & \includegraphics[width=0.99\linewidth, valign=m]{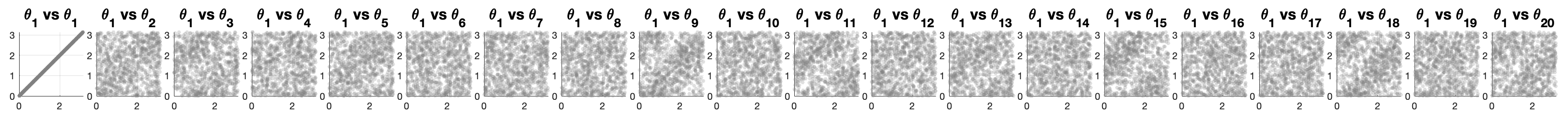} \\
\hline
\end{tabularx}

\caption{\textbf{Linear coordination patterns in the intrinsic angular space.}
    Each row displays the phase correlations for a specific chassis. The subplots map the converged solutions onto the coordinate plane defined by the master angle $\theta_1$ (x-axis) and the $i$-th rotor angle $\theta_i$ (y-axis).
    \textbf{Regular Polygons (Top):} As $N$ increases, the solutions collapse onto linear trajectories with slopes of $+ 1$, indicating strict synchronous coupling. For $N>6$, multiple parallel lines appear, revealing the existence of distinct discrete isomers.
    \textbf{Platonic Solids (Bottom):} These highly symmetric chassis similarly exhibit multiple stacked parallel lines, confirming the presence of distinct symmetric solution loops separated by fixed phase shifts. Note the increasing ``blur'' in chassis with higher $N$.}
\label{fig:scatter_table}
\end{figure*}

By applying Algorithm~\ref{algo:phase_extraction} to every solution $\mathcal{D}^*$ in $\mathcal{S}^*$, we transform the point cloud $\mathcal{S}^*$ into a set of phase vectors $\{\bm{\theta}_m\}_{m=1}^{M^*}$.
To visualize $\mathcal{S}^*$ in these new coordinates and uncover underlying patterns, we employ a pairwise scatter plot technique, the results of which are shown in Figure~\ref{fig:scatter_table}. Each row of Figure~\ref{fig:scatter_table} corresponds to a specific regular chassis (Polygons $N=6, \dots, 11$ and Platonic Solids). For each chassis, we display a sequence of $N$ subplots (the ``squares'') that reveal the global coordination of the system. In each subplot $i$, we plot the intrinsic angle $\theta_{i,m}$ of the $i$-th rotor (y-axis) against the angle of the reference rotor $\theta_{1,m}$ (x-axis) across the full ensemble of solutions. This visualization effectively projects the $N$-dimensional solution manifold onto a series of 2D planes. Note that the choice of $\theta_1$ as the reference is arbitrary; equivalent linear structures emerge regardless of which rotor phase (e.g., $\theta_2, \theta_3, \dots$) is selected as the independent variable.

A striking structural order emerges from the data: rather than filling the space randomly, the solutions collapse onto distinct 1D structures composed of sharp linear segments.
Four geometric features are immediately apparent in these projections, revealing the physical nature of the coordination:

\begin{enumerate}
    \item \textbf{Unitary Slopes:} The trajectories form straight lines with unitary slopes (diagonal lines), which implies $\theta_i \approx \theta_1 + C$. This indicates a \textbf{1:1 synchronous locking}, where rotor $i$ rotates in the same direction (in its local frame) as the master rotor while maintaining a fixed phase shift.
    
    \item \textbf{Parallelism and Isomers:} For most chassis (specifically $N > 6$), we observe multiple ``stacked'' parallel lines for a single range of $\theta_1$, separated by discrete intervals. This indicates that while the continuous coordination law (the slope) remains invariant, the solution landscape possesses distinct discrete \textit{isomers}. These are symmetric permutations of the solution where the relative phase relationships are shifted by fixed quanta. Empirical observation suggests the number of concurrent isomers scales as $N-5$.

    \item \textbf{Closedness:} The linear segments are not open-ended; the endpoints correspond exactly to the starting points given the periodic topology of the torus. This results in perfectly closed curves in the 2-torus projections $\mathbb{T}^2= S^1\times S^1$ (corresponding to pairs $(\theta_1, \theta_i)$). This closure implies that the full solution set in $\tangentTorus$ also forms closed loops, indicating a cyclical behavior where the master parameter $\theta_1$ traverses the full interval $[0, \pi)$.

    \item \textbf{Blurring:} 
    \secondrev{As $N$ increases ($N \ge 8$), the sharpness of the lines degrades, and for $N \ge 11$, the scatter becomes sufficiently blurred that the linear structure is difficult to resolve visually; as we will demonstrate in Sec~\ref{sec:high_n_analysis}, this is not a limit of numerical precision, but rather the physical onset of the landscape flattening into higher-dimensional optimal hypersurfaces around the 1D backbones.}
\end{enumerate}

\secondrev{These observations strongly suggest that for $N \le 7$, the $2N$-dimensional optimization problem naturally collapses strictly onto a union of simple 1D manifolds, whereas for $N \ge 8$, these 1D loops serve as the core trajectories within expanded optimal spaces.}

This implies a \emph{secondary dimensionality reduction}: the landscape of solutions collapses from the $N$ dimensions of the Tangent Torus to exactly one dimension. The complex optimization landscape is effectively reduced to a system of coupled linear mechanical gears running on projective rails.

The linearity of these patterns allows us to hypothesize that the global solution space can be modeled as a graph of disconnected and closed curves in $\tangentTorus$. We thus formally define the phenomenon of \textbf{Affine Phase Locking}:

\begin{definition}[Affine Phase Locking]
A solution manifold exhibits Affine Phase Locking if all $N$ angles defining a solution on the $\tangentTorus$ are linear functions of a single global driving parameter $\lambda \in [0, \pi)$. The configuration of the $i$-th rotor angle on the $k$-th topological branch (\emph{Isomer}) is given by:
\begin{equation}
    \theta_i^{(k)}(\lambda) = \chi_i^{(k)} \lambda + \delta_i^{(k)} \pmod{\pi}
    \label{eq:phase_locking}
\end{equation}
where $\chi_i^{(k)} \in \{-1, +1\}$ is the locking slope (\emph{Chirality}) and $\delta_i^{(k)} \in [0, \pi)$ is the \emph{Phase Offset} of the $k$-th isomer.
\end{definition}

In the following section, we exploit this structure by formally extracting the coordination laws from the data.

\subsection{Extraction of Branches and Phase Offsets}
\label{subsec:extraction_mode}

To validate the phase-locking hypothesis and identify the locking parameters from the raw numerical  data shown in Fig.~\ref{fig:scatter_table}, we employed the custom topological extraction routine described in Algorithm~\ref{alg:branch_clustering}.

\begin{algorithm}[t]
\caption{Topological Branch \& Phase Extraction}
\label{alg:branch_clustering}
\begin{algorithmic}[1]
\Require Set of intrinsic phase vectors $\Theta = \{\bm{\theta}^{(1)}, \dots, \bm{\theta}^{({M^*})}\}$ derived from $\mathcal{S}^*$.
\Ensure Branch count $K$, Set of Phase Offsets $\{\bm{\delta}^{(1)}, \dots, \bm{\delta}^{(K)}\}$, Set of Chirality Vectors $\{\bm{\chi}^{(1)}, \dots, \bm{\chi}^{(K)}\}$.

\State \textit{Step 1: Projective Unwrapping}
\State Correct $\pi$-discontinuities in $\Theta$ to map data from modular space $[0,\pi)^N$ to covering space $\mathbb{R}^N$.

\State \textit{Step 2: Manifold Analysis (PCA)}
\State $\mathbf{X} \leftarrow [\bm{\theta}^{(1)}, \dots, \bm{\theta}^{({M^*})}]^\top$.
\State $[\mathbf{coeff}, \mathbf{score}, \mathbf{latent}] \leftarrow \text{PCA}(\mathbf{X})$.
\State $\bm{\lambda} \leftarrow \mathbf{score}(:, 1)$. \Comment{Global Parameter $\lambda$}
\If {$\mathbf{latent}(1) / \sum \mathbf{latent} < 0.999$}
    \State \Return \textbf{Error:} Solution space is not a 1D Manifold.
\EndIf

\State \textit{Step 3: Branch Clustering}
\State Apply density-based clustering (e.g., DBSCAN) on $\mathbf{X}$ to identify disjoint subsets $\mathcal{C}_1, \dots, \mathcal{C}_K$.

\State \textit{Step 4: Parameter Identification}
\For{$k = 1$ to $K$}
    \State Extract subset indices $\mathcal{I}_k$ for cluster $\mathcal{C}_k$.
    \State $\bm{\lambda}_k \leftarrow \bm{\lambda}(\mathcal{I}_k)$. \Comment{Local parameterization}
    \For{rotor $i = 1$ to $N$}
        \State $\mathbf{y} \leftarrow \mathbf{X}(\mathcal{I}_k, i)$. \Comment{Phases of rotor $i$}
        \State Fit linear model: $\mathbf{y} \approx \chi_{slope} \bm{\lambda}_k + \delta_{int}$.
        \State $\bm{\chi}^{(k)}(i) \leftarrow \operatorname{sign}(\chi_{slope})$. \Comment{Chirality of rotor $i$ on branch $k$}
        \State $\bm{\delta}^{(k)}(i) \leftarrow \delta_{int} \pmod{\pi}$. \Comment{Isomer Phase Offset}
    \EndFor
\EndFor

\State \textbf{Return} $K$, $\{\bm{\delta}^{(1)}, \dots, \bm{\delta}^{(K)}\}$, $\{\bm{\chi}^{(1)}, \dots, \bm{\chi}^{(K)}\}$.
\end{algorithmic}
\end{algorithm}

\begin{table*}[t]
    \centering
    \caption{\textbf{Affine Phase Coordination Parameters.} The table reports the phase shift $\delta_i^{(k)}$ computed by Algorithm~\ref{alg:branch_clustering} for all global solution projections on $\tangentTorus$ obtained for regular polygons and Platonic solids with $6\leq N\leq 10$ (the successful extraction cases). All phases are normalized by $\pi$. The branch index $k$ is assigned by increasing order of the phase shift $\delta_{2}^{(k)}$. \textit{Spread} indicates the tightness of the numerical cluster used to identify the branches, per Eq.~\eqref{eq:cluster_spread}. \textit{Max Fit Err} denotes the worst-case deviation of the computed phase shift from the displayed exact rational fraction of $\pi$.}
    \label{tab:estimated_phase_offsets}
    
    \footnotesize 
    \renewcommand{\arraystretch}{1.1} 
    \setlength{\tabcolsep}{3.0pt} 
    
    \resizebox{0.8\textwidth}{!}{%
        \begin{tabular}{ll c c *{10}{c}}
            \toprule
            \textbf{Chassis} & \textbf{Branch} & \textbf{Spread} & \textbf{Max Fit Err}
            & $\delta_{1}^{(k)}$ & $\delta_{2}^{(k)}$ & $\delta_{3}^{(k)}$ & $\delta_{4}^{(k)}$ & $\delta_{5}^{(k)}$ & $\delta_{6}^{(k)}$ & $\delta_{7}^{(k)}$ & $\delta_{8}^{(k)}$ & $\delta_{9}^{(k)}$ & $\delta_{10}^{(k)}$ \\
            & (k) & (deg) & (deg) &$\vdash$ & \multicolumn{8}{c}{(rad/$\pi$)} & $\dashv$ \\
            \midrule
            
            
            \textbf{CRPol6} 
            & 1 & $\pm 0.00^\circ$ & $0.00^\circ$
              & $0$ & $3/6$ & $0$ & $3/6$ & $0$ & $3/6$ & \textendash & \textendash & \textendash & \textendash \\
            \cmidrule{1-14}
    
            \multirow{2}{*}{\textbf{CRPol7}} 
            & 1 & $\pm 0.00^\circ$ & $0.00^\circ$
              & $0$ & $3/7$ & $6/7$ & $2/7$ & $5/7$ & $1/7$ & $4/7$ & \textendash & \textendash & \textendash \\
            & 2 & $\pm 0.00^\circ$ & $0.00^\circ$
              & $0$ & $4/7$ & $1/7$ & $5/7$ & $2/7$ & $6/7$ & $3/7$ & \textendash & \textendash & \textendash \\
            \cmidrule{1-14}
    
            \multirow{3}{*}{\textbf{CRPol8}} 
            & 1 & $\pm 0.00^\circ$ & $0.00^\circ$
              & $0$ & $3/8$ & $6/8$ & $1/8$ & $4/8$ & $7/8$ & $2/8$ & $5/8$ & \textendash & \textendash \\
            & 2 & $\pm 1.06^\circ$ & $0.04^\circ$
              & $0$ & $4/8$ & $0$ & $4/8$ & $0$ & $4/8$ & $0$ & $4/8$ & \textendash & \textendash \\
            & 3 & $\pm 0.00^\circ$ & $0.00^\circ$
              & $0$ & $5/8$ & $2/8$ & $7/8$ & $4/8$ & $1/8$ & $6/8$ & $3/8$ & \textendash & \textendash \\
            \cmidrule{1-14}
    
            \multirow{4}{*}{\textbf{CRPol9}} 
            & 1 & $\pm 0.00^\circ$ & $0.00^\circ$
              & $0$ & $3/9$ & $6/9$ & $0$ & $3/9$ & $6/9$ & $0$ & $3/9$ & $6/9$ & \textendash \\
            & 2 & $\pm 7.72^\circ$ & $0.09^\circ$
              & $0$ & $4/9$ & $8/9$ & $3/9$ & $7/9$ & $2/9$ & $6/9$ & $1/9$ & $5/9$ & \textendash \\
            & 3 & $\pm 7.61^\circ$ & $0.10^\circ$
              & $0$ & $5/9$ & $1/9$ & $6/9$ & $2/9$ & $7/9$ & $3/9$ & $8/9$ & $4/9$ & \textendash \\
            & 4 & $\pm 0.00^\circ$ & $0.00^\circ$
              & $0$ & $6/9$ & $3/9$ & $0$ & $6/9$ & $3/9$ & $0$ & $6/9$ & $3/9$ & \textendash \\
            \cmidrule{1-14}
    
            \multirow{5}{*}{\textbf{CRPol10}} 
            & 1 & $\pm 0.00^\circ$ & $0.00^\circ$
              & $0$ & $3/10$ & $6/10$ & $9/10$ & $2/10$ & $5/10$ & $8/10$ & $1/10$ & $4/10$ & $7/10$ \\
            & 2 & $\pm 19.11^\circ$ & $0.38^\circ$
              & $0$ & $4/10$ & $8/10$ & $2/10$ & $6/10$ & $0$ & $4/10$ & $8/10$ & $2/10$ & $6/10$ \\
            & 3 & $\pm 25.80^\circ$ & $0.44^\circ$
              & $0$ & $5/10$ & $0$ & $5/10$ & $0$ & $5/10$ & $0$ & $5/10$ & $0$ & $5/10$ \\
            & 4 & $\pm 19.21^\circ$ & $1.23^\circ$
              & $0$ & $6/10$ & $2/10$ & $8/10$ & $4/10$ & $0$ & $6/10$ & $2/10$ & $8/10$ & $4/10$ \\
            & 5 & $\pm 0.00^\circ$ & $0.00^\circ$
              & $0$ & $7/10$ & $4/10$ & $1/10$ & $8/10$ & $5/10$ & $2/10$ & $9/10$ & $6/10$ & $3/10$ \\
    
            \midrule
            \midrule
    
    
            \textbf{COct6} 
            & 1 & $\pm 0.00^\circ$ & $0.00^\circ$
              & $0$ & $0$ & $0$ & $3/6$ & $3/6$ & $3/6$ & \textendash & \textendash & \textendash & \textendash \\
            \cmidrule{1-14}
    
            \multirow{3}{*}{\textbf{CCub8}} 
            & 1 & $\pm 0.00^\circ$ & $0.00^\circ$
              & $0$ & $0$ & $1/2$ & $1/2$ & $0$ & $0$ & $1/2$ & $1/2$ & \textendash & \textendash \\
            & 2 & $\pm 0.00^\circ$ & $0.00^\circ$
              & $0$ & $2/3$ & $2/3$ & $0$ & $1/6$ & $1/2$ & $1/2$ & $1/6$ & \textendash & \textendash \\
            & 3 & $\pm 0.00^\circ$ & $0.00^\circ$
              & $0$ & $5/6$ & $1/3$ & $1/2$ & $1/3$ & $1/2$ & $0$ & $5/6$ & \textendash & \textendash \\
    
            \bottomrule
        \end{tabular}%
    }
\end{table*}

\begin{figure*}[t]
\centering
\footnotesize
\begin{tabularx}{\textwidth}{@{\hspace{0pt}} >{\centering\arraybackslash}m{0.08\linewidth} @{\hspace{1pt}} | @{\hspace{0pt}} X @{\hspace{0pt}}}
\hline
\textbf{ID} & \multicolumn{1}{c}{\textbf{Pairwise Correlation Scatters Fitted by Multiple Isomers via Phase Shift Affine Clustering}} \\ 
            & \multicolumn{1}{c}{($\theta_1$ vs. $\theta_i$ for $i=1 \dots N$)} \\
\hline

\rowcolor[gray]{0.95} \multicolumn{2}{l}{\textit{Regular Polygons}} \\
\hline
\textbf{CRPol6} & \includegraphics[width=0.99\linewidth, valign=m]{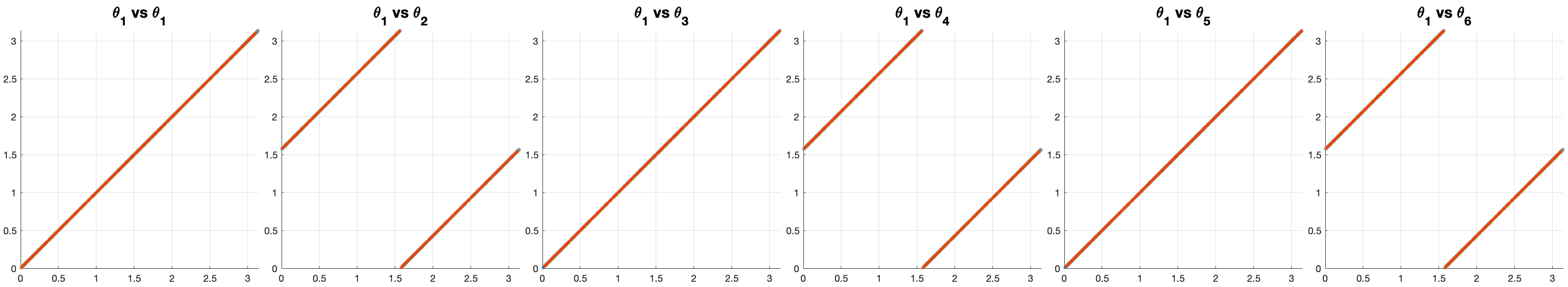} \\
\hline
\textbf{CRPol7} & \includegraphics[width=0.99\linewidth, valign=m]{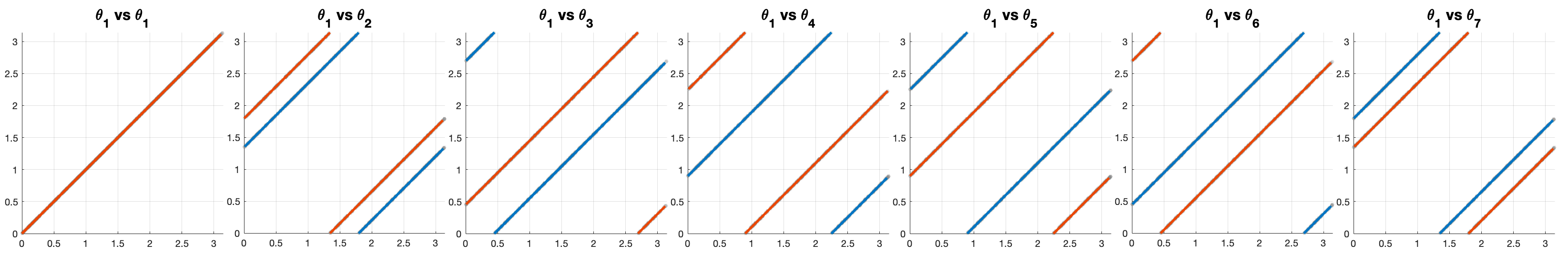} \\
\hline
\textbf{CRPol8} & \includegraphics[width=0.99\linewidth, valign=m]{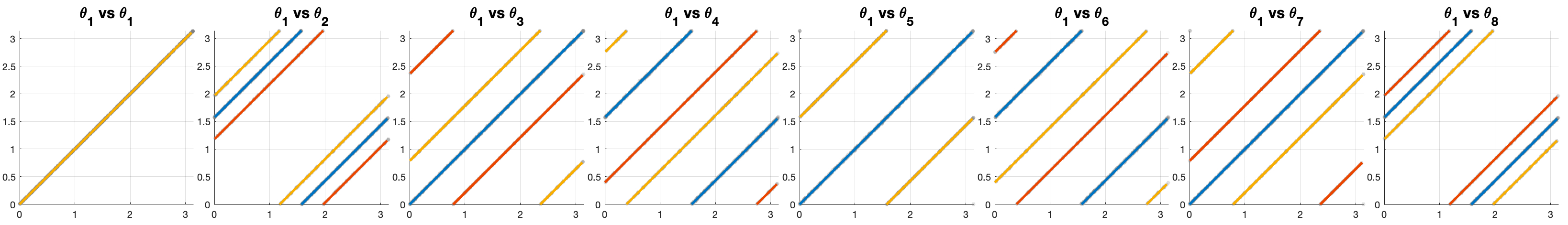} \\
\hline
\textbf{CRPol9} & \includegraphics[width=0.99\linewidth, valign=m]{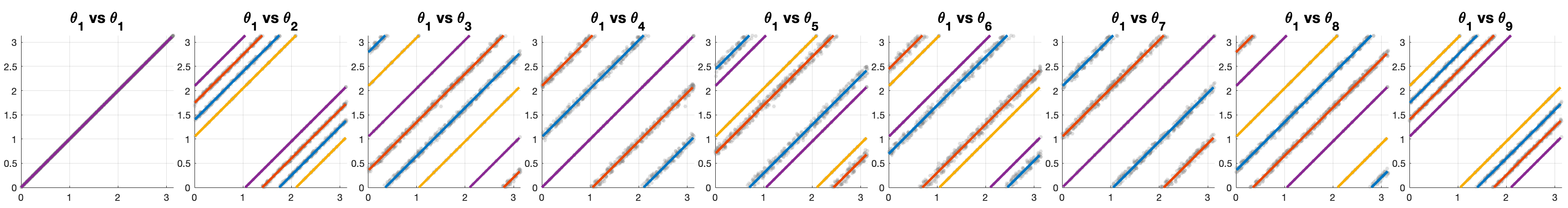} \\
\hline
\textbf{CRPol10} & \includegraphics[width=0.99\linewidth, valign=m]{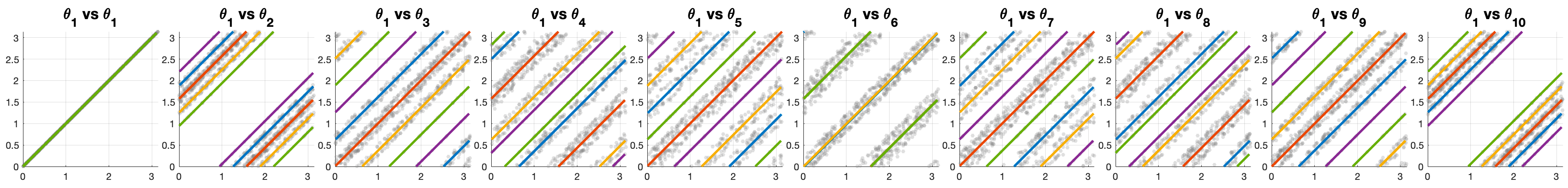} \\
\hline
\textbf{CRPol11} (rejected) & \includegraphics[width=0.99\linewidth, valign=m]{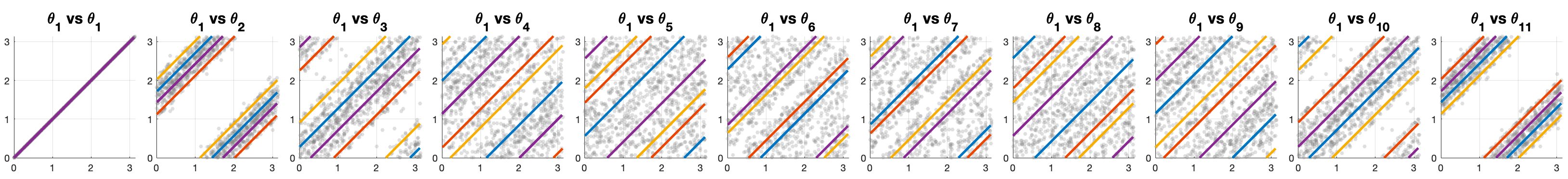} \\

\Xhline{2pt}
\rowcolor[gray]{0.95} \multicolumn{2}{l}{\textit{Platonic Solids}} \\
\hline
\textbf{COct6} & \includegraphics[width=0.99\linewidth, valign=m]{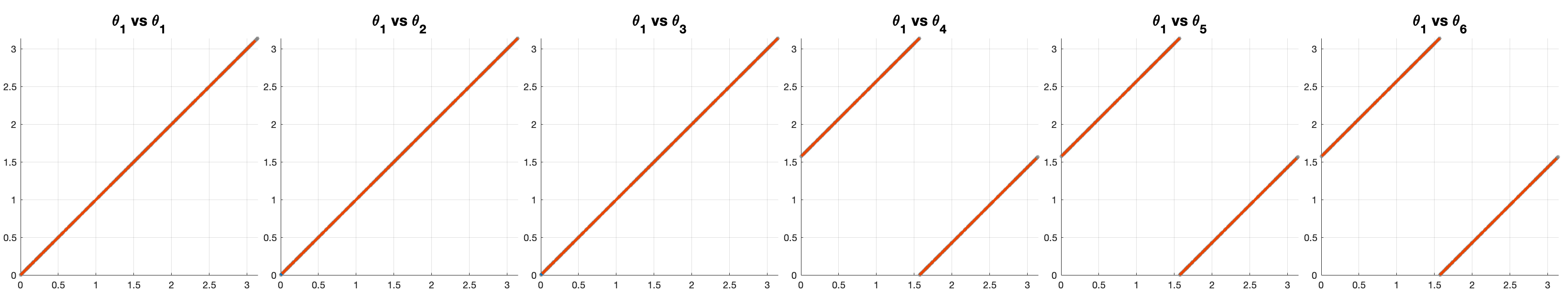} \\
\hline
\textbf{CCub8} & \includegraphics[width=0.99\linewidth, valign=m]{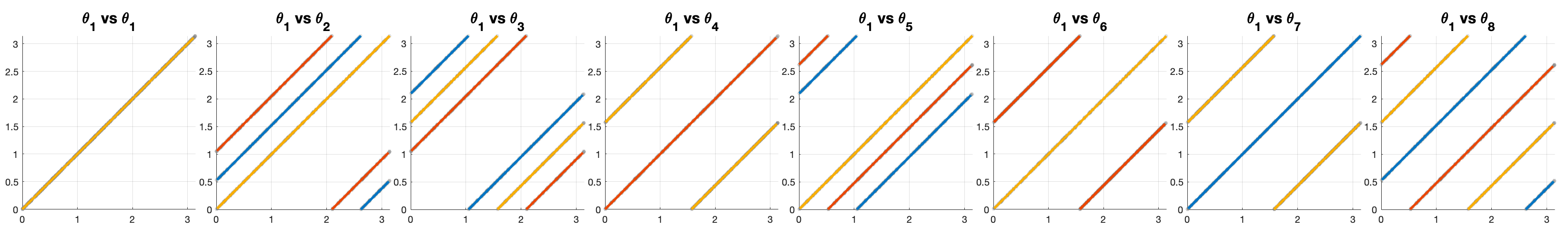} \\
\hline
\textbf{CIco12}  (rejected) & \includegraphics[width=0.99\linewidth, valign=m]{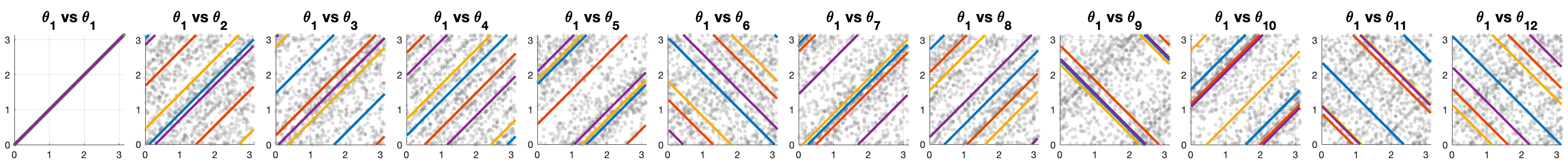} \\
\hline
\textbf{CDod20}  (rejected) & \includegraphics[width=0.99\linewidth, valign=m]{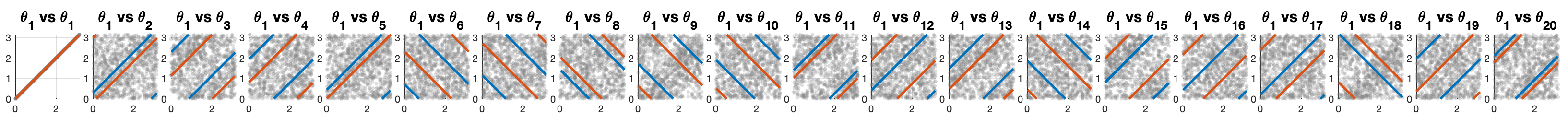} \\
\hline
\end{tabularx}

\caption{\textbf{Visual results of the topological branch and phase extraction of the landscape of optimal solutions.} This figure demonstrates the application of Algorithm~\ref{alg:branch_clustering} to the solution data from Fig.~\ref{fig:scatter_table} and the subsequent 1D manifold branch extraction. Each row displays the multi-branch solution manifold for a specific regular chassis. The data points are overlaid with the fitted affine phase-locking curves, colored to distinguish the distinct $K$ topological branches (isomers) identified by the algorithm for each chassis. The cluster spreads for all chassis with $N\leq 10$ fall within the admissible range (see Table~\ref{tab:estimated_phase_offsets}). In contrast, CRPol11, CIco12, and CDod20 were rejected due to excessive spread but are included here to illustrate the failure modes of the extraction.}
\label{fig:branch_extraction}
\end{figure*}

The algorithm partitions the unwrapped phase data into $K$ disjoint subsets, representing distinct topological branches or \textit{isomers}. Within each branch $k$, a linear regression of the $i$-th rotor angle $\theta_i$ against the principal manifold coordinate $\lambda$ yields the locking slope $\chi_i^{(k)}$ and the phase offset $\delta_i^{(k)}$.

To assess the quality of the extraction and establish a criterion for acceptance, we introduce the \textbf{Branch Cluster Spread}. This error metric quantifies the tightness of the physical coordination--specifically, how closely the numerical solutions adhere to the idealized 1D manifold. 
Define the set of intrinsic phase vectors $\Theta = \{\bm{\theta}^{(1)}, \dots, \bm{\theta}^{({M^*})}\}$ derived from $\mathcal{S}^*$.
For every branch $\mathcal{B}^*_k \subset \Theta$ ($k=1,\dots,K$) identified by Algorithm~\ref{alg:branch_clustering}, we define the \emph{cluster spread} as the Root Mean Square (RMS) deviation from the centroid $\boldsymbol{\mu}^{(k)}$:
\begin{equation}
    \theta_{\text{spread}}^{(k)} = \sqrt{ \frac{1}{|\mathcal{B}^*_k|} \sum_{\bm{\theta} \in \mathcal{B}^*_k} \left\| \bm{\theta} \ominus \boldsymbol{\mu}^{(k)} \right\|^2 }.
    \label{eq:cluster_spread}
\end{equation}
where $\ominus$ denotes the geodesic distance on the projective circle $[0, \pi)$.

The \textit{Spread} column in Table~\ref{tab:estimated_phase_offsets} reports these values for each identified branch.
We classify an error below $\pm 1^\circ$ as excellent and below $\pm 10^\circ$ as good. We define $\pm 30^\circ$ as the maximum tolerable threshold for a result to be accepted.

We applied this methodology to the full library of regular chassis. Table~\ref{tab:estimated_phase_offsets} and Fig.~\ref{fig:branch_extraction} provide a numerical and visual summary, respectively, of the extracted affine coordination parameters. Algorithm~\ref{alg:branch_clustering} yielded  clustering results  below the admissibility threshold for all regular chassis with $N \leq 10$.

Notably, we do not explicitly report the estimated integer slopes $\chi_i^{(k)}$. Consistent with the visual inspection of the empirical data in Fig.~\ref{fig:scatter_table}, the algorithmic extraction reveals a strict \emph{homochirality} in these coordinates: specifically, $\chi_i^{(k)}=1$ for all $i=1,\dots,N$ across all observed branches for chassis with $N \leq 10$. Consequently, we omit the chirality coefficient in subsequent derivations, treating it as unity.

\secondrev{Conversely, the cluster spread begins to increase slightly for $N = 8$ until it consistently exceeds the admissibility threshold for chassis with $N > 10$. This degradation substantiates the ``blurring'' phenomenon described previously, physically marking the transition where the 1D loops become backbones embedded within broader multidimensional hypersurfaces, see Sec.~\ref{sec:high_n_analysis}.}
For completeness, Fig.~\ref{fig:branch_extraction} illustrates the failure to cluster data for the hendecagon ($N=11$), the icosahedron ($N=12$), and the dodecahedron ($N=20$). While we address these limitations in Sec.~\ref{subsec:conj_resolution}, the remainder of this section focuses on the successful cases ($N\leq 10$).

To facilitate the discovery of underlying patterns and governing laws, we aimed to present the results in an interpretable format, specifically as rational fractions of $\pi$. In this context, we introduced the \textbf{Rational Fit Error}. This metric quantifies the deviation of the extracted phase offsets from an exact rational fraction of $\pi$. The extracted phases $\delta_i^{(k)}$ were approximated to the nearest rational fraction, prioritizing $N$ as the denominator where applicable. The \textit{Max Fit Error} column in Table~\ref{tab:estimated_phase_offsets} reports the worst-case deviation among all $N$ rotor phases for the given branch.

\firstrev{
\subsection{Topological Degradation and Hypersurface Expansion as $N$ Increases}
\label{subsec:conj_resolution}

Attempts to cluster the data for regular chassis with $N > 10$ to isolate the predicted $N-5$ branches prove highly elusive. As illustrated in Fig.~\ref{fig:branch_extraction}, for these higher-order polygons, the distinct topological branches coalesce, rendering clear separation via Algorithm~\ref{alg:branch_clustering} impossible. 

For the verified cases ($N \geq 10$), solution branches are tightly defined with spreads consistently below $26^\circ$ (see Table~\ref{tab:estimated_phase_offsets}). In stark contrast, applying the extraction algorithm to the regular Hendecagon ($N=11$) yields clusters with spreads exceeding $51^\circ$, while the Icosahedron ($N=12$) data degrades into a single unstructured cloud with a spread of $\approx 148^\circ$. 

Initially, one might conjecture that this obfuscation is strictly a numerical resolution limit of the solver—assuming the energy barriers (Log-Volume gradients) separating the $K=N-5$ branches simply fall below finite-precision thresholds. However, this algorithmic difficulty is actually symptomatic of a fundamental topological degradation. \secondrev{As the number of rotors increases beyond $N=7$, the $N-5$ distinct 1D continuous valleys still exist;} however, the surrounding landscape flattens in orthogonal directions, causing these isolated loops to become embedded core trajectories within broader, multi-dimensional optimal hypersurfaces. This flattening causes the optimizer to drift freely in the neighborhood of the core isomers while still remaining confined to the tangent torus. We prove numerically this physical dimensional expansion of the design null space in Section~\ref{sec:high_n_analysis} by tracking the collapse of the numerical Hessian's eigenvalues.
}

\section{Study on the Evolution of the Solution Landscape for Increasing~$N$}
\label{sec:high_n_analysis}

While the primary focus of this work lies in establishing the $N-5$ scaling law—which dictates the existence of $N-5$ continuous one-dimensional manifolds within the solution landscape—a natural avenue for future research concerns the topological evolution of this space in higher-dimensional regimes (as discussed in Sec.~\ref{subsec:conj_resolution}). At lower values of $N$, the design null space manifests strictly as isolated one-dimensional curves. To establish a preliminary foundation for the study of higher-order systems, it is critical to examine whether this 1D characteristic remains invariant or if the null space geometrically expands into a higher-dimensional hypersurface. Accordingly, this section presents an exploratory numerical analysis designed to probe the local topology of the high-$N$ solution space, mapping the structural evolution of the distinct branches and quantifying the expanded dimensionality of these flat valleys.

\subsection{Methodology and Computational Setup}
To investigate the geometry of the null space, we employ a localized eigenvalue analysis of the system's Hessian along the identified 1D branches. For each configuration—defined by the system scale $N$, branch index $k$, and the associated topological star parameter $q$—we evaluate the number of effectively flat dimensions of the cost function, denoted as $n_H$. As formalized in \eqref{eq:phase_locking}, each branch represents a continuous 1D curve parameterized by $\lambda \in [0, \pi]$. We discretize this parameter interval into 50 uniform samples and compute the system's Hessian at each discrete configuration along the trajectory. For each sample, the local dimensionality of the null space, $n_H$, is defined by the number of principal eigenvalues of the Hessian that fall below a strict numerical threshold ($\approx 10^{-7}$). These near-zero eigenvalues indicate orthogonal directions of vanishing curvature, mathematically corresponding to a flat region in the optimization landscape.

The presence of near-zero eigenvalues alone is insufficient to differentiate a contiguous global valley from a collection of disjoint, localized numerical artifacts. To formally verify that the $N-5$ predicted branches act as continuous, privileged trajectories, we first evaluate the geometric alignment of the 1D manifold. By confirming that the tangent vector to the parameterized branch remains strictly contained within the null space of the local Hessian at every infinitesimal step, we prove the trajectory represents a smooth, zero-variation self-motion. Furthermore, to verify the structural continuity of the entire optimal subspace along this path, we evaluate a Subspace Alignment metric. By calculating the principal angles (inner products) between the null space basis vectors of adjacent configurations, we mathematically confirm that the optimal subspace does not undergo abrupt topological shifts, proving the valley is physically contiguous. Finally, to empirically establish the dimensionality of this landscape, we extract the sorted principal eigenvalues ($\mu_1$ through $\mu_{N}$) of the numerical Hessian; the emergence of additional near-zero eigenvalues formally establishes the boundary and expansion of the surrounding flat hypersurface.

\begin{figure*}[t]
    \centering
    \includegraphics[width=0.99\textwidth]{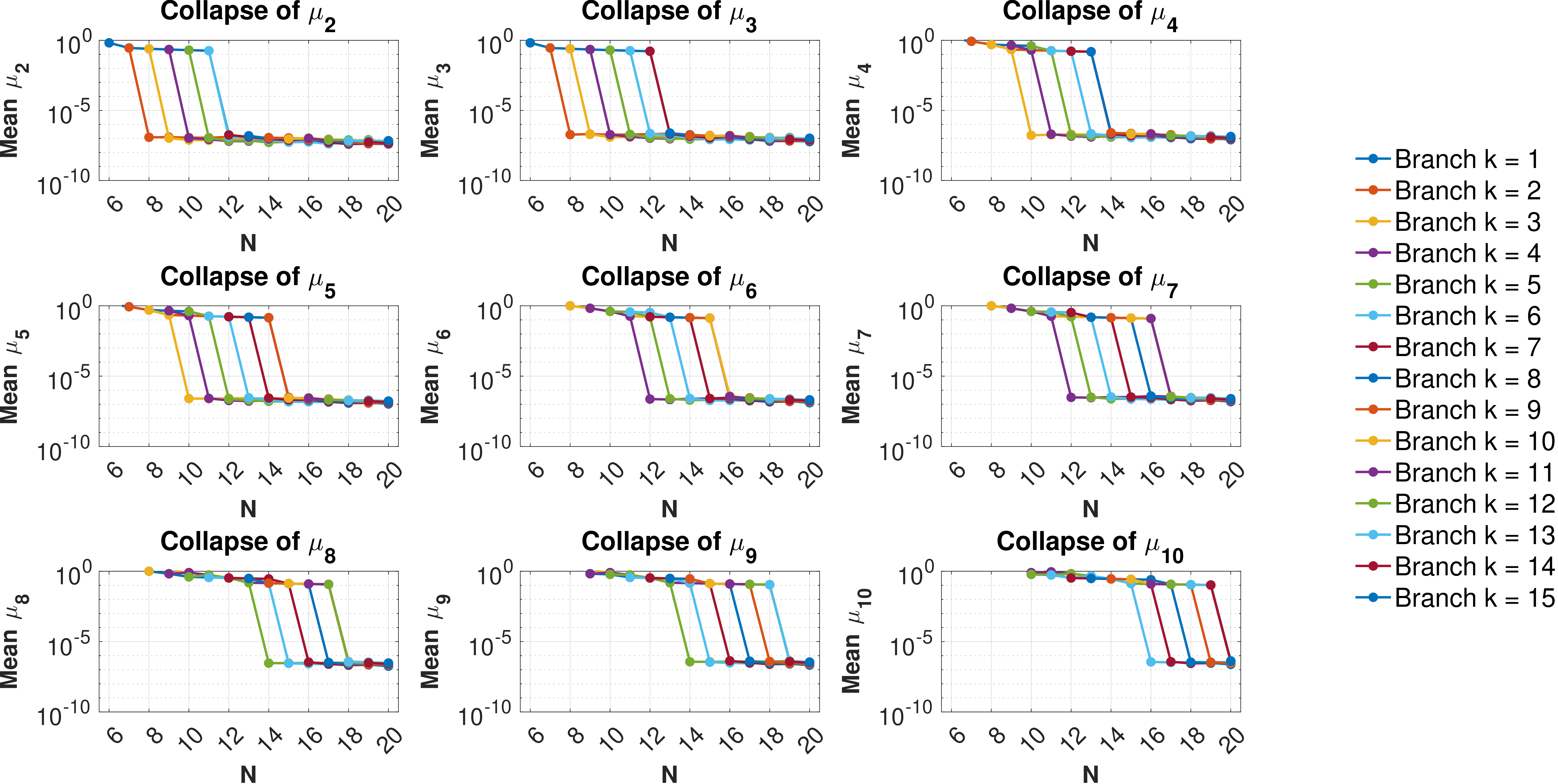}
    \caption{\firstrev{Evolution of orthogonal landscape stiffness and the dimensional expansion of the design null space (see Tables~\ref{tab:high_n_data_part1} and~\ref{tab:high_n_data_part2} in Appendix~\ref{app:tables_hessian}). The subplots trace the mean absolute principal eigenvalues ($\mu_i$) of the numerical Hessian evaluated along the continuous 1D trajectories, parameterized by system scale ($N$) and branch index ($k$). The logarithmic scale isolates the discrete transition of transverse dimensions from locally strictly convex (high stiffness) to mathematically flat ($\mu_i < 10^{-7}$). The cascading collapse of higher-order eigenvalues, particularly for central branches (intermediate $k$), provides direct graphical confirmation of the $N$-dependent geometric expansion of the local hypersurface.}}
    \label{fig:landscape_flattening}
\end{figure*}

\subsection{Numerical Results and Interpretation of Topological Evolution}

The results of the subspace alignment and valley flattening analysis for $6 \le N \le 20$ are presented in Tables~\ref{tab:high_n_data_part1} and~\ref{tab:high_n_data_part2} in Appendix~\ref{app:tables_hessian}. For the sake of visualization the eigenvalues are displayed until $\mu_{10}$, however $n_H$ is computed using all the eigenvalues.

The data provide clear numerical evidence regarding the evolution of the solution subspace. In the strictly low-$N$ regime ($N \le 7$), the number of flat dimensions is exactly $n_H = 1$, confirming that the physical solution spaces in these configurations are isolated 1D curves.

However, a geometric transition occurs starting at $N=8$, where the null space begins to expand into a higher-dimensional volume ($n_H = 3$ for branch $k=2$). This trend continues as system size increases. By $N=10$, we observe branches yielding $n_H = 5$. By the time the system scales to $N=20$, the central core branches exhibit significantly expanded null spaces, with the dimensionality of the flat valley reaching up to $n_H = 15$. 

These results confirm that the 1D trajectories predicted by the $N-5$ scaling law act as contiguous submanifolds embedded within higher-dimensional flat hypersurfaces of different sizes depending on $q$. Furthermore, the local dimensionality of this expanded null space scales with $N$ and attains a maximum for branches characterized by intermediate values of the index $q$.

Three numerical indicators validate this physical interpretation:
\begin{enumerate}
    \item \textbf{Subspace Alignment}: Across all measured scales and branches, the subspace alignment evaluates to unity within machine precision. This ensures that the identified flat dimensions are mathematically smooth and represent contiguous, navigable physical manifolds rather than discontinuous numerical anomalies.
    \item \textbf{Mean Eigenvalue Bounds}: By observing the mean eigenvalue columns (Mean $\mu_1$ to $\mu_{10}$), a clear structural boundary is evident. For principal directions falling within the designated $n_H$ dimensions, the associated eigenvalue remains strictly bounded between $10^{-8}$ and $10^{-7}$, denoting zero curvature. The moment the dimension index exceeds $n_H$ (e.g., $\mu_2$ for $N=7, k=1$ where $n_H=1$, or $\mu_4$ for $N=8, k=2$ where $n_H=3$), the mean eigenvalue exhibits a discrete jump of several orders of magnitude. This sharp gradient defines the physical boundary of the flat valley.
    \item \textbf{Branch Differentiation}: The topological variations (represented by $k$ and $q$) dictate the geometry of the expanding subspace, verifying the existence of distinct, stable branches whose null space dimensionality scales dynamically with $N$.
\end{enumerate}

To visually synthesize the dimensional expansion detailed in the preceding tables, Figure~\ref{fig:landscape_flattening} illustrates the cascading collapse of the orthogonal constraint stiffness as the system scales. By plotting the mean absolute magnitude of the higher-order principal eigenvalues ($\mu_2$ through $\mu_{10}$) across the evaluated branches, a clear geometric bifurcation becomes evident. While the primary eigenvalue ($\mu_1$) remains functionally zero across all configurations—defining the contiguous floor of the 1D valley—successive transverse dimensions quickly drop into the numerical null space ($\approx 10^{-7}$) at specific $N$ thresholds. This graphical representation highlights the branch-dependent nature of the phase transition: central topological branches (intermediate $k$ values) experience an accelerated expansion of the null space, shedding orthogonal stiffness earlier and opening into higher-dimensional flat volumes more rapidly than peripheral branches.

\subsection{Future Projections and Physical Significance}
The finding that the $N-5$ scaling law is associated with the emergence of multiple flat dimensions around a central core carries practical consequences for multirotor design. Because the dimensionality of the flat valleys scales with $N$, high-order systems ($N > 10$) possess a continuous hypersurface of valid configurations.

This structural flexibility provides a foundation for future research. Rather than being confined to a single optimization path, the higher-dimensional null space enables the integration of secondary constraints—such as structural dynamics, energy minimization, or real-time reconfiguration—without violating the primary isotropy invariants of the system. Mapping and exploiting these multi-dimensional flat valleys will be central to the development of scalable design algorithms for complex, fully actuated aerial vehicles.

\begin{table*}[t]
\centering
\caption{\textbf{Unified Taxonomy of the Solution Manifold Landscapes for Maximally Isotropic MRAV Design.}
This classification synthesizes the topological behaviors observed across the full chassis library (cf. Table~\ref{tab:chassis_zoo}). It categorizes chassis based on the landscape of the optimal rotor line-of-force directions within $\Manifold=(\RPtwo)^N$, as identified by the solver for each specific positional arrangement. The categories distinguish between unstructured or pre-asymptotic states (Types I--III), characteristic of irregular chassis, and coherent manifold collapses (Type IV), typical of quasi-regular and regular chassis. The latter establishes the primary domain of interest for high-performance reconfigurable design.}
\label{tab:unified_classification}
\small
\renewcommand{\arraystretch}{1.35} 

\renewcommand{\tabularxcolumn}[1]{p{#1}}
\begin{tabularx}{\textwidth}{@{} 
    >{\bfseries\raggedright\arraybackslash}p{0.04\textwidth}   
    >{\raggedright\arraybackslash}p{0.16\textwidth}            
    >{\raggedright\arraybackslash}p{0.15\textwidth}            
    >{\raggedright\arraybackslash}X                              
@{}}
\toprule
\textbf{Class} & \textbf{Chassis} & \textbf{Topological Landscape} & \textbf{Phenomenological Description} \\
\midrule

Type I & 
\textbf{CTriPr6}, \textbf{CQRPol6}, \newline
\textbf{CQRPol7}, \textbf{CQRPol9} & 
\textbf{Discrete Point Set} \newline
$\dim(\mathcal{S}^*) \approx 0$ & 
\textbf{Sparse \& Rigid.} The optimizer converges to a sparse set of distinct, isolated points ($<10$). The landscape is rigid, offering no continuous degrees of freedom for reconfiguration. \\
\midrule

Type II & 
\textbf{CPentBi7}, \textbf{CTriCup9} & 
\textbf{Unstructured Cloud} \newline
$\mathcal{S}^* \subset (\RPtwo)^N$ & 
\textbf{Clustered Scattering.} Solutions form large, nebulous "islands" or high-entropy clouds covering portions of the moduli discs representing $\Manifold$. There is no global coherence or connectivity, implying a lack of correlation between rotor orientations. \\
\midrule

Type III & 
\textbf{CQRPol8}, \textbf{CSqAnti8}, \newline
\textbf{CQRPol10} & 
\textbf{Partial Collapse on the Tangent Torus} \newline
$\mathcal{S}^* \to \tangentTorus$ & 
\textbf{Condensing / Hybrid.} A "pre-collapse" phase of the solutions where subsets of directions aggregate into semi-elliptic bands approximating the planes of the tangent torus, while others remain scattered. The chassis symmetry is insufficient to fully enforce the tangent torus constraint. \\
\midrule

\multicolumn{4}{@{}l}{\textbf{Type IV: Coherent Collapse on the Tangent Torus ($\mathcal{S}^* \subseteq \tangentTorus$)}} \\
\multicolumn{4}{@{}l}{\textit{Solutions are strictly confined to the $N$-dimensional Tangent Torus. This represents the first-level  dimensionality collapse from $2N$ to $N$. }} \\
\multicolumn{4}{@{}l}{\textit{Sub-classification is defined by internal phase correlations.}} \\
\cmidrule(r){1-4}

IV-A & 
\textbf{CQCub8}, \textbf{CCubOct12}, \newline
\textbf{CHexPr12} & 
\textbf{No Phase Locking} \newline
(No 1D Structure) & 
\textbf{Uncorrelated Tangency.} Solutions reside on the tangent torus but lack strong phase correlations. The manifold retains residual thickness or forms shallow valleys, exploring the full volume of $\tangentTorus$ without collapsing onto a lower-dimensional submanifold. \\
\arrayrulecolor{black!30}\midrule

IV-B & 
\textbf{CRPol6}, \textbf{CRPol7}, \newline
\textbf{COct6}, \textbf{CCub8} & 
\textbf{1D Phase Locking with Branches} \newline
$\mathcal{S}^* \cong \bigcup S^1$ \newline 
$\dim(\mathcal{S}^*$) = 1& 
\textbf{The N-5 Empirical Scaling Law.}
The solution landscape exhibits strong synchronization, characterized by a perfect collapse onto $K=N-5$ branches of closed 1D curves with unitary slopes. This is the second-level dimensionality collapse, from $N$ to $1$. These branches represent discrete symmetric isomers separated by fixed phase shifts. For polygons, these shifts are rigorously predicted by the discovered law~\eqref{eq:polygon_phase_law} (Star Polygon Isomorphism).
\\
\midrule
\vspace{-1em}
\secondrev{IV-C} & \vspace{-1em}
\secondrev{\textbf{CRPol8+}, \newline
\textbf{CIco12}, \textbf{CDod20}} & \vspace{-1em}
\secondrev{\textbf{Topological Flattening} \newline
(Hypersurface Expansion)} & \vspace{-1em}
\secondrev{\textbf{The N-5 Scaling Law as a Topological Backbone.} Strong fitting of the solution landscape with the tangent torus remains evident, but the discrete isolated lines characteristic of IV-B begin to blur. As proven via Hessian analysis (Sec.~\ref{sec:high_n_analysis}), this is not merely a numerical resolution limit; starting at $N=8$, the strictly 1D valleys expand into higher-dimensional hypersurfaces. The $N-5$ 1D isomers persist as the fundamental topological core, but the flattened landscape allows continuous expansion in some of the orthogonal directions.} \\
\arrayrulecolor{black}\bottomrule
\end{tabularx}
\end{table*}

\section{The Empirical $N-5$ Scaling Law and Chassis Classification}
\label{sec:empirical_law}

The analysis of the extracted parameters in Sec.~\ref{sec:scaling_law} and ~\ref{sec:high_n_analysis} reveals three fundamental topological constraints governing the landscape of solutions for regular chassis:

\begin{enumerate}
    \item \textbf{Unitary Chirality:} For all identified branches across all chassis types, the chirality is strictly unitary ($\chi_i^{(k)} =1$).
    This implies that the solutions are obtained by rotating the $N$ thrust force lines all the same way in the local frames of  $\tangentTorus$.
    
    \item \textbf{The N-5 Branching Law:} The number of distinct disconnected solution branches, $K$, scales linearly with the number of rotors $N$. 
    \secondrev{Specifically, for all Regular Polygons and Platonic Solids where the 1D structures can be successfully extracted ($6 \le N \le 10$)—whether as strictly isolated branches ($N \le 7$) or core backbones ($N \ge 8$)—the count satisfies:}
    \begin{equation}
        K = N - 5
    \end{equation}
    This holds true for the Polygons ($N=6 \to K=1$,\ldots, $N=10 \to K=5$) as well as the Platonic solids (Octahedron $N=6 \to K=1$, Cube $N=8 \to K=3$).
    
    \item \textbf{Rational Phase Quantization:} While the phases for the Platonic solids map to simple rational multiples of $\pi$ (e.g., thirds and halves), the Regular Polygons exhibit a rigorous arithmetic progression. The phase shift for any rotor $i$ in any branch $k$ is deterministically governed by the rotor count $N$ as a multiple of $\pi/N$. This dependence is formalized in Eq.~\eqref{eq:polygon_phase_law} and is further analyzed in Section~\ref{sec:star_polygons}.
\end{enumerate}

Based on these observations \secondrev{and the analysis in Sec.~\ref{sec:scaling_law} and~\ref{sec:high_n_analysis},} we formalize the empirical law for regular polygons as follows:

\begin{empirical}[The N-5 Law]
\secondrev{For a generic fully actuated/omnidirectional platform with a regular chassis $\mathcal{C}$ (Regular Polygon or Platonic Solid) with number of vertices (rotors) equal to $N$, the landscape of maximal isotropic configurations, i.e., minimizing the log volume cost function defined in \eqref{eq:cost_logvolume}, features exactly:
\begin{equation}
    K = N - 5
\end{equation}
1D topological branches. For low complexity chassis ($N \le 7$), the solution space consists strictly of these disconnected closed curves, whereas for higher-dimensional architectures ($N \ge 8$), these curves act as the fundamental topological backbones embedded within multidimensional optimal hypersurfaces. In all cases, these 1D trajectories are completely contained within the tangent torus $\tangentTorus$ and defined by Eq.~\eqref{eq:phase_locking}, in which each phase coordinate of $\tangentTorus$ rotates at the same unitary speed (mono-chirality) with constant phase shifts $\delta_{i}^{(k)}$ reported in Tab.~\ref{tab:estimated_phase_offsets}, with $i=1,\ldots,N$ and $k=1,\ldots,K$.}

Furthermore, in the case in which $\mathcal{C}$ is a regular polygon the phase shift $\delta_{v}^{(k)}$ for the $v$-th vertex ($v=1\dots N$) in the $k$-th branch ($k=1\dots N-5$) is given exactly by:
\begin{equation}
    \delta_{v}^{(k)} = (v-1) \frac{k+2}{N} \pi \pmod{\pi}.
    \label{eq:polygon_phase_law}
\end{equation}
\end{empirical}

\subsection{Taxonomy of Topological Collapses}
\label{subsec:taxonomy}

This empirical investigation culminates in a comprehensive classification of the MRAV design space. We have uncovered that the solution landscape is not monolithic; rather, it exhibits a non-trivial spectrum of topological behaviors strictly governed by the regularity and symmetry of the chassis. 

Table~\ref{tab:unified_classification} synthesizes these findings into a unified taxonomy. This classification maps the progression of the solution space from disordered, rigid point sets (Type I) to the highly structured, phase-locked manifolds (Type IV) that emerge in the most symmetric configurations. This table serves as the primary map for understanding how physical symmetry dictates the availability and structure of optimal design solutions.

\section{Geometric Isomorphism to Star Polygons}
\label{sec:star_polygons}

\textit{Note: The following section provides a geometric interpretation of the empirical results presented above. It offers a theoretical justification based on topological observations rather than a rigorous mathematical proof. The goal is to highlight a structural isomorphism between the solution landscape and classical geometry, potentially stimulating further inquiry into the topological properties of the landscape of solutions in the phase space $\tangentTorus$.}

\subsection{Mapping Phase Shifts to Star Densities}

The rational quantization of the phase shifts described by Eq.~\eqref{eq:polygon_phase_law} is not arbitrary. For a regular polygon chassis with $N$ rotors, the extracted phase offsets $\delta_i^{(k)}$ follow an arithmetic progression. Specifically, the phase offsets listed in Table~\ref{tab:estimated_phase_offsets} for polygons are all elements of the set:
\begin{equation}
    \left\{ q \frac{\pi}{N} \;\middle|\; q \in \mathbb{Z}, \, 0 \le q < N \right\},
\end{equation}
i.e., for any given solution branch $k$, the sequence of phases across the rotors $i=1\dots N$ corresponds exactly to the sequence of multiples of a specific integer $q$ modulo $N$.
For instance:
\begin{itemize}
    \item \textbf{Hexagon ($N=6$):} The single observed branch corresponds to the generator $q=3$. The phase sequence is:
    $ \{0, 3, 6, 9, 12, 15\} \cdot \tfrac{\pi}{6} \pmod{\pi} \equiv \{0, 3, 0, 3, 0, 3\} \cdot \tfrac{\pi}{6} $
    
    \item \textbf{Heptagon ($N=7$):} We observe two branches corresponding to generators $q=3$ and $q=4$. For $q=3$, the sequence is:
     $\{0, 3, 6, 9, 12, 15, 18\} \cdot \tfrac{\pi}{7} \pmod{\pi} \equiv \{0, 3, 6, 2, 5, 1, 4\} \cdot \tfrac{\pi}{7} $
\end{itemize}

\begin{figure*}[t]
    \centering

    \includegraphics[width=1.0\linewidth]{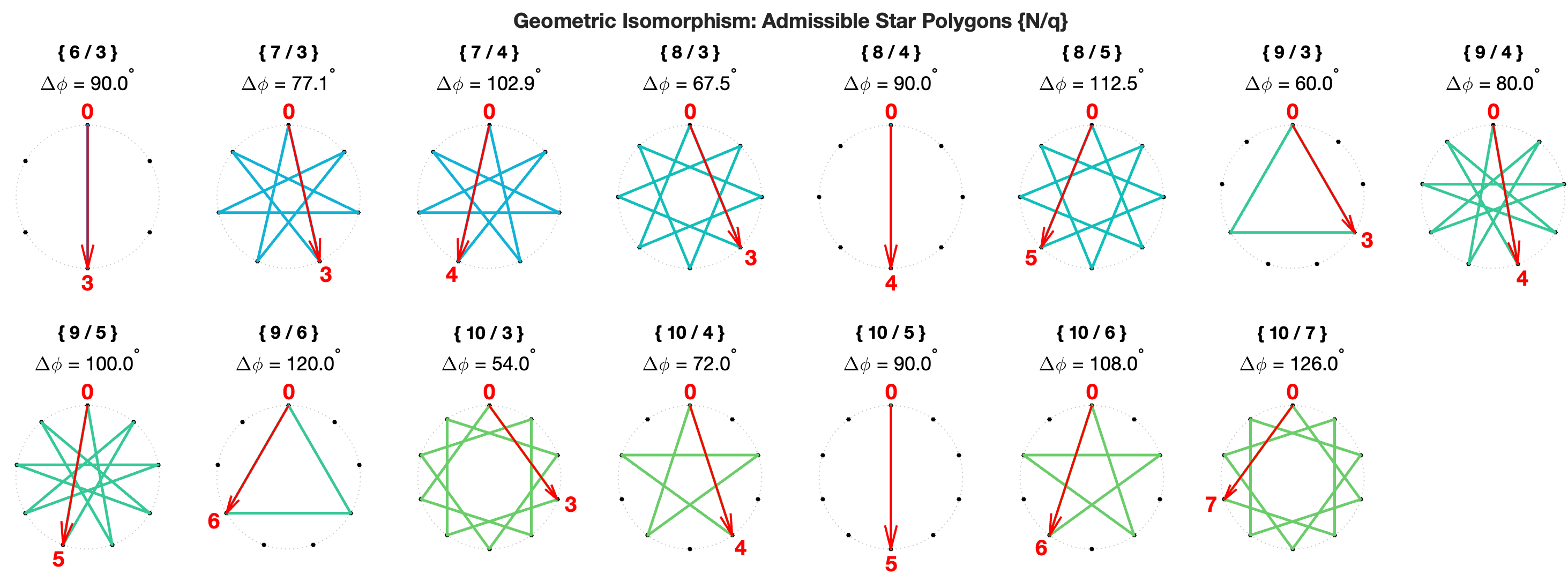}
    
    \caption{\textbf{Star Polygons $\{N/q\}$ which are admissible according to the $N-5$ Law.} 
    A grid visualization of star polygons for $N \in [6, 10]$ and $q \in [3, N-3]$. 
    Each subplot represents a specific geometry where the red arrow indicates the initial phase step ($\Delta\phi$) corresponding to the "chord" length of the polygon. Each star polygon denoted by $\{N,q\}$ in this figure maps to one of the branches of the $N$-vertex polygonal chassis in Tab.~\ref{tab:estimated_phase_offsets}, with $q\in\{3,\ldots, N-3\}$. }
    
    \label{fig:star_polygons}
\end{figure*}

These sequences map directly to the geometry of Star Polygons. A regular star polygon, denoted by the Schläfli symbol $\{N/q\}$, is a self-intersecting polygon created by connecting every $q$-th vertex of a regular $N$-gon \citep{coxeter1973regular}. The integer $q$ represents the \emph{density} (or winding number) of the polygon, see Fig.~\ref{fig:star_polygons} for the representation of all the star polygons matching one-to-one all the $15$ branches listed in Table~\ref{tab:estimated_phase_offsets}.

The phase step $\Delta \phi = q\pi/N$ derived from our data has a precise geometric interpretation. It is exactly the \emph{inscribed angle} subtended by a single edge of the star polygon $\{N/q\}$. In topological terms, it represents the uniform phase step required to wind the phase coordinate exactly $q$ times around the domain $[0, \pi)$ over the course of the $N$ rotors directions while going through all adjacent  vertices of the polygonal chassis.

By comparing this definition with our empirical law in Eq.~\eqref{eq:polygon_phase_law}, we identify a direct isomorphism between the topological branch index $k$ and the star density $q$. Equating the phase progression terms yields:
\begin{equation}
    q \equiv k + 2.
\end{equation}
This mapping reveals that each solution branch corresponds to a specific ``mode'' of coordination. 

While the rotors physically reside on a convex hull $\{N/1\}$ (the chassis $\mathcal{C}$), their thrust lines, in order to correspond to optimal solutions on the $\tangentTorus$, must coordinate according to the topology of the star polygon $\{N/q\}$. 
This mapping constitutes a geometric isomorphism that serves a dual explanatory purpose:
\begin{itemize}
    \item \textbf{Numerically}, it is simple: the phase step is literally $\Delta\phi = q\pi/N$, matching the inscribed angle of the star geometry exactly.
    \item \textbf{Physically}, it is deep: it implies the optimization problem is seeking ``harmonic modes.'' The system is locking onto integer winding numbers (topological invariants) rather than converging to arbitrary continuous values.
\end{itemize}

\subsection{Justification of the $N-5$ Law and Spatial Spanning}
This geometric isomorphism provides an analytic justification for the empirically observed $N-5$ scaling law. The set of all possible integer densities $q$ for a polygon of $N$ vertices is the complete set of harmonic modes $\{0, \dots, N-1\}$. However, the optimization landscape inherently acts as a geometric filter. To achieve a non-degenerate 6-DOF wrench ellipsoid, the orientation of adjacent propellers along the perimeter must exhibit sufficient angular spread. Modes that fail to provide this spread are systematically rejected:

\begin{enumerate}
    \item \textbf{Degenerate Mode} ($q=0$): 
    This mode corresponds to all rotors having identical phase angles ($\Delta\phi = 0$). Geometrically, this is analogous to the staves of a barrel (if $\theta=\pi/2$) or sticks lying flat on a table (if $\theta=0$). In these configurations, the lines of force are parallel or coplanar in a way that renders the system rank-deficient, possessing zero controllability in specific spatial degrees of freedom.

    \item \textbf{Unary and Binary Modes} ($q \in \{1, 2\}$ and their symmetric inverses $q \in \{N-1, N-2\}$): 
    \firstrev{The unary mode ($q=1$) implies a minimal phase step of $\Delta\phi = \pi/N$. While not strictly mathematically singular, the angular separation between adjacent control vectors is too small to form a robust 6D basis. This creates a loosely woven structure where the force basis vectors are highly correlated, yielding an anisotropic manipulability ellipsoid. Similarly, the binary mode ($q=2$) doubles this step to $2\pi/N$, but empirical data confirms this remains below the critical threshold required for isotropic spatial conditioning. The optimizer consistently discards these 4 modes because they fail to spherically span the wrench space.}

    \item \textbf{Star Modes} ($q \in \mathcal{Q}_{valid}$, defined in~\eqref{eq:q_valid}): 
    The surviving branches are those where the winding number $q$ produces a phase step closer to orthogonality ($\Delta\phi \approx \pi/2$). This is analogous to a ``woven basket,'' where the lines of force cross each other at steep angles. This high-frequency interlocking provides the necessary angular spread to maximize the volume of the wrench polytope, ensuring robust isotropic force closure.
\end{enumerate}

\firstrev{Consequently, the set of \emph{admissible} densities corresponds to the integers strictly between the binary modes and their symmetric inverses:
\begin{equation}
    \mathcal{Q}_{valid} = \{ q \in \mathbb{Z} \mid 3 \le q \le N-3 \}
    \label{eq:q_valid}
\end{equation}
The cardinality of this set, which defines the total number of optimal topological branches $K$, is the count of integers in the closed interval $[3, N-3]$:
\begin{equation}
    K = (N - 3) - 3 + 1 = N - 5
\end{equation}

This fundamental counting aligns perfectly with the physical requirements of full actuation. A minimum of $N=6$ rotors is required to span a non-degenerate 6D wrench ellipsoid. Correspondingly, for $N=6$, the formula yields exactly $6-5 = 1$ valid branch ($q=3$). 

For every additional rotor added to the perimeter ($N > 6$), the system gains one degree of design redundancy. This redundancy is not expressed by enlarging the continuous dimension of the optimal valley, but rather by the emergence of a completely new discrete harmonic mode (a new closed curve in the phase space). Thus, the $N-5$ law is the direct consequence of the continuous spatial spanning requirements filtering out exactly 5 insufficient harmonic modes, leaving the remaining discrete star polygons to define the valid structural symmetries of the platform.}

\section{Practical Implications: Design Redundancy and Synthesis Strategies}
\label{sec:implications}

\subsection{Design Redundancy vs. Kinematic Redundancy}

The topological classification established in Table~\ref{tab:unified_classification} reveals a fundamental, yet previously unexplored, property of multi-rotor systems. We propose to distinguish this property from standard redundancy as follows:

\begin{itemize}
    \item \textbf{Kinematic Redundancy (Operational):} The existence of extra degrees of freedom in the \emph{actuation space} ($N > 6$), allowing the controller to reconfigure the robot's pose without disturbing the end-effector task (dynamic nullspace).
    
    \item \textbf{Design Redundancy (Morphological):} The existence of continuous degrees of freedom in the \emph{parameter space} ($\dim(\mathcal{S}^*) > 0$), allowing the engineer to reconfigure the robot's \emph{structure} without disturbing the platform's optimality (static design nullspace).
\end{itemize}

Standard optimization usually results in a single global minimum, locking the design into a rigid configuration (Type I, \textbf{Design-Rigid}). However, our results demonstrate that highly symmetric chassis (Type IV) defy this expectation. They exhibit \textbf{Design Redundancy}: the global optimum is not a point or a set of sparse discrete points, but a continuous one-dimensional manifold composed of one or more closed branches. 

This implies that for Type IV-B chassis (and IV-C by our conjecture), optimality is not a specific shape, but a continuous family of shapes. This provides a rigorous theoretical foundation for \emph{Morphing MRAVs}: robots that can continuously reorient their rotors in flight--moving along the valleys of the solution manifold--to dynamically alter their geometric profile (e.g., to squeeze through narrow gaps or minimize aerodynamic drag) while maintaining maximum isotropic authority. 
\secondrev{Furthermore, the existence of $N-5$ distinct topological isomers—whether as strictly disconnected branches for $N \le 7$ or as core backbones for $N \ge 8$—suggests a layer of topological fault tolerance: these symmetric configurations may serve as alternative optimal operating modes, allowing the system to shift configurations if a mechanical failure obstructs a specific structural state.}

\subsection{Generalization of Existing Designs}

The topological framework presented here reinterprets isolated designs found in the literature as subsets of a broader solution space. A prominent example is the specific isotropic geometry derived for the \emph{Omnirotor} ($N=8$) by~
\cite{Brescianini2018}. 

Our analysis identifies their algebraic solution not as a unique global optimum, but as a single coordinate point ($\theta_1 \approx 35^\circ$) located on \emph{Branch 1} of the continuous solution manifold for the cube chassis, cf Tab.~\ref{tab:estimated_phase_offsets}. 
While the original work treated this configuration as a static fixed point, our results prove it resides on a \secondrev{1D topological backbone within a broader multidimensional optimal hypersurface}. This implies that even established designs such as the Omnirotor possess latent \emph{Design Redundancy}: they can theoretically be morphed along the identified solution trajectory—and into the surrounding flattened landscape—without compromising their isotropic wrench authority.

\subsection{The Tangent Heuristic for Efficient Design}
\label{subsec:tangent_heuristic}

The empirical observation that global optima for Type III and Type IV chassis lie exclusively (or asymptotically) within the Tangent Torus $\tangentTorus$ suggests a powerful computational strategy. We propose the \textbf{Tangent Heuristic} as a standard initialization for high-DOF MRAV design:

\begin{enumerate}
    \item \textbf{Dimensional Collapse ($2N \to N$):} 
    By constraining the search domain to the torus $\tangentTorus$ rather than the full projective space $(\RPtwo)^N$, the optimization problem is effectively halved in dimensionality. This mitigates the "curse of dimensionality" for large $N$, with the potential to transform an intractable global search into a manageable local optimization.
    
    \item \textbf{The ``Tangent Prior'':} 
    For irregular or quasi-regular chassis (Type III) where the solution is not strictly tangential, the Tangent Torus serves as an extremely high-fidelity \emph{warm start}. Initializing the solver on $\tangentTorus$ bypasses the vast majority of local minima associated with non-tangential configurations.
    
    \item \textbf{Decoupled Tolerance Analysis:} 
The torus formulation allows designers to separate manufacturing tolerances into two orthogonal components: \emph{Phase Errors} (perturbations along $\tangentTorus$) and \emph{Tilt Errors} (perturbations normal to $\tangentTorus$). 
Since the solution manifold is flat along the phase direction (due to the $N-5$ redundancy), the design is naturally robust to phase misalignments. Conversely, sensitivity is concentrated in the normal direction. This informs the mechanical design: high-precision machining is required for the arm inclination (to ensure tangency), while the rotational alignment of the motor mounts can be more relaxed.
\end{enumerate}

\firstrev{It is important to emphasize the practical boundaries of this dimensionality reduction for engineers working in airframe optimization. 
\begin{enumerate}
    \item The collapse from $2N$ to $N$ dimensions characterizes the specific subspace containing the global optima for the Log-Volume cost function; it does \textit{not} encompass all possible wrench polytopes attainable in the full $2N$ space. If a designer restricts their optimization search exclusively to the Tangent Torus $\tangentTorus$, they are actively filtering the search space to find only highly isotropic, omnidirectional architectures. Consequently, this $N$-dimensional representation deliberately leaves highly anisotropic designs--such as standard planar multirotors optimized for unidirectional payload capacity--out of the optimization loop. 

    \item While the mathematical reduction assumes the $(\mathbb{R}P^2)^N$ bidirectional manifold, this continuous $N$-dimensional landscape \secondrev{provides the theoretical foundation to reconstruct the discrete design envelopes} for unidirectional systems via Minkowski reconstruction, as \secondrev{discussed} in Section~\ref{sec:unidirectional_extension}.
 
    \item While the $N-5$ distinct 1D continuous valleys \secondrev{(and their surrounding flattened hypersurfaces for $N \ge 8$)} represent strictly invariant geometric control authority, they serve as theoretical baselines.
    In physical dynamic reconfiguration (morphing) scenarios, complex aerodynamic interferences--such as rotor downwash overlapping, inflow variations, and drag torque--will inevitably alter the system's true energetic efficiency. During high-speed or high-load morphing, the actual minimum-energy configuration may shift slightly off the geometrically optimal 1D manifold.

\end{enumerate}
}

\firstrev{

\subsection{Applicability to Unidirectional Thrust and Gravity Compensation}
\label{sec:unidirectional_extension}

The geometric framework and the associated design manifold $\mathcal{M} = (\mathbb{R}P^2)^N$ derived in Section~\ref{subsec:proj_design_manifold} inherently assume bidirectional thrust capabilities ($\bm{u} \in [-1, 1]^N$, normalized for simplicity). In industrial applications, MRAVs predominantly utilize unidirectional rotors ($\bm{u} \in [0, 1]^N$). This hardware limitation breaks the origin symmetry of the actuator space and introduces a strict requirement for constant gravity compensation which must be contained in the highly reduced workspace. A critical theoretical question is whether these operational constraints invalidate the optimization of the Log-Volume potential $J_{\text{vol}}$  and the topological dimensionality reduction.

We demonstrate that the continuous geometric optimization of the dimensionless grasp matrix $\bar{\bm{A}}$ is mathematically decoupled from the discrete constraints of unidirectional thrust and gravity. This decoupling is established through two geometric properties: \emph{Volume Equivalence} and \emph{Minkowski Reconstruction}.

\subsubsection{Volume Equivalence}
The unidirectional control input space $\mathcal{U}_{uni} = [0, 1]^N$ is an exact affine transformation of the bidirectional space $\mathcal{U}_{bi} = [-1, 1]^N$, defined by $\bm{u}_{uni} = \frac{1}{2} \bm{u}_{bi} + \frac{1}{2}\mathbf{1}_N$. Applying the linear mapping of the dimensionless grasp matrix $\bar{\bm{A}}$, the resulting dimensionless unidirectional wrench space $\mathcal{W}_{uni}$ is a half-scale geometric clone of the bidirectional space $\mathcal{W}_{bi}$, shifted by a constant offset vector $\bm{w}_0$:
\begin{equation}
\mathcal{W}_{uni} = \bar{\bm{A}} \left( \frac{1}{2} \mathcal{U}_{bi} + \frac{1}{2} \mathbf{1}*N \right) = \frac{1}{2} \mathcal{W}_{bi} + \bm{w}*0.
\end{equation}
Because rigid translations in $\mathbb{R}^6$ do not alter hypervolume, the volume of the unidirectional wrench polytope scales by a strict, constant factor: $\text{Vol}(\mathcal{W}_{uni}) = 2^{-6} \text{Vol}(\mathcal{W}_{bi})$.

Furthermore, the cost function $J_{\text{vol}}$ acts on the Grasp Gramian $\det(\bar{\bm{A}}\bar{\bm{A}}^\top)$, which characterizes the volume of the inscribed ellipsoid. By the Cauchy-Binet theorem, $\det(\bar{\bm{A}}\bar{\bm{A}}^\top)$ is exactly the sum of the squares of all $6 \times 6$ subdeterminants of $\bar{\bm{A}}$. Geometrically, the exact volume of the fully realized wrench zonotope (the parallelotope formed by the columns of $\bar{\bm{A}}$) is the sum of the absolute values of those exact same $6 \times 6$ subdeterminants. \secondrev{Because maximizing the sum of squared subdeterminants is not analytically identical to maximizing the sum of their absolute values, the cost function $J_{\text{vol}}$ does not directly maximize the exact true zonotope volume. Instead, minimizing $J_{\text{vol}}$ acts as a mathematically smooth, differentiable proxy that maximizes the bounding surrogate hyper-ellipsoid. This geometric proxy strongly correlates with and robustly expands the operational capabilities of both the bidirectional and unidirectional polytopes.}

\subsubsection{Gravity Compensation via Minkowski Reconstruction}
To hover at a generic attitude, the required gravity compensation wrench $\bm{w}_g$ (mapped to the body frame $\mathcal{F}_B$) must strictly lie within the available wrench space. While a single unidirectional configuration $\mathcal{W}_{uni}$ is shifted and may not contain $\bm{w}_g$ for all $SO(3)$ orientations, the physical mounting of the chassis $\mathcal{C}$ offers discrete flexibility.

By physically inverting a subset of the rotors (mounting them upside down while preserving their line of action $\mathcal{D}_i \in \mathbb{R}P^2$), we effectively change the domain of those specific inputs from $[0, 1]$ to $[-1, 0]$. There are $2^N$ possible physical mounting configurations, corresponding to all sign permutations of the columns of $\bar{\bm{A}}$. Because the union of all $2^N$ one-sided domains perfectly reconstructs the global bidirectional domain (i.e., $[-1, 0] \cup [0, 1] = [-1, 1]$), the union of the $2^N$ resulting shifted unidirectional polytopes perfectly reconstructs the original bidirectional polytope:
\begin{equation}
\bigcup_{p=1}^{2^N} \mathcal{W}_{uni}^{(p)} = \mathcal{W}_{bi}.
\end{equation}
Consequently, if the optimally designed bidirectional space $\mathcal{W}_{bi}$ is large enough to contain the gravity vector $\bm{w}_g$ for any required orientation, there strictly exists at least one discrete rotor mounting configuration $p \in \{1, \dots, 2^N\}$ such that $\bm{w}_g \in \mathcal{W}_{uni}^{(p)}$.

Thus we can conclude with the following statement.
\secondrev{\begin{remark}[Relevance to Unidirectional ESCs]
The optimal geometric allocation determined on $\mathcal{M}$ establishes a theoretical upper bound that strongly informs unidirectional hardware design. The continuous geometric optimum is identified first, and specific unidirectional rotor polarities are assigned discretely post-optimization to encompass the desired operational gravity envelope. However, it must be noted that a fixed physical multirotor utilizing unidirectional ESCs cannot execute the exact same continuous, full-space dynamic morphing trajectories as a bidirectional platform without encountering unilateral actuator saturation in certain spatial directions.
\end{remark}}

}

\firstrev{

\section{Application Scenario: Dynamic Reconfiguration and Aerodynamic Footprinting}
\label{sec:dynamic_reconfiguration}

Conventional underactuated multirotor platforms generate a predominantly vertical aerodynamic downwash, coupling flight mechanics directly to the environmental wake. In fully actuated omnidirectional platforms, the skewed rotor geometry ($\mathbf{d}_i \in \SphereTwo$) intrinsically produces multidirectional flow fields. While often treated as a secondary aerodynamic effect, this spatially distributed wake can be actively modulated for functional surface interactions. 

Directing oblique airflow across surfaces has practical utility in several aerial operations. Examples include the dispersion of surface treatments (e.g., drying agents or chemical solvents) and the pneumatic clearance of debris prior to structural inspections. Furthermore, dynamically morphing the aerodynamic footprint can facilitate tasks such as localized agricultural dispersion, active thermal regulation of industrial components, or targeted gas sampling, all without requiring lateral vehicle translation. By exploiting the inherent design redundancy of the platform, it is possible to generate a continuously varying aerodynamic wake while remaining entirely within the design null space of \eqref{eq:optimization}. This allows for the alteration of environmental interactions without perturbing the internal flight state (e.g., control authority or the $L_2$ norm of the propeller thrusts).

It should be noted that this target area sweeping scenario is provided as a demonstrative proof-of-concept rather than an exhaustive optimization study. The objective is to illustrate how the geometric redundancies formalized by the Projective Design Manifold can be leveraged for functional operations beyond primary flight mechanics. Other potential applications of this active footprinting include adapting the rotor geometry to environmental spatial constraints, vectoring inflow for localized onboard sensor sampling, or rapidly reconfiguring the platform to generate aerodynamic drag moments for braking. Ultimately, the formal characterization of $\tangentTorus$ provides a mathematically rigorous foundation for developing functional self-motions in fully actuated aerial systems.

\subsection{Morphing Simulation on the Tangent Torus}

To illustrate the practical utility of this continuous design null space, we simulate a dynamic reconfiguration scenario for a fully actuated hexarotor ($N=6$) operating within the Tangent Torus $\tangentTorus$. The morphing trajectory is parameterized by a single continuous degree of freedom, $\lambda \in [0, \pi)$, which governs the rotor lines of action according to the affine phase-locking law $\theta_i(\lambda) = \lambda + \delta_i$. 

During the maneuver, the vehicle is commanded to maintain a constant horizontal attitude ($\bm{\omega}_B = \mathbf{0}$) while tracking a sinusoidal position profile along the inertial $X$-axis: $\mathbf{p}_B(t) = [A \cos(\omega t), 0, 0]^\top$, where $A$ is the motion amplitude and $\omega$ is the angular frequency. 
According to the Newton-Euler formalism defined in \eqref{eq:dynamics}, tracking this trajectory requires a time-varying lateral inertial force $m \ddot{\mathbf{p}}_B(t)$. The instantaneous required wrench $\bm{w}_{req}(t)$ to maintain this trajectory alongside gravity compensation is:
\begin{equation}
    \bm{w}_{req}(t) = \begin{bmatrix} -m A \omega^2 \cos(\omega t) & 0 & mg & 0 & 0 & 0 \end{bmatrix}^\top
\end{equation}
For any morphing configuration $\lambda$, the control allocation is resolved by computing the Moore-Penrose pseudoinverse of the instantaneous grasp matrix, $\bm{u}(\lambda, t) = \bm{A}(\lambda)^\dagger \bm{w}_{req}(t)$, yielding the required individual signed thrusts $u_i \in \mathbb{R}$ for each rotor.

\begin{figure*}[t]
  \centering
  \begin{subfigure}[b]{0.245\linewidth}
    \centering
    \includegraphics[width=\linewidth]{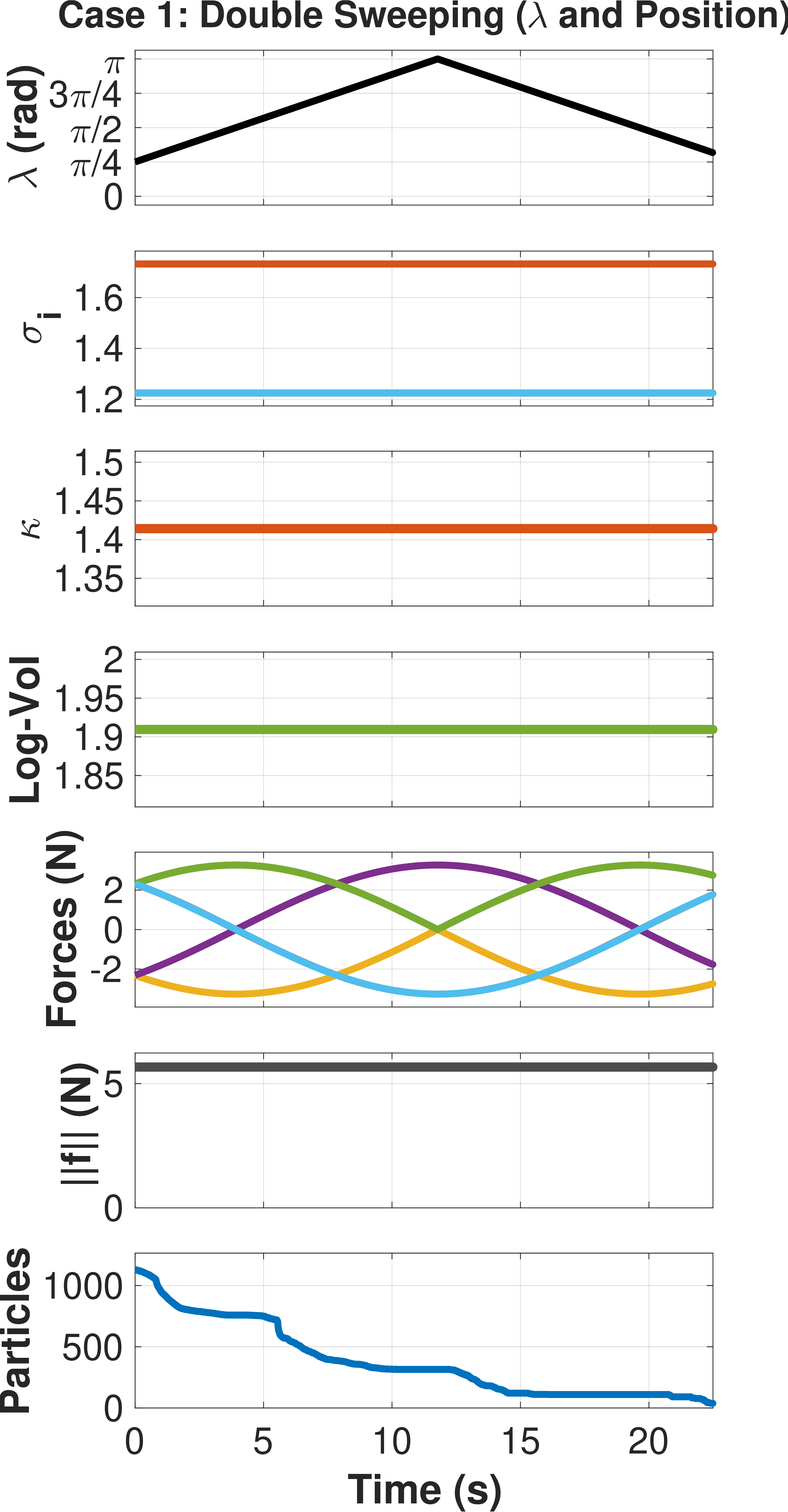}
    \caption{Case 1: both $\lambda$ and position sweep}
    \label{fig:log_case_1}
  \end{subfigure}\hfill
  \begin{subfigure}[b]{0.245\textwidth}
    \centering
    \includegraphics[width=\linewidth]{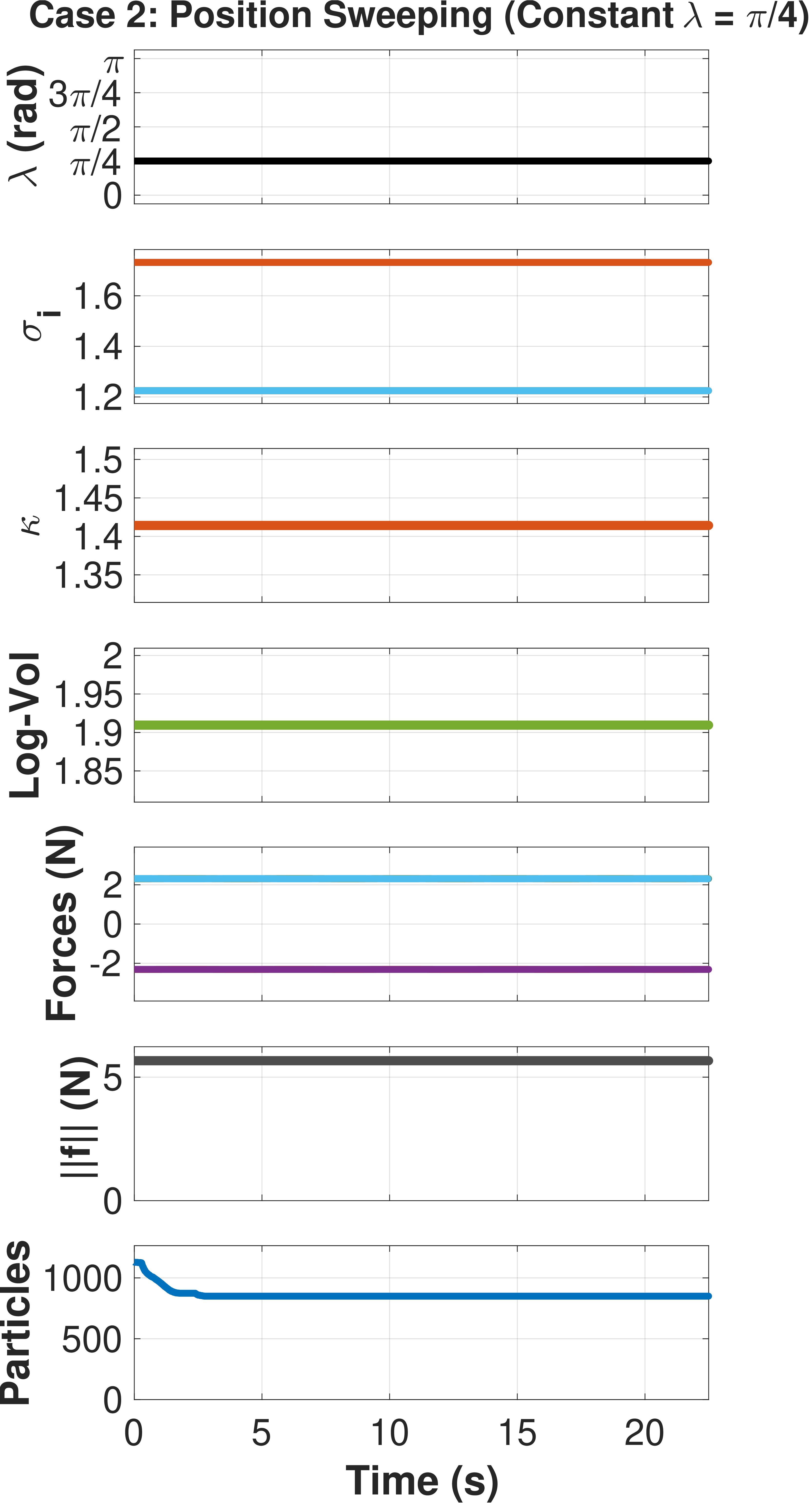}
    \caption{Case 2: positional sweep and constant $\lambda=\pi/4$}
    \label{fig:log_case_2}
  \end{subfigure}\hfill
  \begin{subfigure}[b]{0.245\textwidth}
    \centering
    \includegraphics[width=\linewidth]{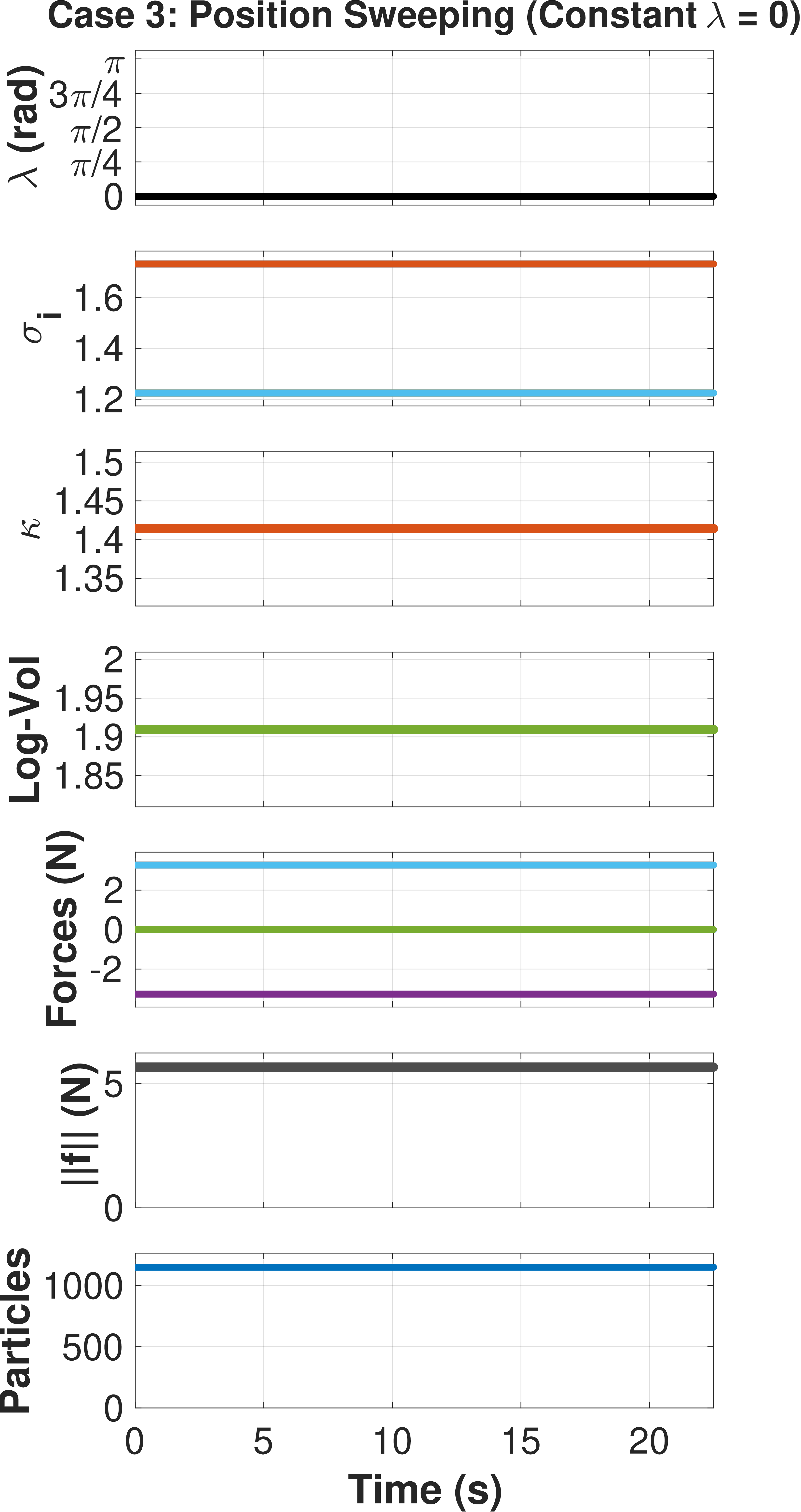}
    \caption{Case 3: positional sweep and constant $\lambda=0$}
    \label{fig:log_case_3}
  \end{subfigure}\hfill
  \begin{subfigure}[b]{0.245\textwidth}
    \centering
    \includegraphics[width=\linewidth]{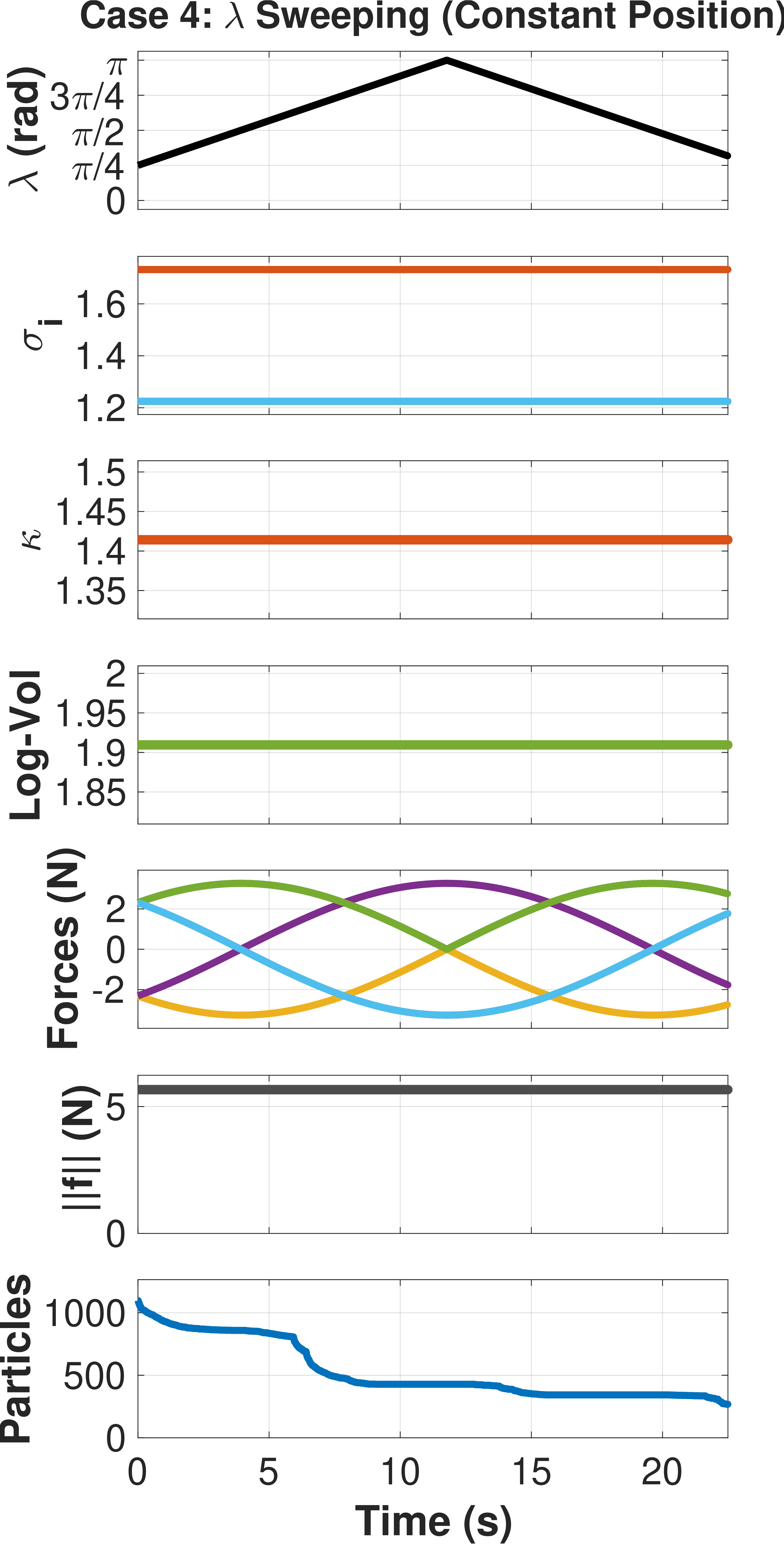}
    \caption{Case 4: $\lambda$ sweep and constant position}
    \label{fig:log_case_4}
  \end{subfigure}

  \vspace{3ex} 

  \begin{subfigure}[b]{0.245\textwidth}
    \centering
    \includegraphics[width=\linewidth, height=0.6\linewidth]{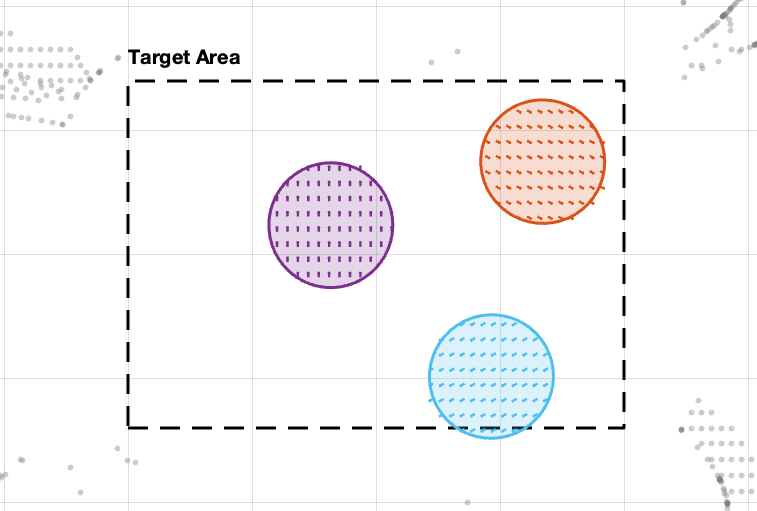} 
    \caption{Case 1: Final Distribution}
    \label{fig:visual_case_1}
  \end{subfigure}\hfill
  \begin{subfigure}[b]{0.245\textwidth}
    \centering
    \includegraphics[width=\linewidth, height=0.6\linewidth]{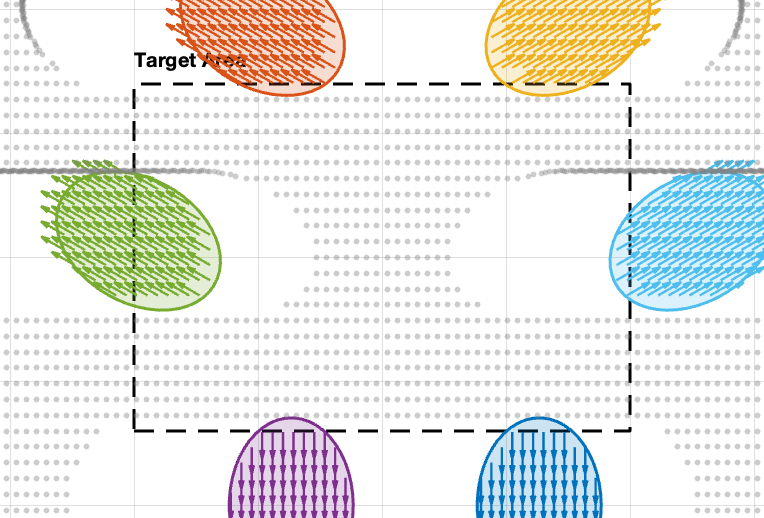} 
    \caption{Case 2: Final Distribution}
    \label{fig:visual_case_2}
  \end{subfigure}\hfill
  \begin{subfigure}[b]{0.245\textwidth}
    \centering
    \includegraphics[width=\linewidth, height=0.6\linewidth]{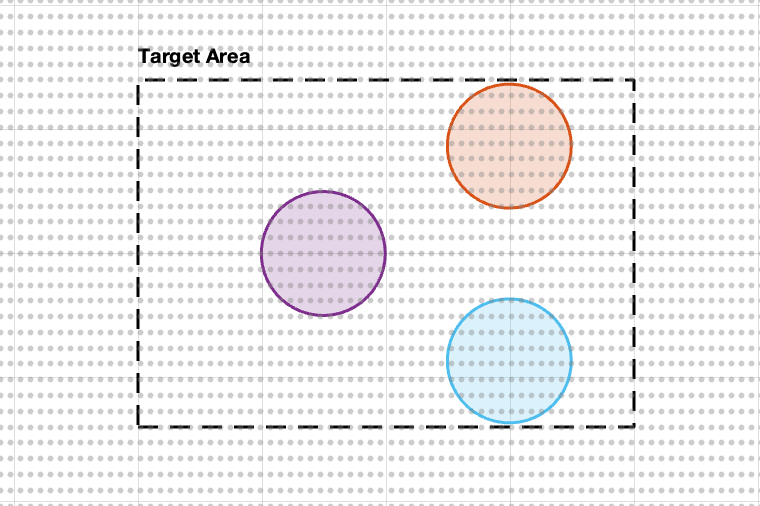} 
    \caption{Case 3: Final Distribution}
    \label{fig:visual_case_3}
  \end{subfigure}\hfill
  \begin{subfigure}[b]{0.245\textwidth}
    \centering
    \includegraphics[width=\linewidth, height=0.6\linewidth]{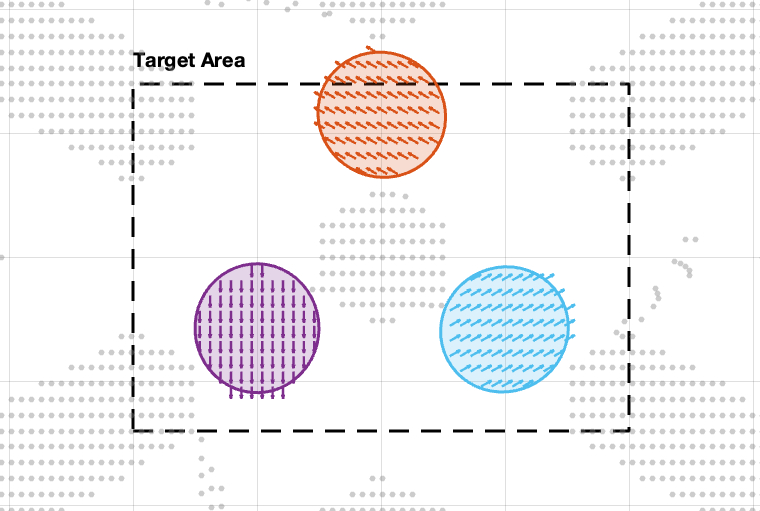} 
    \caption{Case 4: Final Distribution}
    \label{fig:visual_case_4}
  \end{subfigure}

   \vspace{1ex} 

  \caption{\firstrev{\textbf{Comparative results of the active aerodynamic footprinting simulation for target area sweeping.}
  \textbf{Row 1:} Synchronized timeseries of the morphing parameter $\lambda$, individual signed thrusts $\bm{u}$, total control effort $\|\bm{u}\|$, and the particle count within the monitoring area. Y-axes are uniform across cases to facilitate direct comparison.
  \textbf{Row 2:} Visual representation of the ground plane ($z=-H$) at the conclusion of each simulation, illustrating the final particle distribution (grey dots) and the cumulative sweeping path. The bounded rectangle denotes the target monitoring area.
  \textbf{Analysis:} Case 1 (Double Sweep) exhibits the highest clearance rate, displacing $>95\%$ of the particles. Case 2 ($\lambda=\pi/4$) clears approximately $25\%$. Case 3 ($\lambda=0$, corresponding to standard vertical thrust with respect to the sweeping plane) clears negligible particles, illustrating the inefficiency of vertical flows for lateral advection. Case 4 ($\lambda$-sweep from a stationary hover) clears approximately $66\%$. The total control effort $\|\bm{u}\|$ remains invariant with respect to $\lambda$ during static trajectory phases, confirming the null-space isolation of the maneuver.}}
  \label{fig:campaign_results}
\end{figure*}

\subsection{Aerodynamic Wake and Particle Advection Model}

To evaluate the environmental impact of this reconfiguration, the aerodynamic downwash of the platform is projected onto a horizontal ground plane located at a depth $z = -H$. Utilizing standard actuator disk momentum theory, the aerodynamic wake of the $i$-th rotor is modeled as a cylindrical streamtube originating at the rotor position $\mathbf{p}_i$ and oriented along the thrust-opposing unit vector $\mathbf{v}_{w,i}(\lambda) = -\operatorname{sgn}(u_i) \mathbf{d}_i(\lambda)$. The magnitude of the induced velocity within the streamtube is proportional to the square root of the absolute control input, $V_i \propto \sqrt{|u_i(\lambda, t)|}$.

The intersection of each streamtube cylinder with the horizontal ground plane generates an elliptical footprint $\mathcal{E}_i(\lambda)$. Within each domain $\mathcal{E}_i$, we compute the 2D projection of the induced velocity vector field. To quantify the surface interaction footprint, a massless particle transport model is introduced on the ground plane. An ensemble of particles is initialized with a uniform spatial distribution $\mathcal{P}_0 = \{\mathbf{x}_k(0) \in \mathbb{R}^2\}$. The kinematics of the $k$-th particle are governed by a first-order advection model, where the particle velocity is proportional to the superposition of the projected aerodynamic induced velocities from all intersecting rotor streamtubes:
\begin{equation}
    \dot{\mathbf{x}}_k(t) = c_v \sum_{i=1}^{N} \mathcal{I}_i(\mathbf{x}_k, \lambda) \sqrt{|u_i(\lambda, t)|} \, \mathbf{v}_{w,i}^{xy}(\lambda)
\end{equation}
Here, $c_v \in \mathbb{R}^+$ is a scaling constant, $\mathbf{v}_{w,i}^{xy}$ represents the planar projection of the $i$-th rotor's wake vector, and $\mathcal{I}_i$ is an indicator function that equals $1$ if the spatial position $\mathbf{x}_k$ resides within the projected elliptical boundary $\mathcal{E}_i(\lambda)$, and $0$ otherwise. 

As the chassis morphs continuously with respect to $\lambda$, the control allocation implicitly redistributes the individual thrust efforts. This geometric variation manifests as dynamic translations and deformations of the elliptical footprints, accompanied by proportional variations in the projected flow intensities. The time-varying vector field subsequently redistributes the particle ensemble, actively sweeping the target area.

\subsection{Task Comparison and Null-Space Morphing Invariance}

Figure~\ref{fig:campaign_results} provides a comparative analysis of  four canonical sweeping strategies. The timeseries data and final visual distributions demonstrate the advantage of using the motion in the design nullspace. Case 3, which mimics a standard multirotor generating vertical thrust relative to the sweep plane ($\theta_i=0$), results in zero particle advection, confirming its inefficiency for lateral surface-interaction tasks. Conversely, static oblique flow ($\lambda=\pi/4$, Case 2) clears approximately $25\%$ of the target area, while dynamically sweeping $\lambda$ from a stationary hover (Case 4) increases clearance to $33\%$. Combining dynamic lateral translation with continuous reconfiguration (Case 1) results in near-complete clearance ($>95\%$), demonstrating that active downwash vectoring significantly outperforms static configurations for this specific task. 

To empirically validate that this morphological transition occurs independently of the vehicle's flight authority, we monitor the total $L_2$ norm of the control input vector, $\|\bm{u}\|$, throughout the simulation. For a specified trajectory wrench $\bm{w}_{req}(t)$, the pseudoinverse allocation yields a baseline control effort defined by $\|\bm{u}\|^2 = \bm{w}_{req}^\top (\bm{A}\bm{A}^\top)^{-1} \bm{w}_{req}$. Because the continuous $(N-5)$-dimensional manifolds represent configurations where the condition number and the orthogonal constraint stiffness are mathematically invariant, the matrix product $\bm{A}\bm{A}^\top$ remains strictly constant with respect to $\lambda$. Consequently, as the parameter $\lambda$ is swept to vector the aerodynamic wake, the individual rotor inputs $u_i(\lambda, t)$ dynamically redistribute to maintain Newton-Euler equilibrium, yet the total omnidirectional control effort $\|\bm{u}\|$ remains entirely decoupled from the internal morphological states. The third subplots of Fig.~\ref{fig:campaign_results} corroborate this theoretical null-space invariance, demonstrating that $\|\bm{u}\|$ remains practically invariant during the maneuvers. This confirms that the affine phase-locking paths act as zero-energy self-motions, enabling extensive environmental interaction without compromising platform efficiency.

}

\section{Reproducibility and Software Package}
\label{sec:reproducibility}

\subsection{The OmniDesign Optimizer Suite}
\label{subsec:software}

To ensure full reproducibility of the topological scaling laws and optimization results presented in this work, we have released the complete \textbf{OmniDesign Optimizer} software suite as an open-source MATLAB framework. The package is designed to allow independent verification of the \emph{N-5 Law} via the branch and phase extraction algorithm, the \emph{tangent torus projection}, and the generation of globally optimal chassis geometries.

The software suite is organized into \firstrev{six} primary modules mirroring the analytical steps of this paper:
\begin{enumerate}
    \item \textbf{Global Optimizer}: Implements the global manifold exhaustion strategy on $(\mathbb{RP}^2)^N$ to generate the raw point clouds of optimal configurations.
    \item \textbf{Manifold Parameterization}: Validates the reduction of the solution space to the tangent torus $\tangentTorus$. It fits semi-ellipsoidal trajectories to the raw optimization data to extract the intrinsic angular coordinates.
    \item \textbf{Topological Analysis}: Performs the high-dimensional PCA and clustering required to isolate the $K$ distinct isomer branches. It quantitatively validates the scaling law by computing the "Cluster Spread" and identifying disconnected topological loops.
    \item \textbf{Performance Benchmarking}: A dedicated benchmarking script that reproduces the comparative analysis presented. It performs the ablation study (Log-Volume vs. Condition Number) and automatically generates the timing data for Table~\ref{tab:comp_time} and the landscape visualizations for Figure~\ref{fig:ablation_comparison}.
    \item \textbf{Manifold Motion \firstrev{and Dynamic Simulation}}: \firstrev{A suite of tools} that reconstructs the 3D continuous motion of the thrust vectors along the identified optimal branches. It visually validates the "zero-cost motion" property by demonstrating the invariance of the singular values along the trajectory\firstrev{, and features an interactive simulation to empirically prove control effort invariance and visualize aerodynamic wake sweeping}.
    \firstrev{\item \textbf{Hessian Expansion Study}: Evaluates the Subspace Alignment and local numerical Hessian along the 1D branches, mapping the topological expansion of the flat multi-dimensional valleys in High-$N$ configurations.}
\end{enumerate}

\noindent \textbf{Availability:}
The source code, along with the datasets for the $N=6$ to $N=20$ chassis analyzed in this study, is available at:
\begin{center}
    \url{https://github.com/antoniofranchi/omnidesign-optimizer/}
\end{center}

\subsection{Multimedia Extensions: Dynamic Manifold Visualization}
\label{subsec:multimedia}

To provide intuitive validation of the topological findings, we include \firstrev{three} supplementary video animations demonstrating the \emph{Cube} ($N=8$)\firstrev{,} \emph{Regular Octagon} ($N=8$)\firstrev{, and \emph{Hexarotor} ($N=6$)} chassis in motion. These visualizations explicitly verify the existence of the \emph{Design Nullspace} by simulating the platform's behavior as it traverses the solution manifold over time.

The animations visualize the system state parameterized by the global phase variable $\lambda(t)$, which varies continuously from $0$ to $\pi$. The visual layout is divided into two synchronized panels:

\subsubsection*{Top Panel: Physical Reconfiguration}
We display three simultaneous instances of the platform corresponding to the $K=3$ distinct topological branches (isomers) identified by the $N-5$ Law for $N=8$. As $\lambda(t)$ evolves, the rotor force lines rotate within the Tangent Torus $\tangentTorus$. Crucially, the animation demonstrates \emph{Affine Phase Locking}: the rotors reorient in perfect synchrony, maintaining the specific constant phase offsets $\delta_i^{(k)}$ characteristic of each branch (as derived in Table~\ref{tab:estimated_phase_offsets}). This visualizes the infinite set of optimal points available in the solution landscape.

\subsubsection*{Bottom Panel: Wrench Space Metrics}
Below the physical models, we visualize the corresponding control authority metrics in real-time:
\begin{enumerate}
    \item \textbf{Wrench Ellipsoids:} We render the Force and Moment manipulability ellipsoids, which represent the 3D projections of the full 6-dimensional singular value hyper-ellipsoid into the translational ($\mathbb{R}^3_f$) and rotational ($\mathbb{R}^3_\tau$) subspaces.
    \item \textbf{Singular Value Spectrum:} We plot the bars of the six singular values $\sigma_1 \dots \sigma_6$ of the dimensionless grasp matrix $\bar{\bm{A}}(\lambda)$.
    \item \textbf{Optimality Scalar Fields:} We track the Condition Number $\kappa$ and the (non-negated) Log-Volume metric $V = \sum \ln(\sigma_i)$, which serves as the maximization objective.
\end{enumerate}

\subsubsection*{Comparative Analysis and Invariance}
The animations reveal a fundamental property of the Type IV solution manifolds: \textbf{Isotropic Invariance}. For the three optimal branches, the singular values $\sigma_i$ remain perfectly constant and equal (indicating optimal isotropy) throughout the entire trajectory of $\lambda$. Consequently, the force and moment ellipsoids maintain a fixed spherical shape and volume, despite the continuous rotation of the physical actuators. This confirms that the identified topological loops correspond to \emph{minimum-cost morphing trajectories}.

For contrast, the animations include three control groups: (i) a configuration constrained to the Tangent Torus $\tangentTorus$ but with incorrect phase offsets (decoherent phase), and (ii-iii) two configurations with random rotor orientations. In these sub-optimal cases, the visualizations show severe degradation:
\begin{itemize}
    \item The singular values, whether constant or oscillating, are significantly far from the optimum.
    \item The Condition Number $\kappa$ and the (non-negated) Log-Volume $V$ increase.
    \item The manipulability ellipsoids undergo dynamic deformation (breathing/shearing), indicating time-varying loss of control authority in specific directions.
\end{itemize}
These comparisons highlight also that tangency alone is insufficient; strict adherence to the extracted affine phase-locking laws is required to achieve global optimality.

\firstrev{\subsubsection*{Dynamic Reconfiguration and Wake Vectoring}
Beyond kinematic validation, the third simulation demonstrates the practical utility of the optimal nullspace using a fully-actuated hexarotor executing a constrained lateral navigation task. As the platform dynamically reconfigures its morphology along the 1D manifold ($\lambda$), the simulation renders the real-time projection of the aerodynamic downwash onto the ground plane. Telemetry from this simulation empirically proves \emph{Control Effort Invariance}: the total $L_2$ norm of the control input vector ($||\bm{f}||$) remains strictly constant during the morphological transition, confirming that traversal of the optimal continuous manifold constitutes true zero-energy self-motion.}

\section{Concluding Remarks}
\label{sec:conclusions}

This work presents a comprehensive topological characterization of the design space for fully actuated and omnidirectional MRAVs. 
By shifting the perspective from local parameter optimization to a global manifold analysis, we revealed a fundamental collapse in the solution landscape: for symmetric chassis, the "optimal design" is rarely an isolated point. \secondrev{For low-complexity symmetric chassis ($N \le 7$), it strictly collapses into a coherent 1-DOF continuous mechanism governed by the underlying geometry, which subsequently expands into multidimensional optimal hypersurfaces anchored by these 1D core backbones for higher-order systems ($N \ge 8$). }
We formalized this structural baseline through the \textit{N-5 Law} (for both polygonal and solid regular chassis) and the star polygon generative isomorphism (for regular polygonal chassis), which together allow designers to predict the core topology and discrete isomer count of the solution space based solely on chassis symmetry, providing a deterministic foundation that eliminates the need for pure stochastic numerical search.

Immediate future work will focus on the rigorous group-theoretic formalization of the Star Polygon Isomorphism for the high-$N$ limit ($N \ge 8$), investigating the exact analytical boundaries of the hypersurface expansion \secondrev{ where the strict 1D phase-locking physically relaxes into flattened higher-dimensional optimal landscapes.} Furthermore, we aim to investigate "Topological Robustness"--specifically, characterizing how these resonant phase-locked states and their surrounding optimal hypersurfaces deform or bifurcate when the chassis geometry deviates from perfect regularity (the transition from Type IV back to Type I).

\firstrev{Looking forward, a highly promising avenue for future research is the formalization of a Pareto optimal configuration space. Such a framework would map the complete geometric continuum bridging purely isotropic architectures (as explored in this work) with highly efficient, unidirectional designs (such as standard planar multirotors). Identifying a generative mapping that contains only non-dominated wrench polytopes would provide a universal, task-agnostic tool for global airframe optimization.}

\section*{Acknowledgments}

The author thanks Chiara Gabellieri for her insightful scientific exchange and constant encouragement throughout the development of this work. Acknowledgment is also due to Fiorella Romano for being an early adopter of the manifold visualization idea and providing feedback that helped me enhance the understandability of technical results for a larger audience.

The author also thanks the anonymous reviewers. Their insightful feedback and constructive criticism genuinely helped to improve the quality of the original work.

The author also acknowledges the LLM Gemini 3 (late 2025 and early 2026) for assistance in proofreading, and writing and optimizing the MATLAB scripts used for data processing, plot generations and proofreading  The author manually verified the functionality of all code and the accuracy of the resulting outputs. The final, validated codebase is available in the supplementary material and the repository provided in Sec.~\ref{sec:reproducibility}.

\section*{Project Funding}

The work was partially funded by the European Commission Horizon Europe Framework under project Autoassess (101120732).

\bibliographystyle{SageH} 
\bibliography{bib-custom}

\begin{thebibliography}{40}
\providecommand{\natexlab}[1]{#1}
\providecommand{\url}[1]{\texttt{#1}}
\providecommand{\urlprefix}{URL }
\expandafter\ifx\csname urlstyle\endcsname\relax
  \providecommand{\doi}[1]{DOI:\discretionary{}{}{}#1}\else
  \providecommand{\doi}{DOI:\discretionary{}{}{}\begingroup
  \urlstyle{rm}\Url}\fi

\bibitem[{Aboudorra et~al.(2024)Aboudorra, Gabellieri, Brantjes, Sablé and
  Franchi}]{Aboudorra2024}
Aboudorra Y, Gabellieri C, Brantjes R, Sablé Q and Franchi A (2024) Modelling,
  analysis, and control of omnimorph: an omnidirectional morphing multi-rotor
  uav.
\newblock \emph{Journal of Intelligent \& Robotic Systems} 110(21).

\bibitem[{Al-zubaidi and Stol(2025)}]{AlZubaidi2025Comparison}
Al-zubaidi S and Stol KA (2025) A comparison of optimised, fixed-tilt,
  fully-actuated multirotor uav configurations.
\newblock In: \emph{16th International Micro Air Vehicle Conference and
  Competition (IMAV)}, IMAV2025-13.

\bibitem[{Arza et~al.(2025)Arza, Rehberg, Weiss, Kulkarni and
  Alexis}]{Arza2025Performance}
Arza E, Rehberg W, Weiss P, Kulkarni M and Alexis K (2025) Performance-guided
  task-specific optimization for multirotor design.
\newblock \emph{arXiv preprint arXiv:2510.04724} .

\bibitem[{Bodie et~al.(2024)Bodie, Brunner and Allenspach}]{Bodie2024Book}
Bodie K, Brunner M and Allenspach M (2024) \emph{Omnidirectional Tilt-Rotor
  Flying Robots for Aerial Physical Interaction: Modelling, Control, Design and
  Experiments}, volume 157.
\newblock Springer Tracts in Advanced Robotics.

\bibitem[{Brescianini and D'Andrea(2016)}]{Brescianini2016ICRA}
Brescianini D and D'Andrea R (2016) Design, modeling and control of an
  omni-directional aerial vehicle.
\newblock In: \emph{2016 IEEE International Conference on Robotics and
  Automation (ICRA)}. pp. 3261--3266.
\newblock \doi{10.1109/ICRA.2016.7487497}.

\bibitem[{Brescianini and D'Andrea(2018)}]{Brescianini2018}
Brescianini D and D'Andrea R (2018) An omni-directional multirotor vehicle.
\newblock \emph{Mechatronics} 55: 76--93.

\bibitem[{Brunner et~al.(2022)Brunner, Giacomini, Siegwart and
  Tognon}]{Brunner2022Tank}
Brunner M, Giacomini L, Siegwart R and Tognon M (2022) Energy tank-based
  policies for robust aerial physical interaction with moving objects.
\newblock In: \emph{2022 International Conference on Robotics and Automation
  (ICRA)}. pp. 2054--2060.
\newblock \doi{10.1109/ICRA46639.2022.9812342}.

\bibitem[{Coxeter(1973)}]{coxeter1973regular}
Coxeter HSM (1973) \emph{Regular Polytopes}.
\newblock 3rd edition. New York: Dover Publications.
\newblock ISBN 978-0486614809.

\bibitem[{Franchi(2019)}]{Franchi2019Interaction}
Franchi A (2019) Interaction control of platforms with multi-directional total
  thrust.
\newblock In: \emph{Aerial Robotic Manipulation}, \emph{Springer Tracts in
  Advanced Robotics}, volume 129. Springer, pp. 175--189.

\bibitem[{Garofano-Soldado et~al.(2024)Garofano-Soldado, Gonzalez-Morgado,
  Heredia and Ollero}]{Garofano-2024}
Garofano-Soldado A, Gonzalez-Morgado A, Heredia G and Ollero A (2024)
  Assessment and modeling of the aerodynamic ground effect of a fully-actuated
  hexarotor with tilted propellers.
\newblock \emph{IEEE Robotics and Automation Letters} 9(2): 1907--1914.
\newblock \doi{10.1109/LRA.2024.3350975}.

\bibitem[{Gonzalez-Morgado et~al.(2025)Gonzalez-Morgado, Nekoo, Heredia and
  Ollero}]{GONZALEZMORGADO2025276}
Gonzalez-Morgado A, Nekoo SR, Heredia G and Ollero A (2025)
  Bidirectional-thruster multirotor for perimeter pipe inspections ({BiMPPI}):
  A nonlinear optimal integral-{SDRE} design.
\newblock \emph{{ISA} Transactions} 161: 276--288.
\newblock \doi{https://doi.org/10.1016/j.isatra.2025.03.015}.

\bibitem[{González-Morgado et~al.(2023)González-Morgado, Álvarez Cía,
  Heredia and Ollero}]{Morgado-2023}
González-Morgado A, Álvarez Cía C, Heredia G and Ollero A (2023)
  Fully-actuated, corner contact aerial robot for inspection of hard-to-reach
  bridge areas.
\newblock In: \emph{2023 Int. Conf. on Unmanned Aircraft Systems (ICUAS)}. pp.
  1191--1198.
\newblock \doi{10.1109/ICUAS57906.2023.10156580}.

\bibitem[{Hamandi et~al.(2024)Hamandi, Ali, Evangeliou, Chaikalis, Tzes,
  Kyriakopoulos and Khorrami}]{Hamandi2024OmniOcta}
Hamandi M, Ali AM, Evangeliou N, Chaikalis D, Tzes A, Kyriakopoulos K and
  Khorrami F (2024) Mechatronic design of an omnidirectional octorotor uav.
\newblock In: \emph{2024 10th International Conference on Automation, Robotics
  and Applications (ICARA)}. pp. 300--304.

\bibitem[{Hamandi et~al.(2021)Hamandi, Usai, Sable, Staub, Tognon and
  Franchi}]{Hamandi2021Taxonomy}
Hamandi M, Usai F, Sable Q, Staub N, Tognon M and Franchi A (2021) Design of
  multirotor aerial vehicles: a taxonomy based on input allocation.
\newblock \emph{The International Journal of Robotics Research (IJRR)} 40(8-9):
  1015--1044.
\newblock \doi{10.1177/02783649211025998}.

\bibitem[{Jiang and Voyles(2013)}]{Jiang2013Dexterous}
Jiang G and Voyles R (2013) Hexrotor {UAV} platform enabling dextrous
  interaction with structures-flight test.
\newblock In: \emph{2013 IEEE International Symposium on Safety, Security, and
  Rescue Robotics (SSRR)}. pp. 1--6.
\newblock \doi{10.1109/SSRR.2013.6719377}.

\bibitem[{Kamel et~al.(2018)Kamel, Verling, Elkhatib, Sprecher, Wulkop, Taylor,
  Siegwart and Gilitschenski}]{Kamel2018Voliro}
Kamel M, Verling S, Elkhatib O, Sprecher C, Wulkop P, Taylor Z, Siegwart R and
  Gilitschenski I (2018) The voliro omniorientational hexacopter: An agile and
  maneuverable tiltable-rotor aerial vehicle.
\newblock \emph{IEEE Robotics \& Automation Magazine} 25(4): 34--44.
\newblock \doi{10.1109/MRA.2018.2866758}.

\bibitem[{Lee et~al.(2025)Lee, Cheng, Wu, Lim, Siegwart and
  Hovakimyan}]{Lee2025Omnirotor}
Lee H, Cheng S, Wu Z, Lim J, Siegwart R and Hovakimyan N (2025) Geometric
  tracking control of omnidirectional multirotors for aggressive maneuvers.
\newblock \emph{IEEE Robotics and Automation Letters} 10(2): 1130--1137.
\newblock \doi{10.1109/LRA.2024.3518922}.

\bibitem[{Lei et~al.(2017)Lei, Ji, Wang, Bai and Xu}]{Lei2017Aerodynamic}
Lei Y, Ji Y, Wang C, Bai Y and Xu Z (2017) Aerodynamic design on the non-planar
  rotor system of a multi-rotor flying robot ({MFR}).
\newblock In: \emph{2017 IEEE 3rd International Symposium in Robotics and
  Manufacturing Automation (ROMA)}. pp. 1--5.
\newblock \doi{10.1109/ROMA.2017.8231740}.

\bibitem[{Lewis and Overton(1996)}]{lewis1996eigenvalue}
Lewis AS and Overton ML (1996) Eigenvalue optimization.
\newblock \emph{Acta Numerica} 5: 149--190.

\bibitem[{Lin et~al.(2025)Lin, Ji, Wu, Wu, Han and Gao}]{Lin-2025}
Lin J, Ji S, Wu Y, Wu T, Han Z and Gao F (2025) Float drone: A fully-actuated
  coaxial aerial robot for close-proximity operations.
\newblock In: \emph{2025 IEEE/RSJ Int. Conf. on Intelligent Robots and
  Systems}. pp. 7216--7223.
\newblock \doi{10.1109/IROS60139.2025.11245967}.

\bibitem[{McCarthy et~al.(2024)McCarthy, Thomas, Danielson, Phillips and
  Fierro}]{McCarthy2024Space}
McCarthy RM, Thomas T, Danielson C, Phillips S and Fierro R (2024) Control for
  an omnidirectional multi-rotor uav for space applications.
\newblock In: \emph{AIAA SciTech Forum}, AIAA 2024-0508.

\bibitem[{Niessen and {Betaflight Dev Team}(2024)}]{DShotStandard}
Niessen F and {Betaflight Dev Team} (2024) Digital shot ({DShot}) {ESC}
  protocol: Bidirectional telemetry standards.
\newblock \urlprefix\url{https://betaflight.com/docs/development/DShot}.
\newblock Industry standard for real-time motor feedback.

\bibitem[{Nigro et~al.(2021)Nigro, Pierri and Caccavale}]{Nigro2021}
Nigro M, Pierri F and Caccavale F (2021) Control of an omnidirectional uav for
  transportation and manipulation tasks.
\newblock \emph{Applied Sciences} 11(22): 10991.
\newblock \doi{10.3390/app112210991}.

\bibitem[{Nikitas(2024)}]{Nikitas2024Thesis}
Nikitas D (2024) \emph{Optimized design of a tilted propeller aerial robot for
  ballast tank contact inspections}.
\newblock Master's Thesis, University of Twente.

\bibitem[{Nikou et~al.(2015)Nikou, Gavridis and
  Kyriakopoulos}]{Nikou2015Mechanical}
Nikou A, Gavridis GC and Kyriakopoulos KJ (2015) Mechanical design, modelling
  and control of a novel aerial manipulator.
\newblock In: \emph{2015 IEEE International Conference on Robotics and
  Automation (ICRA)}. pp. 4698--4703.
\newblock \doi{10.1109/ICRA.2015.7139851}.

\bibitem[{Nocedal and Wright(2006)}]{nocedal2006optimization}
Nocedal J and Wright SJ (2006) \emph{Numerical optimization}.
\newblock Second edition. Springer.

\bibitem[{Ollero et~al.(2022)Ollero, Tognon, Suarez, Lee and
  Franchi}]{AerialRobotic}
Ollero A, Tognon M, Suarez A, Lee D and Franchi A (2022) Past, present, and
  future of aerial robotic manipulators.
\newblock \emph{IEEE Transactions on Robotics} 38(1): 626--645.
\newblock \doi{10.1109/TRO.2021.3084395}.

\bibitem[{Park et~al.(2016)Park, Her, Kim and Lee}]{Park2016ODAR}
Park S, Her J, Kim J and Lee D (2016) Design, modeling and control of
  omni-directional aerial robot.
\newblock In: \emph{2016 IEEE/RSJ International Conference on Intelligent
  Robots and Systems (IROS)}. pp. 1570--1575.
\newblock \doi{10.1109/IROS.2016.7759254}.

\bibitem[{Perin et~al.(2024)Perin, Bertoni, Viezzer, Michieletto and
  Cenedese}]{Perin-2024}
Perin M, Bertoni M, Viezzer N, Michieletto G and Cenedese A (2024) Star-shaped
  tilted hexarotor maneuverability: Analysis of the role of the tilt cant
  angles.
\newblock In: \emph{2024 IEEE 20th Int. Conf. on Automation Science and
  Engineering}. pp. 1791--1797.
\newblock \doi{10.1109/CASE59546.2024.10711709}.

\bibitem[{Rajappa et~al.(2015)Rajappa, Ryll, B\"ulthoff and
  Franchi}]{Rajappa2015Modeling}
Rajappa S, Ryll M, B\"ulthoff HH and Franchi A (2015) Modeling, control and
  design optimization for a fully-actuated hexarotor aerial vehicle with tilted
  propellers.
\newblock In: \emph{Proceedings of the IEEE International Conference on
  Robotics and Automation (ICRA)}. pp. 4006--4013.
\newblock \doi{10.1109/ICRA.2015.7139759}.

\bibitem[{Rashad et~al.(2020)Rashad, Goerres, Aarts, Engelen and
  Stramigioli}]{Rashad2020Review}
Rashad R, Goerres J, Aarts R, Engelen JBC and Stramigioli S (2020) Fully
  actuated multirotor {UAV}s: A literature review.
\newblock \emph{IEEE Robotics \& Automation Magazine} 27(3): 97--107.
\newblock \doi{10.1109/MRA.2019.2955964}.

\bibitem[{Ryll et~al.(2015)Ryll, B\"ulthoff and Giordano}]{Ryll2015Holocopter}
Ryll M, B\"ulthoff HH and Giordano PR (2015) A novel overactuated quadrotor
  unmanned aerial vehicle: Modeling, control, and experimental validation.
\newblock \emph{IEEE Transactions on Control Systems Technology} 23(2):
  540--556.

\bibitem[{Ryll et~al.(2019)Ryll, Muscio, Pierri, Cataldi, Antonelli, Caccavale,
  Bicego and Franchi}]{Ryll2019Paradigm}
Ryll M, Muscio G, Pierri F, Cataldi E, Antonelli G, Caccavale F, Bicego D and
  Franchi A (2019) 6d interaction control with aerial robots: The flying
  end-effector paradigm.
\newblock \emph{The International Journal of Robotics Research} 38(9):
  1045--1062.
\newblock \doi{10.1177/0278364919856694}.

\bibitem[{Skaug(2024)}]{BLHeli32}
Skaug S (2024) {BLHeli\_32} firmware for 32-bit electronic speed controllers.
\newblock \urlprefix\url{https://github.com/bitdump/BLHeli}.
\newblock Widespread firmware for high-performance multirotor ESCs.

\bibitem[{Szász et~al.(2022)Szász, Allenspach, Han, Tognon and
  Katzschmann}]{Szasz2022}
Szász R, Allenspach M, Han M, Tognon M and Katzschmann RK (2022) Modeling and
  control of an omnidirectional micro aerial vehicle equipped with a soft
  robotic arm.
\newblock In: \emph{2022 IEEE 5th International Conference on Soft Robotics
  (RoboSoft)}. pp. 01--08.

\bibitem[{Tognon and Franchi(2018)}]{Tognon2018Omni}
Tognon M and Franchi A (2018) Omnidirectional aerial vehicles with
  unidirectional thrusters: Theory, optimal design, and control.
\newblock \emph{IEEE Robotics and Automation Letters (RA-L)} 3(3): 2277--2282.
\newblock \doi{10.1109/LRA.2018.2802544}.

\bibitem[{Toratani(2012)}]{Toratani2012}
Toratani D (2012) Research and development of doublet tetrahedron
  hexa-rotorcraft ({DOT-HR}).
\newblock In: \emph{Proceedings of the 28th Congress of the International
  Council of the Aeronautical Sciences (ICAS)}. Brisbane, Australia.

\bibitem[{von Frankenberg and Nokleby(2018)}]{Frankenberg2018}
von Frankenberg F and Nokleby SB (2018) Inclined landing testing of an
  omni-directional unmanned aerial vehicle.
\newblock \emph{Transactions of the Canadian Society for Mechanical
  Engineering} 42(1).
\newblock \doi{10.1139/tcsme-2017-0008}.

\bibitem[{Y{\"u}ksel et~al.(2019)Y{\"u}ksel, Secchi, B{\"u}lthoff and
  Franchi}]{Yuksel2019Aerial}
Y{\"u}ksel B, Secchi C, B{\"u}lthoff HH and Franchi A (2019) Aerial physical
  interaction via {IDA-PBC}.
\newblock \emph{The International Journal of Robotics Research} 38(4):
  403--421.
\newblock \doi{10.1177/0278364919835605}.

\bibitem[{Zhong et~al.(2025)Zhong, Liang, Chen, Zhang, Mao and
  Wang}]{PhyintReview}
Zhong H, Liang J, Chen Y, Zhang H, Mao J and Wang Y (2025) Prototype, modeling,
  and control of aerial robots with physical interaction: A review.
\newblock \emph{IEEE Transactions on Automation Science and Engineering} 22:
  3528--3542.

\end{thebibliography}

\makeatletter
\setlength{\@fptop}{0pt}

\setlength{\@fpsep}{20pt}

\setlength{\@fpbot}{0pt plus 1fil}
\makeatother

\begin{appendices}

\section{Data Tables for Valley Flattening Assessment in Higher dimensions}
\label{app:tables_hessian}

\clearpage

\begin{table*}[b]
\firstrev{
\centering
\caption{N-5 Law Validation, Subspace Alignment \& Valley Flattening (Part 1)}
\label{tab:high_n_data_part1}
\resizebox{\textwidth}{!}{%
\begin{tabular}{ccccccccccccccc}
\toprule
\textbf{$N$} & \textbf{Branch} & \textbf{Star} & \textbf{Flat Dims} & \textbf{Subspace} & \textbf{Mean $\mu_1$} & \textbf{Mean $\mu_2$} & \textbf{Mean $\mu_3$} & \textbf{Mean $\mu_4$} & \textbf{Mean $\mu_5$} & \textbf{Mean $\mu_6$} & \textbf{Mean $\mu_7$} & \textbf{Mean $\mu_8$} & \textbf{Mean $\mu_9$} & \textbf{Mean $\mu_{10}$} \\
\textbf{} & \textbf{($k$)} & \textbf{ ($q$)} & \textbf{($n_H$)} & \textbf{ Align} & \textbf{} & \textbf{} & \textbf{} & \textbf{} & \textbf{} & \textbf{} & \textbf{} & \textbf{} & \textbf{} & \textbf{} \\
\midrule
6 & 1 & 3 & 1 & 1.0000 & 9.91e-08 & 6.67e-01 & 6.67e-01 & 1.33e+00 & 1.33e+00 & 2.00e+00 & NaN & NaN & NaN & NaN \\
\midrule
7 & 1 & 3 & 1 & 1.0000 & 9.83e-08 & 2.86e-01 & 2.86e-01 & 8.57e-01 & 8.57e-01 & 1.43e+00 & 1.43e+00 & NaN & NaN & NaN \\
7 & 2 & 4 & 1 & 1.0000 & 1.34e-07 & 2.86e-01 & 2.86e-01 & 8.57e-01 & 8.57e-01 & 1.43e+00 & 1.43e+00 & NaN & NaN & NaN \\
\midrule
8 & 1 & 3 & 1 & 1.0000 & 1.50e-07 & 2.50e-01 & 2.50e-01 & 5.00e-01 & 5.00e-01 & 1.00e+00 & 1.00e+00 & 1.00e+00 & NaN & NaN \\
8 & 2 & 4 & 3 & 1.0000 & 6.54e-08 & 1.21e-07 & 1.87e-07 & 5.00e-01 & 5.00e-01 & 1.00e+00 & 1.00e+00 & 1.50e+00 & NaN & NaN \\
8 & 3 & 5 & 1 & 1.0000 & 1.39e-07 & 2.50e-01 & 2.50e-01 & 5.00e-01 & 5.00e-01 & 1.00e+00 & 1.00e+00 & 1.00e+00 & NaN & NaN \\
\midrule
9 & 1 & 3 & 1 & 1.0000 & 1.32e-07 & 2.22e-01 & 2.22e-01 & 4.44e-01 & 4.44e-01 & 6.67e-01 & 6.67e-01 & 6.67e-01 & 6.67e-01 & NaN \\
9 & 2 & 4 & 3 & 1.0000 & 5.50e-08 & 1.23e-07 & 2.03e-07 & 2.22e-01 & 2.22e-01 & 6.67e-01 & 6.67e-01 & 1.11e+00 & 1.11e+00 & NaN \\
9 & 3 & 5 & 3 & 1.0000 & 4.82e-08 & 1.05e-07 & 1.96e-07 & 2.22e-01 & 2.22e-01 & 6.67e-01 & 6.67e-01 & 1.11e+00 & 1.11e+00 & NaN \\
9 & 4 & 6 & 1 & 1.0000 & 1.33e-07 & 2.22e-01 & 2.22e-01 & 4.44e-01 & 4.44e-01 & 6.67e-01 & 6.67e-01 & 6.67e-01 & 6.67e-01 & NaN \\
\midrule
10 & 1 & 3 & 1 & 1.0000 & 1.25e-07 & 2.00e-01 & 2.00e-01 & 4.00e-01 & 4.00e-01 & 4.00e-01 & 4.00e-01 & 4.00e-01 & 6.00e-01 & 6.00e-01 \\
10 & 2 & 4 & 3 & 1.0000 & 5.30e-08 & 1.14e-07 & 1.85e-07 & 2.00e-01 & 2.00e-01 & 4.00e-01 & 4.00e-01 & 8.00e-01 & 8.00e-01 & 8.00e-01 \\
10 & 3 & 5 & 5 & 1.0000 & 3.13e-08 & 7.69e-08 & 1.25e-07 & 1.68e-07 & 2.56e-07 & 4.00e-01 & 4.00e-01 & 8.00e-01 & 8.00e-01 & 1.20e+00 \\
10 & 4 & 6 & 3 & 1.0000 & 4.95e-08 & 1.06e-07 & 1.87e-07 & 2.00e-01 & 2.00e-01 & 4.00e-01 & 4.00e-01 & 8.00e-01 & 8.00e-01 & 8.00e-01 \\
10 & 5 & 7 & 1 & 1.0000 & 1.01e-07 & 2.00e-01 & 2.00e-01 & 4.00e-01 & 4.00e-01 & 4.00e-01 & 4.00e-01 & 4.00e-01 & 6.00e-01 & 6.00e-01 \\
\midrule
11 & 1 & 3 & 1 & 1.0000 & 1.13e-07 & 1.82e-01 & 1.82e-01 & 1.82e-01 & 1.82e-01 & 3.64e-01 & 3.64e-01 & 3.64e-01 & 3.64e-01 & 5.45e-01 \\
11 & 2 & 4 & 3 & 1.0000 & 4.92e-08 & 1.12e-07 & 1.90e-07 & 1.82e-01 & 1.82e-01 & 3.64e-01 & 3.64e-01 & 5.45e-01 & 5.45e-01 & 5.45e-01 \\
11 & 3 & 5 & 5 & 1.0000 & 2.90e-08 & 7.33e-08 & 1.26e-07 & 1.84e-07 & 2.54e-07 & 1.82e-01 & 1.82e-01 & 5.45e-01 & 5.45e-01 & 9.09e-01 \\
11 & 4 & 6 & 5 & 1.0000 & 3.97e-08 & 8.64e-08 & 1.33e-07 & 2.00e-07 & 2.63e-07 & 1.82e-01 & 1.82e-01 & 5.45e-01 & 5.45e-01 & 9.09e-01 \\
11 & 5 & 7 & 3 & 1.0000 & 5.62e-08 & 1.09e-07 & 1.94e-07 & 1.82e-01 & 1.82e-01 & 3.64e-01 & 3.64e-01 & 5.45e-01 & 5.45e-01 & 5.45e-01 \\
11 & 6 & 8 & 1 & 1.0000 & 1.29e-07 & 1.82e-01 & 1.82e-01 & 1.82e-01 & 1.82e-01 & 3.64e-01 & 3.64e-01 & 3.64e-01 & 3.64e-01 & 5.45e-01 \\
\midrule
12 & 1 & 3 & 2 & 1.0000 & 9.01e-08 & 1.78e-07 & 1.67e-01 & 1.67e-01 & 1.67e-01 & 1.67e-01 & 3.33e-01 & 3.33e-01 & 3.33e-01 & 3.33e-01 \\
12 & 2 & 4 & 3 & 1.0000 & 6.66e-08 & 1.28e-07 & 2.04e-07 & 1.67e-01 & 1.67e-01 & 3.33e-01 & 3.33e-01 & 3.33e-01 & 3.33e-01 & 3.33e-01 \\
12 & 3 & 5 & 5 & 1.0000 & 4.47e-08 & 9.71e-08 & 1.46e-07 & 1.99e-07 & 2.67e-07 & 1.67e-01 & 1.67e-01 & 3.33e-01 & 3.33e-01 & 6.67e-01 \\
12 & 4 & 6 & 7 & 1.0000 & 2.77e-08 & 6.40e-08 & 1.04e-07 & 1.46e-07 & 1.88e-07 & 2.35e-07 & 3.10e-07 & 3.33e-01 & 3.33e-01 & 6.67e-01 \\
12 & 5 & 7 & 5 & 1.0000 & 2.91e-08 & 7.01e-08 & 1.15e-07 & 1.66e-07 & 2.35e-07 & 1.67e-01 & 1.67e-01 & 3.33e-01 & 3.33e-01 & 6.67e-01 \\
12 & 6 & 8 & 3 & 1.0000 & 6.84e-08 & 1.32e-07 & 2.27e-07 & 1.67e-01 & 1.67e-01 & 3.33e-01 & 3.33e-01 & 3.33e-01 & 3.33e-01 & 3.33e-01 \\
12 & 7 & 9 & 2 & 1.0000 & 9.56e-08 & 1.76e-07 & 1.67e-01 & 1.67e-01 & 1.67e-01 & 1.67e-01 & 3.33e-01 & 3.33e-01 & 3.33e-01 & 3.33e-01 \\
\midrule
13 & 1 & 3 & 3 & 1.0000 & 6.66e-08 & 1.25e-07 & 2.11e-07 & 1.54e-01 & 1.54e-01 & 1.54e-01 & 1.54e-01 & 3.08e-01 & 3.08e-01 & 3.08e-01 \\
13 & 2 & 4 & 3 & 1.0000 & 5.84e-08 & 1.16e-07 & 2.16e-07 & 1.54e-01 & 1.54e-01 & 1.54e-01 & 1.54e-01 & 3.08e-01 & 3.08e-01 & 3.08e-01 \\
13 & 3 & 5 & 5 & 1.0000 & 4.16e-08 & 8.74e-08 & 1.36e-07 & 1.84e-07 & 2.72e-07 & 1.54e-01 & 1.54e-01 & 3.08e-01 & 3.08e-01 & 4.62e-01 \\
13 & 4 & 6 & 7 & 1.0000 & 2.79e-08 & 6.36e-08 & 9.42e-08 & 1.27e-07 & 1.68e-07 & 2.18e-07 & 2.96e-07 & 1.54e-01 & 1.54e-01 & 4.62e-01 \\
13 & 5 & 7 & 7 & 1.0000 & 3.26e-08 & 7.05e-08 & 1.05e-07 & 1.52e-07 & 1.96e-07 & 2.50e-07 & 3.21e-07 & 1.54e-01 & 1.54e-01 & 4.62e-01 \\
13 & 6 & 8 & 5 & 1.0000 & 5.50e-08 & 1.11e-07 & 1.60e-07 & 2.17e-07 & 2.89e-07 & 1.54e-01 & 1.54e-01 & 3.08e-01 & 3.08e-01 & 4.62e-01 \\
13 & 7 & 9 & 3 & 1.0000 & 6.46e-08 & 1.27e-07 & 2.09e-07 & 1.54e-01 & 1.54e-01 & 1.54e-01 & 1.54e-01 & 3.08e-01 & 3.08e-01 & 3.08e-01 \\
13 & 8 & 10 & 3 & 1.0000 & 8.34e-08 & 1.54e-07 & 2.42e-07 & 1.54e-01 & 1.54e-01 & 1.54e-01 & 1.54e-01 & 3.08e-01 & 3.08e-01 & 3.08e-01 \\
\midrule
14 & 1 & 3 & 4 & 1.0000 & 4.36e-08 & 9.24e-08 & 1.47e-07 & 2.36e-07 & 1.43e-01 & 1.43e-01 & 1.43e-01 & 1.43e-01 & 2.86e-01 & 2.86e-01 \\
14 & 2 & 4 & 4 & 1.0000 & 5.94e-08 & 1.15e-07 & 1.84e-07 & 2.53e-07 & 1.43e-01 & 1.43e-01 & 1.43e-01 & 1.43e-01 & 2.86e-01 & 2.86e-01 \\
14 & 3 & 5 & 5 & 1.0000 & 3.49e-08 & 8.39e-08 & 1.23e-07 & 1.82e-07 & 2.57e-07 & 1.43e-01 & 1.43e-01 & 2.86e-01 & 2.86e-01 & 2.86e-01 \\
14 & 4 & 6 & 7 & 1.0000 & 3.28e-08 & 6.82e-08 & 1.03e-07 & 1.46e-07 & 1.92e-07 & 2.46e-07 & 3.26e-07 & 1.43e-01 & 1.43e-01 & 2.86e-01 \\
14 & 5 & 7 & 9 & 1.0000 & 2.13e-08 & 5.17e-08 & 8.80e-08 & 1.24e-07 & 1.61e-07 & 1.99e-07 & 2.44e-07 & 2.95e-07 & 3.68e-07 & 2.86e-01 \\
14 & 6 & 8 & 7 & 1.0000 & 3.66e-08 & 7.72e-08 & 1.18e-07 & 1.61e-07 & 2.14e-07 & 2.58e-07 & 3.36e-07 & 1.43e-01 & 1.43e-01 & 2.86e-01 \\
14 & 7 & 9 & 5 & 1.0000 & 4.74e-08 & 9.20e-08 & 1.44e-07 & 2.04e-07 & 2.78e-07 & 1.43e-01 & 1.43e-01 & 2.86e-01 & 2.86e-01 & 2.86e-01 \\
14 & 8 & 10 & 4 & 1.0000 & 5.63e-08 & 1.01e-07 & 1.73e-07 & 2.39e-07 & 1.43e-01 & 1.43e-01 & 1.43e-01 & 1.43e-01 & 2.86e-01 & 2.86e-01 \\
14 & 9 & 11 & 4 & 1.0000 & 4.85e-08 & 1.10e-07 & 1.66e-07 & 2.50e-07 & 1.43e-01 & 1.43e-01 & 1.43e-01 & 1.43e-01 & 2.86e-01 & 2.86e-01 \\
\midrule
15 & 1 & 3 & 5 & 1.0000 & 4.77e-08 & 1.01e-07 & 1.51e-07 & 2.16e-07 & 2.83e-07 & 1.33e-01 & 1.33e-01 & 1.33e-01 & 1.33e-01 & 2.67e-01 \\
15 & 2 & 4 & 5 & 1.0000 & 3.42e-08 & 1.00e-07 & 1.51e-07 & 2.09e-07 & 2.82e-07 & 1.33e-01 & 1.33e-01 & 1.33e-01 & 1.33e-01 & 2.67e-01 \\
15 & 3 & 5 & 5 & 1.0000 & 4.32e-08 & 8.30e-08 & 1.38e-07 & 1.98e-07 & 2.77e-07 & 1.33e-01 & 1.33e-01 & 1.33e-01 & 1.33e-01 & 2.67e-01 \\
15 & 4 & 6 & 7 & 1.0000 & 3.00e-08 & 6.15e-08 & 1.04e-07 & 1.49e-07 & 1.93e-07 & 2.59e-07 & 3.27e-07 & 1.33e-01 & 1.33e-01 & 2.67e-01 \\
15 & 5 & 7 & 9 & 1.0000 & 2.49e-08 & 5.63e-08 & 8.82e-08 & 1.23e-07 & 1.58e-07 & 2.00e-07 & 2.46e-07 & 2.96e-07 & 3.65e-07 & 1.33e-01 \\
15 & 6 & 8 & 9 & 1.0000 & 2.10e-08 & 5.32e-08 & 8.29e-08 & 1.15e-07 & 1.50e-07 & 1.89e-07 & 2.36e-07 & 2.86e-07 & 3.42e-07 & 1.33e-01 \\
15 & 7 & 9 & 7 & 1.0000 & 2.81e-08 & 7.40e-08 & 1.10e-07 & 1.58e-07 & 2.07e-07 & 2.57e-07 & 3.25e-07 & 1.33e-01 & 1.33e-01 & 2.67e-01 \\
15 & 8 & 10 & 5 & 1.0000 & 5.11e-08 & 1.08e-07 & 1.56e-07 & 2.24e-07 & 3.12e-07 & 1.33e-01 & 1.33e-01 & 1.33e-01 & 1.33e-01 & 2.67e-01 \\
15 & 9 & 11 & 5 & 1.0000 & 5.88e-08 & 1.08e-07 & 1.60e-07 & 2.25e-07 & 3.09e-07 & 1.33e-01 & 1.33e-01 & 1.33e-01 & 1.33e-01 & 2.67e-01 \\
15 & 10 & 12 & 5 & 1.0000 & 4.61e-08 & 9.12e-08 & 1.52e-07 & 2.03e-07 & 2.74e-07 & 1.33e-01 & 1.33e-01 & 1.33e-01 & 1.33e-01 & 2.67e-01 \\
\bottomrule
\end{tabular}%
}}
\end{table*}

\begin{table*}[htbp]
\firstrev{
\centering
\caption{N-5 Law Validation, Subspace Alignment \& Valley Flattening (Part 2)}
\label{tab:high_n_data_part2}
\resizebox{\textwidth}{!}{%
\begin{tabular}{ccccccccccccccc}
\toprule
\textbf{$N$} & \textbf{Branch} & \textbf{Star} & \textbf{Flat Dims} & \textbf{Subspace} & \textbf{Mean $\mu_1$} & \textbf{Mean $\mu_2$} & \textbf{Mean $\mu_3$} & \textbf{Mean $\mu_4$} & \textbf{Mean $\mu_5$} & \textbf{Mean $\mu_6$} & \textbf{Mean $\mu_7$} & \textbf{Mean $\mu_8$} & \textbf{Mean $\mu_9$} & \textbf{Mean $\mu_{10}$} \\
\textbf{} & \textbf{($k$)} & \textbf{ ($q$)} & \textbf{($n_H$)} & \textbf{ Align} & \textbf{} & \textbf{} & \textbf{} & \textbf{} & \textbf{} & \textbf{} & \textbf{} & \textbf{} & \textbf{} & \textbf{} \\
\midrule
16 & 1 & 3 & 6 & 1.0000 & 4.48e-08 & 1.01e-07 & 1.47e-07 & 1.95e-07 & 2.61e-07 & 3.46e-07 & 1.25e-01 & 1.25e-01 & 1.25e-01 & 1.25e-01 \\
16 & 2 & 4 & 6 & 1.0000 & 5.09e-08 & 9.72e-08 & 1.50e-07 & 2.04e-07 & 2.58e-07 & 3.45e-07 & 1.25e-01 & 1.25e-01 & 1.25e-01 & 1.25e-01 \\
16 & 3 & 5 & 6 & 1.0000 & 3.77e-08 & 8.53e-08 & 1.37e-07 & 1.82e-07 & 2.47e-07 & 3.35e-07 & 1.25e-01 & 1.25e-01 & 1.25e-01 & 1.25e-01 \\
16 & 4 & 6 & 7 & 1.0000 & 4.63e-08 & 9.34e-08 & 1.38e-07 & 1.86e-07 & 2.39e-07 & 3.00e-07 & 3.86e-07 & 1.25e-01 & 1.25e-01 & 2.50e-01 \\
16 & 5 & 7 & 9 & 1.0000 & 2.19e-08 & 5.62e-08 & 9.12e-08 & 1.27e-07 & 1.66e-07 & 2.00e-07 & 2.50e-07 & 3.05e-07 & 3.75e-07 & 1.25e-01 \\
16 & 6 & 8 & 11 & 1.0000 & 2.65e-08 & 5.69e-08 & 8.50e-08 & 1.20e-07 & 1.51e-07 & 1.87e-07 & 2.27e-07 & 2.64e-07 & 3.08e-07 & 3.62e-07 \\
16 & 7 & 9 & 9 & 1.0000 & 4.73e-08 & 8.71e-08 & 1.24e-07 & 1.64e-07 & 2.07e-07 & 2.47e-07 & 2.96e-07 & 3.48e-07 & 4.31e-07 & 1.25e-01 \\
16 & 8 & 10 & 7 & 1.0000 & 2.92e-08 & 7.11e-08 & 1.14e-07 & 1.66e-07 & 2.16e-07 & 2.90e-07 & 3.87e-07 & 1.25e-01 & 1.25e-01 & 2.50e-01 \\
16 & 9 & 11 & 6 & 1.0000 & 3.52e-08 & 7.86e-08 & 1.31e-07 & 1.76e-07 & 2.33e-07 & 3.24e-07 & 1.25e-01 & 1.25e-01 & 1.25e-01 & 1.25e-01 \\
16 & 10 & 12 & 6 & 1.0000 & 5.12e-08 & 1.08e-07 & 1.64e-07 & 2.20e-07 & 2.89e-07 & 3.89e-07 & 1.25e-01 & 1.25e-01 & 1.25e-01 & 1.25e-01 \\
16 & 11 & 13 & 6 & 1.0000 & 4.25e-08 & 1.00e-07 & 1.53e-07 & 2.09e-07 & 2.75e-07 & 3.52e-07 & 1.25e-01 & 1.25e-01 & 1.25e-01 & 1.25e-01 \\
\midrule
17 & 1 & 3 & 7 & 1.0000 & 3.00e-08 & 7.17e-08 & 1.20e-07 & 1.65e-07 & 2.08e-07 & 2.64e-07 & 3.36e-07 & 1.18e-01 & 1.18e-01 & 1.18e-01 \\
17 & 2 & 4 & 7 & 1.0000 & 3.55e-08 & 8.15e-08 & 1.24e-07 & 1.73e-07 & 2.16e-07 & 2.82e-07 & 3.58e-07 & 1.18e-01 & 1.18e-01 & 1.18e-01 \\
17 & 3 & 5 & 7 & 1.0000 & 4.25e-08 & 8.06e-08 & 1.19e-07 & 1.61e-07 & 2.10e-07 & 2.67e-07 & 3.54e-07 & 1.18e-01 & 1.18e-01 & 1.18e-01 \\
17 & 4 & 6 & 7 & 1.0000 & 3.81e-08 & 8.12e-08 & 1.22e-07 & 1.69e-07 & 2.22e-07 & 2.77e-07 & 3.54e-07 & 1.18e-01 & 1.18e-01 & 1.18e-01 \\
17 & 5 & 7 & 9 & 1.0000 & 3.45e-08 & 7.38e-08 & 1.12e-07 & 1.48e-07 & 1.82e-07 & 2.17e-07 & 2.58e-07 & 3.13e-07 & 3.96e-07 & 1.18e-01 \\
17 & 6 & 8 & 11 & 1.0000 & 1.88e-08 & 4.06e-08 & 7.71e-08 & 1.11e-07 & 1.45e-07 & 1.79e-07 & 2.08e-07 & 2.43e-07 & 2.84e-07 & 3.29e-07 \\
17 & 7 & 9 & 11 & 1.0000 & 2.21e-08 & 5.16e-08 & 8.88e-08 & 1.24e-07 & 1.52e-07 & 1.86e-07 & 2.22e-07 & 2.59e-07 & 3.16e-07 & 3.72e-07 \\
17 & 8 & 10 & 9 & 1.0000 & 3.48e-08 & 6.99e-08 & 1.01e-07 & 1.45e-07 & 1.87e-07 & 2.30e-07 & 2.77e-07 & 3.32e-07 & 4.07e-07 & 1.18e-01 \\
17 & 9 & 11 & 7 & 1.0000 & 3.82e-08 & 8.76e-08 & 1.34e-07 & 1.84e-07 & 2.35e-07 & 2.90e-07 & 3.74e-07 & 1.18e-01 & 1.18e-01 & 1.18e-01 \\
17 & 10 & 12 & 7 & 1.0000 & 3.85e-08 & 8.33e-08 & 1.21e-07 & 1.59e-07 & 2.17e-07 & 2.70e-07 & 3.53e-07 & 1.18e-01 & 1.18e-01 & 1.18e-01 \\
17 & 11 & 13 & 7 & 1.0000 & 2.56e-08 & 7.18e-08 & 1.12e-07 & 1.54e-07 & 2.05e-07 & 2.64e-07 & 3.57e-07 & 1.18e-01 & 1.18e-01 & 1.18e-01 \\
17 & 12 & 14 & 7 & 1.0000 & 3.38e-08 & 7.80e-08 & 1.25e-07 & 1.60e-07 & 2.09e-07 & 2.69e-07 & 3.47e-07 & 1.18e-01 & 1.18e-01 & 1.18e-01 \\
\midrule
18 & 1 & 3 & 8 & 1.0000 & 2.80e-08 & 7.07e-08 & 1.01e-07 & 1.43e-07 & 1.87e-07 & 2.31e-07 & 2.94e-07 & 3.57e-07 & 1.11e-01 & 1.11e-01 \\
18 & 2 & 4 & 8 & 1.0000 & 2.92e-08 & 5.96e-08 & 1.01e-07 & 1.32e-07 & 1.78e-07 & 2.21e-07 & 2.79e-07 & 3.58e-07 & 1.11e-01 & 1.11e-01 \\
18 & 3 & 5 & 8 & 1.0000 & 3.79e-08 & 6.71e-08 & 1.14e-07 & 1.55e-07 & 2.00e-07 & 2.51e-07 & 3.06e-07 & 3.86e-07 & 1.11e-01 & 1.11e-01 \\
18 & 4 & 6 & 8 & 1.0000 & 2.63e-08 & 6.01e-08 & 9.57e-08 & 1.34e-07 & 1.75e-07 & 2.26e-07 & 2.92e-07 & 3.59e-07 & 1.11e-01 & 1.11e-01 \\
18 & 5 & 7 & 9 & 1.0000 & 2.39e-08 & 5.72e-08 & 1.00e-07 & 1.32e-07 & 1.73e-07 & 2.12e-07 & 2.58e-07 & 3.18e-07 & 3.82e-07 & 1.11e-01 \\
18 & 6 & 8 & 11 & 1.0000 & 1.96e-08 & 5.29e-08 & 8.52e-08 & 1.12e-07 & 1.45e-07 & 1.80e-07 & 2.19e-07 & 2.58e-07 & 2.98e-07 & 3.49e-07 \\
18 & 7 & 9 & 13 & 1.0000 & 1.89e-08 & 3.95e-08 & 6.46e-08 & 9.62e-08 & 1.23e-07 & 1.50e-07 & 1.80e-07 & 2.12e-07 & 2.48e-07 & 2.84e-07 \\
18 & 8 & 10 & 11 & 1.0000 & 1.82e-08 & 4.62e-08 & 7.13e-08 & 1.03e-07 & 1.36e-07 & 1.70e-07 & 2.06e-07 & 2.44e-07 & 2.88e-07 & 3.41e-07 \\
18 & 9 & 11 & 9 & 1.0000 & 2.46e-08 & 5.56e-08 & 9.12e-08 & 1.30e-07 & 1.75e-07 & 2.11e-07 & 2.54e-07 & 3.16e-07 & 3.81e-07 & 1.11e-01 \\
18 & 10 & 12 & 8 & 1.0000 & 3.84e-08 & 7.58e-08 & 1.20e-07 & 1.51e-07 & 2.03e-07 & 2.48e-07 & 3.03e-07 & 3.88e-07 & 1.11e-01 & 1.11e-01 \\
18 & 11 & 13 & 8 & 1.0000 & 3.08e-08 & 6.44e-08 & 1.05e-07 & 1.35e-07 & 1.82e-07 & 2.26e-07 & 2.81e-07 & 3.65e-07 & 1.11e-01 & 1.11e-01 \\
18 & 12 & 14 & 8 & 1.0000 & 2.90e-08 & 7.76e-08 & 1.09e-07 & 1.50e-07 & 1.91e-07 & 2.37e-07 & 2.92e-07 & 3.63e-07 & 1.11e-01 & 1.11e-01 \\
18 & 13 & 15 & 8 & 1.0000 & 3.59e-08 & 6.73e-08 & 1.05e-07 & 1.47e-07 & 1.92e-07 & 2.41e-07 & 2.97e-07 & 3.70e-07 & 1.11e-01 & 1.11e-01 \\
\midrule
19 & 1 & 3 & 9 & 1.0000 & 2.64e-08 & 6.39e-08 & 9.64e-08 & 1.34e-07 & 1.63e-07 & 2.06e-07 & 2.55e-07 & 3.09e-07 & 3.79e-07 & 1.05e-01 \\
19 & 2 & 4 & 9 & 1.0000 & 3.73e-08 & 7.14e-08 & 1.07e-07 & 1.50e-07 & 1.90e-07 & 2.29e-07 & 2.76e-07 & 3.28e-07 & 4.03e-07 & 1.05e-01 \\
19 & 3 & 5 & 9 & 1.0000 & 2.88e-08 & 6.50e-08 & 9.88e-08 & 1.42e-07 & 1.80e-07 & 2.30e-07 & 2.81e-07 & 3.38e-07 & 4.19e-07 & 1.05e-01 \\
19 & 4 & 6 & 9 & 1.0000 & 2.98e-08 & 6.43e-08 & 1.05e-07 & 1.49e-07 & 1.93e-07 & 2.33e-07 & 2.80e-07 & 3.37e-07 & 4.12e-07 & 1.05e-01 \\
19 & 5 & 7 & 9 & 1.0000 & 3.81e-08 & 8.06e-08 & 1.18e-07 & 1.56e-07 & 2.00e-07 & 2.41e-07 & 2.89e-07 & 3.49e-07 & 4.19e-07 & 1.05e-01 \\
19 & 6 & 8 & 11 & 1.0000 & 2.14e-08 & 5.32e-08 & 8.35e-08 & 1.18e-07 & 1.48e-07 & 1.86e-07 & 2.22e-07 & 2.58e-07 & 3.06e-07 & 3.53e-07 \\
19 & 7 & 9 & 13 & 1.0000 & 1.79e-08 & 4.26e-08 & 6.66e-08 & 9.34e-08 & 1.25e-07 & 1.59e-07 & 1.88e-07 & 2.24e-07 & 2.61e-07 & 2.96e-07 \\
19 & 8 & 10 & 13 & 1.0000 & 2.23e-08 & 4.92e-08 & 7.65e-08 & 1.10e-07 & 1.40e-07 & 1.69e-07 & 2.01e-07 & 2.35e-07 & 2.69e-07 & 3.08e-07 \\
19 & 9 & 11 & 11 & 1.0000 & 1.78e-08 & 4.75e-08 & 7.50e-08 & 1.00e-07 & 1.32e-07 & 1.68e-07 & 2.02e-07 & 2.43e-07 & 2.87e-07 & 3.37e-07 \\
19 & 10 & 12 & 9 & 1.0000 & 3.21e-08 & 6.89e-08 & 1.03e-07 & 1.42e-07 & 1.74e-07 & 2.16e-07 & 2.66e-07 & 3.20e-07 & 3.98e-07 & 1.05e-01 \\
19 & 11 & 13 & 9 & 1.0000 & 2.61e-08 & 6.11e-08 & 1.01e-07 & 1.43e-07 & 1.83e-07 & 2.26e-07 & 2.77e-07 & 3.41e-07 & 4.13e-07 & 1.05e-01 \\
19 & 12 & 14 & 9 & 1.0000 & 2.13e-08 & 5.38e-08 & 9.43e-08 & 1.27e-07 & 1.63e-07 & 2.05e-07 & 2.43e-07 & 2.94e-07 & 3.72e-07 & 1.05e-01 \\
19 & 13 & 15 & 9 & 1.0000 & 3.49e-08 & 6.91e-08 & 1.07e-07 & 1.47e-07 & 1.91e-07 & 2.38e-07 & 2.96e-07 & 3.44e-07 & 4.20e-07 & 1.05e-01 \\
19 & 14 & 16 & 9 & 1.0000 & 2.49e-08 & 5.56e-08 & 8.62e-08 & 1.25e-07 & 1.67e-07 & 2.03e-07 & 2.51e-07 & 3.03e-07 & 3.78e-07 & 1.05e-01 \\
\midrule
20 & 1 & 3 & 10 & 1.0000 & 1.87e-08 & 4.89e-08 & 8.19e-08 & 1.13e-07 & 1.43e-07 & 1.81e-07 & 2.17e-07 & 2.62e-07 & 3.14e-07 & 3.77e-07 \\
20 & 2 & 4 & 10 & 1.0000 & 2.46e-08 & 5.56e-08 & 8.69e-08 & 1.17e-07 & 1.53e-07 & 1.87e-07 & 2.24e-07 & 2.62e-07 & 3.18e-07 & 3.77e-07 \\
20 & 3 & 5 & 10 & 1.0000 & 2.43e-08 & 4.75e-08 & 7.82e-08 & 1.15e-07 & 1.55e-07 & 1.89e-07 & 2.30e-07 & 2.75e-07 & 3.27e-07 & 3.99e-07 \\
20 & 4 & 6 & 10 & 1.0000 & 1.95e-08 & 5.06e-08 & 8.50e-08 & 1.14e-07 & 1.53e-07 & 1.90e-07 & 2.31e-07 & 2.82e-07 & 3.44e-07 & 4.08e-07 \\
20 & 5 & 7 & 10 & 1.0000 & 2.01e-08 & 5.24e-08 & 7.98e-08 & 1.16e-07 & 1.49e-07 & 1.78e-07 & 2.17e-07 & 2.62e-07 & 3.06e-07 & 3.71e-07 \\
20 & 6 & 8 & 11 & 1.0000 & 1.94e-08 & 4.95e-08 & 7.51e-08 & 1.09e-07 & 1.40e-07 & 1.77e-07 & 2.12e-07 & 2.49e-07 & 2.91e-07 & 3.36e-07 \\
20 & 7 & 9 & 13 & 1.0000 & 1.46e-08 & 4.00e-08 & 6.23e-08 & 9.12e-08 & 1.18e-07 & 1.46e-07 & 1.71e-07 & 2.01e-07 & 2.34e-07 & 2.72e-07 \\
20 & 8 & 10 & 15 & 1.0000 & 1.63e-08 & 4.02e-08 & 5.92e-08 & 8.11e-08 & 1.05e-07 & 1.29e-07 & 1.54e-07 & 1.80e-07 & 2.13e-07 & 2.41e-07 \\
20 & 9 & 11 & 13 & 1.0000 & 2.02e-08 & 4.41e-08 & 7.20e-08 & 1.03e-07 & 1.34e-07 & 1.60e-07 & 1.94e-07 & 2.29e-07 & 2.61e-07 & 2.99e-07 \\
20 & 10 & 12 & 11 & 1.0000 & 1.78e-08 & 5.34e-08 & 8.33e-08 & 1.13e-07 & 1.47e-07 & 1.78e-07 & 2.18e-07 & 2.57e-07 & 2.98e-07 & 3.49e-07 \\
20 & 11 & 13 & 10 & 1.0000 & 3.48e-08 & 6.18e-08 & 9.29e-08 & 1.28e-07 & 1.61e-07 & 2.00e-07 & 2.44e-07 & 2.88e-07 & 3.46e-07 & 4.13e-07 \\
20 & 12 & 14 & 10 & 1.0000 & 2.10e-08 & 5.03e-08 & 8.53e-08 & 1.12e-07 & 1.43e-07 & 1.81e-07 & 2.22e-07 & 2.67e-07 & 3.18e-07 & 3.82e-07 \\
20 & 13 & 15 & 10 & 1.0000 & 2.45e-08 & 6.55e-08 & 9.81e-08 & 1.32e-07 & 1.69e-07 & 2.04e-07 & 2.51e-07 & 3.03e-07 & 3.57e-07 & 4.34e-07 \\
20 & 14 & 16 & 10 & 1.0000 & 1.98e-08 & 4.62e-08 & 7.76e-08 & 1.09e-07 & 1.45e-07 & 1.85e-07 & 2.19e-07 & 2.63e-07 & 3.11e-07 & 3.86e-07 \\
20 & 15 & 17 & 10 & 1.0000 & 3.47e-08 & 6.87e-08 & 1.03e-07 & 1.37e-07 & 1.69e-07 & 2.08e-07 & 2.49e-07 & 3.00e-07 & 3.46e-07 & 4.13e-07 \\
\bottomrule
\end{tabular}%
}}
\end{table*}

\clearpage
\section{Additional Figures from Data Analysis}
\label{app:additional_figs}


\begin{figure}[h]
    \centering
    \framebox{\includegraphics[width=0.90\linewidth]{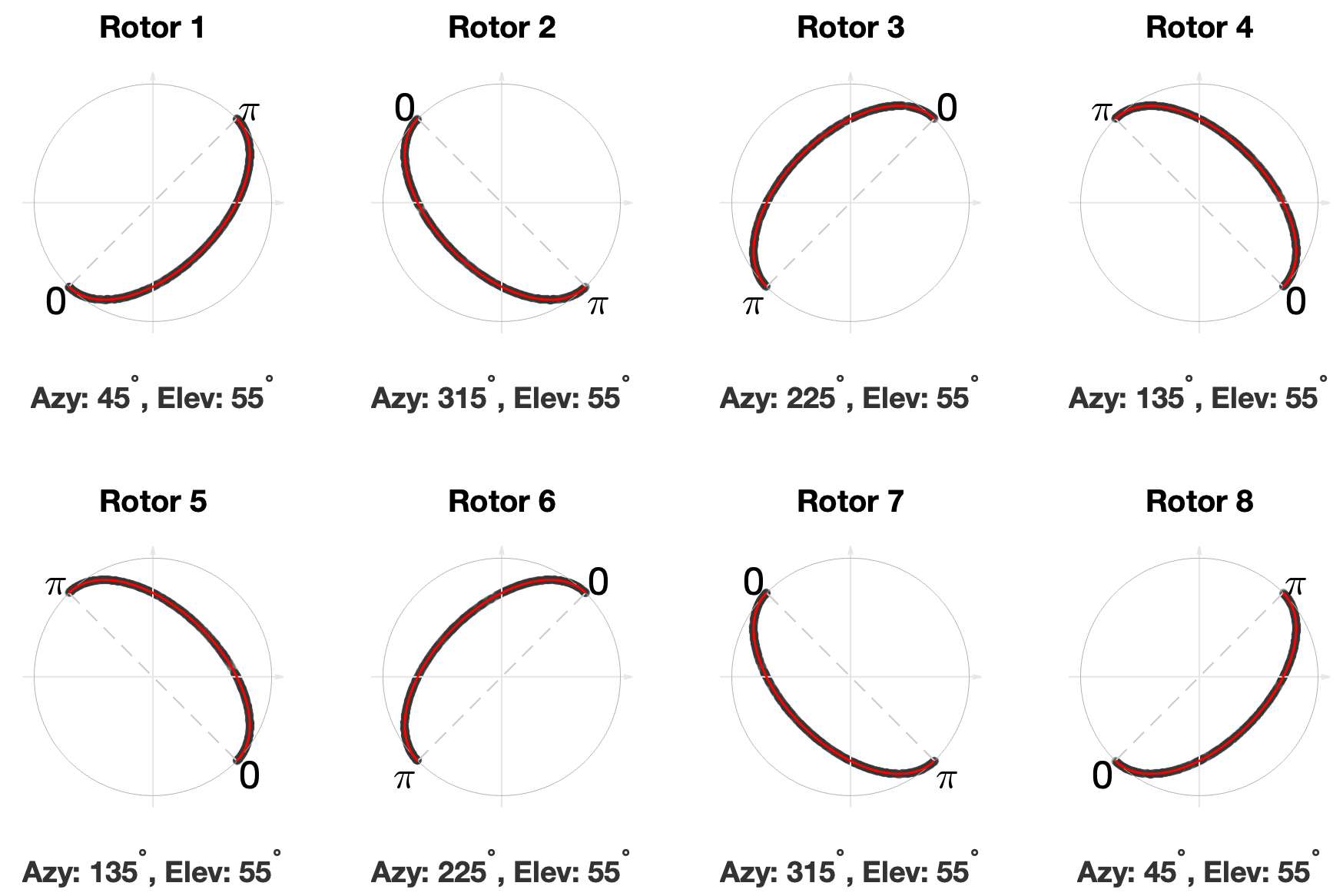}}
    \caption{Representation of the solution landscape data points (${M^*}$ points) in $\Manifold$ for the chassis: \textbf{Cube ($N=8$)}. The gray dots represent the two coordinates of these points on each disc $\RPtwo_i$. The data is perfectly fitted by the red semi-ellipses, whose parameters (rotation and elevation are shown in each subplot). The combination of all the semi-elliptical curves matches exactly the $\tangentTorus$ of the \textbf{Cube ($N=8$)}.}
    \label{fig:fit_cube}
\end{figure}

\begin{figure}[t]
    \centering
    \framebox{\includegraphics[width=0.9\linewidth]{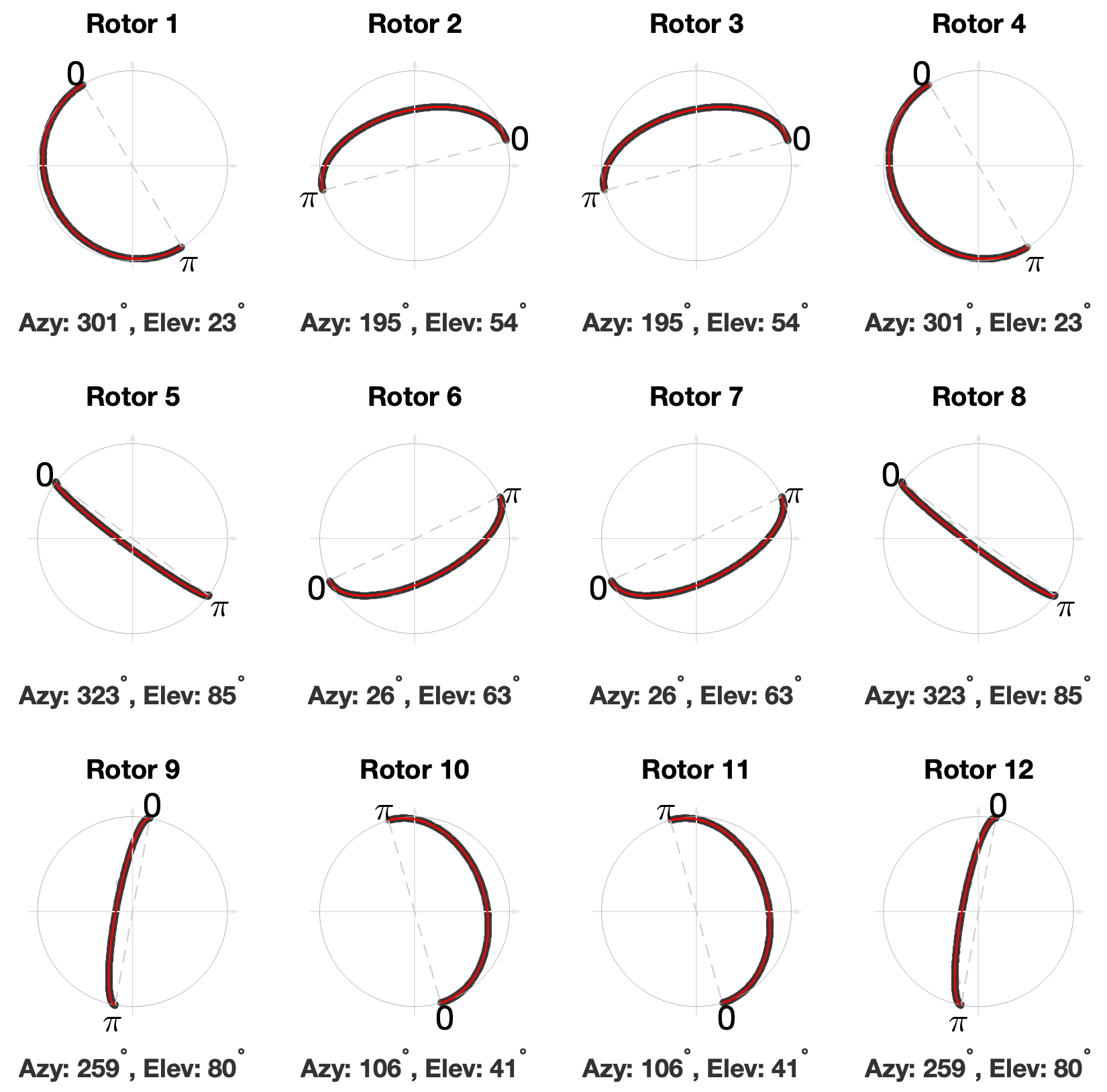}}
    \caption{Representation of the solution landscape data points (${M^*}$ points) in $\Manifold$ for the chassis: \textbf{Icosahedron ($N=12$)}. The gray dots represent the two coordinates of these points on each disc $\RPtwo_i$. The data is perfectly fitted by the red semi-ellipses, whose parameters (rotation and elevation are shown in each subplot). The combination of all the semi-elliptical curves matches exactly the $\tangentTorus$ of the \textbf{Icosahedron ($N=12$)}.}
    \label{fig:fit_icosahedron}
\end{figure}

\begin{figure}[t]
    \centering
    \framebox{\includegraphics[width=0.99\linewidth]{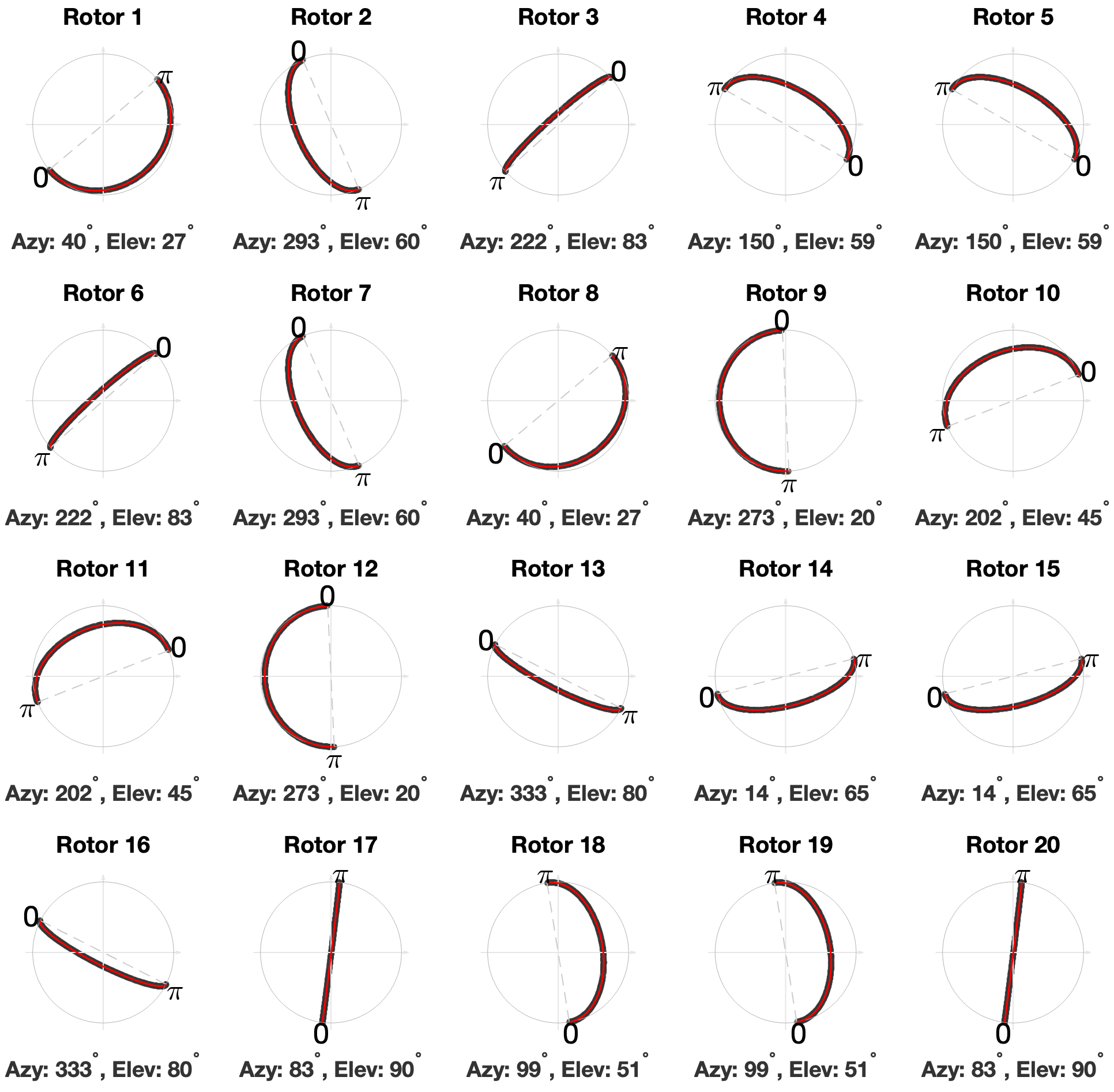}}
    \caption{Representation of the solution landscape data points (${M^*}$ points) in $\Manifold$ for the chassis: \textbf{Dodecahedron ($N=20$)}. The gray dots represent the two coordinates of these points on each disc $\RPtwo_i$. The data is perfectly fitted by the red semi-ellipses, whose parameters (rotation and elevation are shown in each subplot). The combination of all the semi-elliptical curves matches exactly the $\tangentTorus$ of the \textbf{Dodecahedron ($N=20$)}.}
    \label{fig:fit_dodecahedron}
\end{figure}


\begin{figure}[t]
    \centering
    \framebox{\includegraphics[width=0.8\linewidth]{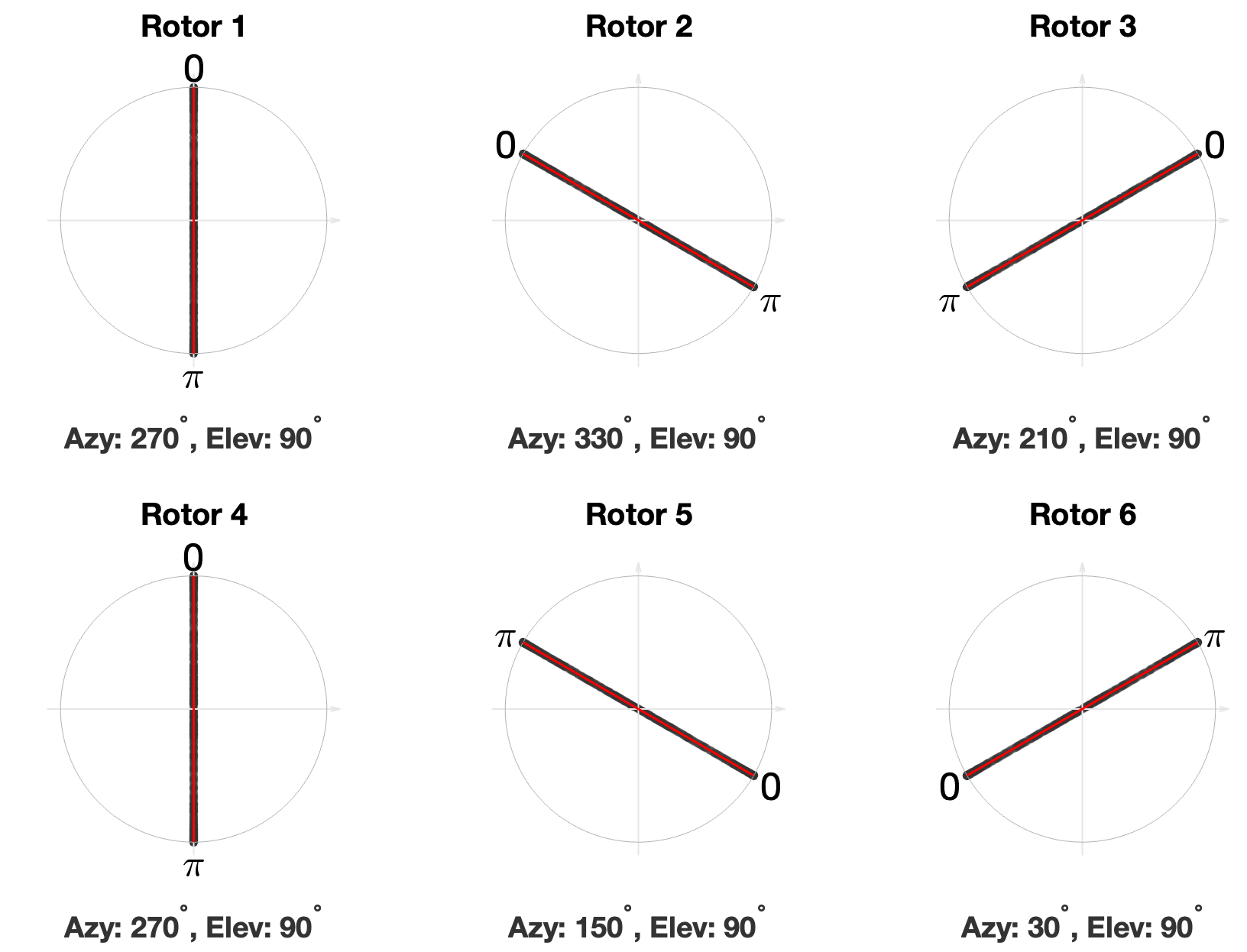}}
    \caption{Representation of the solution landscape data points (${M^*}$ points) in $\Manifold$ for the chassis: \textbf{Hexagon ($N=6$)}. The gray dots represent the two coordinates of these points on each disc $\RPtwo_i$. The data is perfectly fitted by the red semi-ellipses, whose parameters (rotation and elevation) are shown in each subplot. The parameters recovered by the fit reveal that the combination of all semi-elliptical curves corresponds exactly to the $\tangentTorus$ of the \textbf{Hexagon ($N=6$)}.}
    \label{fig:fit_poly6}
\end{figure}

\begin{figure}[t]
    \centering
    \framebox{\includegraphics[width=0.99\linewidth]{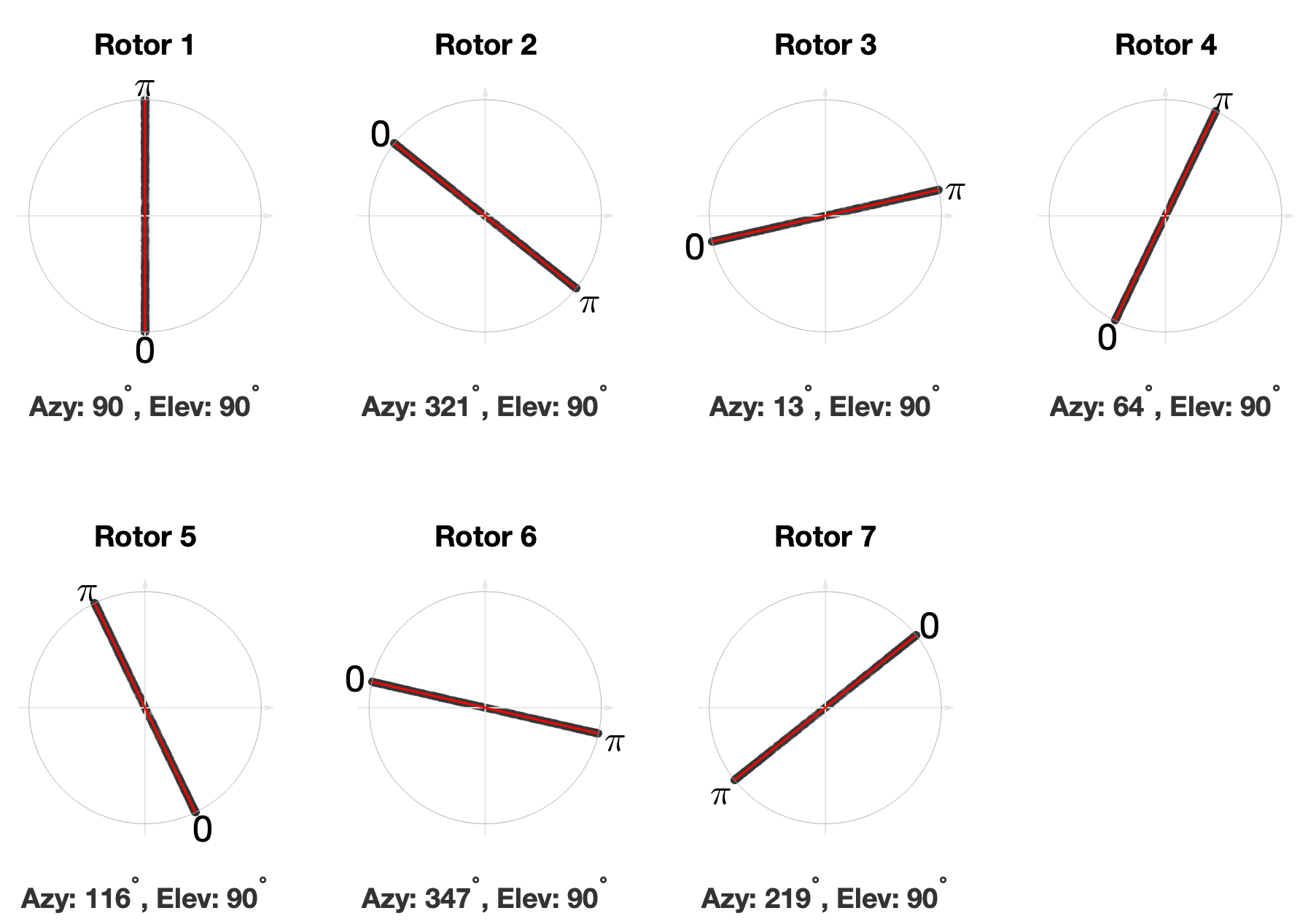}}
    \caption{Representation of the solution landscape data points (${M^*}$ points) in $\Manifold$ for the chassis: \textbf{Heptagon ($N=7$)}. The gray dots represent the two coordinates of these points on each disc $\RPtwo_i$. The data is perfectly fitted by the red semi-ellipses, whose parameters (rotation and elevation are shown in each subplot). The combination of all the semi-elliptical curves matches exactly the $\tangentTorus$ of the \textbf{Heptagon ($N=7$)}.}
    \label{fig:fit_poly7}
\end{figure}

\begin{figure}[t]
    \centering
    \framebox{\includegraphics[width=0.99\linewidth]{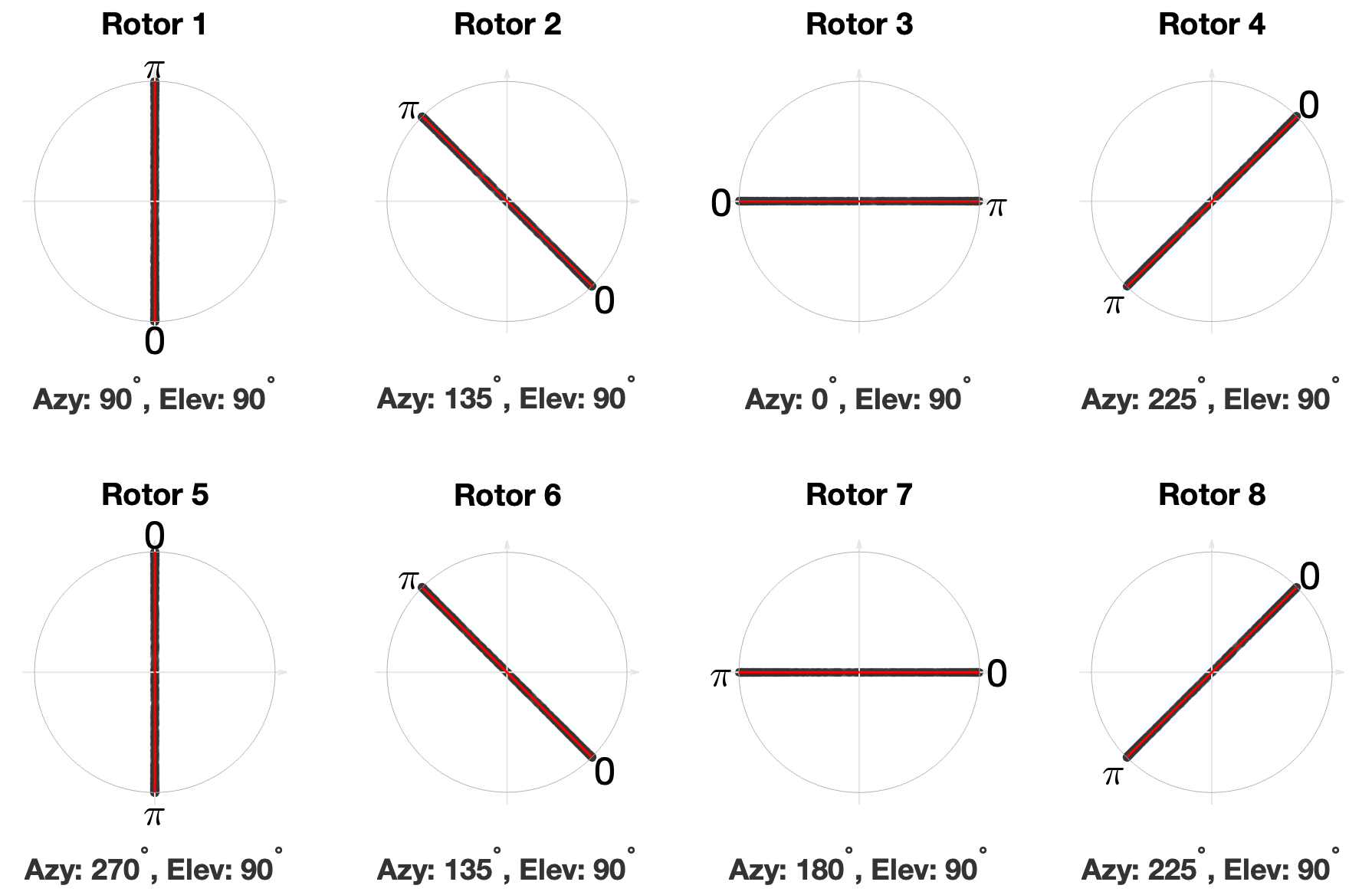}}
    \caption{Representation of the solution landscape data points (${M^*}$ points) in $\Manifold$ for the chassis: \textbf{Octagon ($N=8$)}. The gray dots represent the two coordinates of these points on each disc $\RPtwo_i$. The data is perfectly fitted by the red semi-ellipses, whose parameters (rotation and elevation are shown in each subplot). The combination of all the semi-elliptical curves matches exactly the $\tangentTorus$ of the \textbf{Octagon ($N=8$)}.}
    \label{fig:fit_poly8}
\end{figure}

\begin{figure}[t]
    \centering
    \framebox{\includegraphics[width=0.85\linewidth]{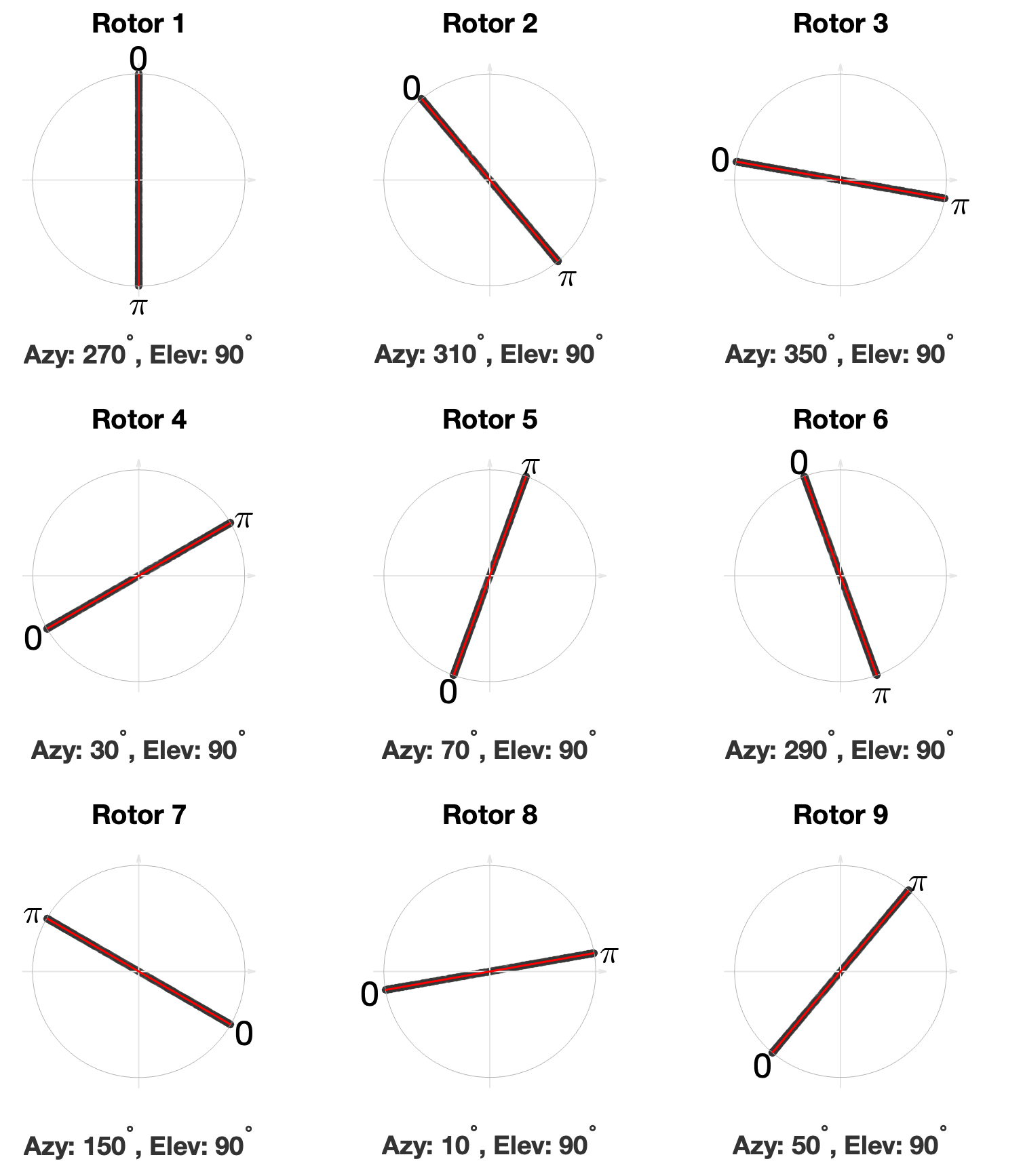}}
    \caption{Representation of the solution landscape data points (${M^*}$ points) in $\Manifold$ for the chassis: \textbf{Nonagon ($N=9$)}. The gray dots represent the two coordinates of these points on each disc $\RPtwo_i$. The data is perfectly fitted by the red semi-ellipses, whose parameters (rotation and elevation are shown in each subplot). The combination of all the semi-elliptical curves matches exactly the $\tangentTorus$ of the \textbf{Nonagon ($N=9$)}.}
    \label{fig:fit_poly9}
\end{figure}

\begin{figure}[t]
    \centering
    \framebox{\includegraphics[width=0.99\linewidth]{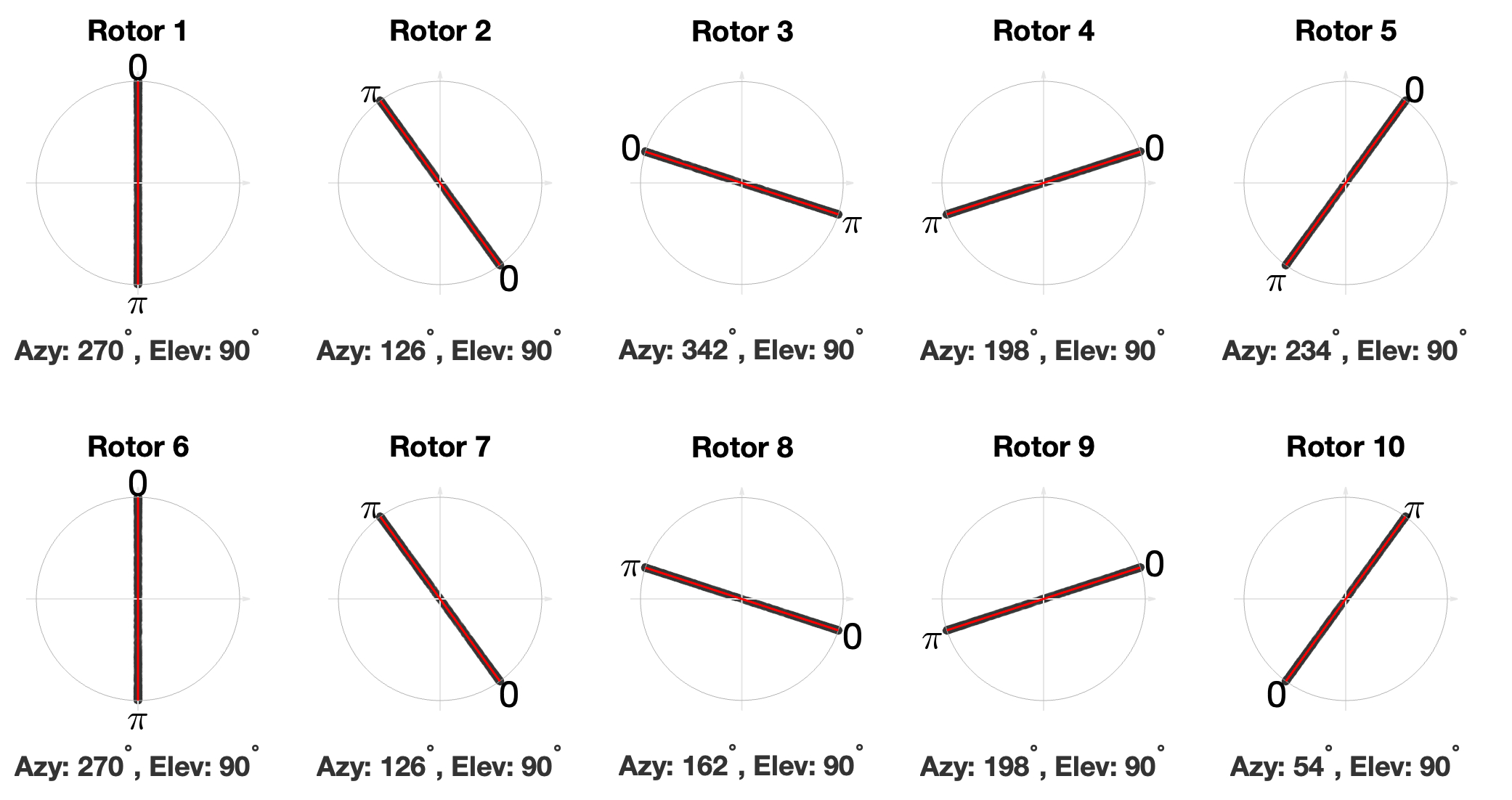}}
    \caption{Representation of the solution landscape data points (${M^*}$ points) in $\Manifold$ for the chassis: \textbf{Decagon ($N=10$)}. The gray dots represent the two coordinates of these points on each disc $\RPtwo_i$. The data is perfectly fitted by the red semi-ellipses, whose parameters (rotation and elevation are shown in each subplot). The combination of all the semi-elliptical curves matches exactly the $\tangentTorus$ of the \textbf{Decagon ($N=10$)}.}
    \label{fig:fit_poly10}
\end{figure}

\begin{figure*}[t]
    \centering
 
    \setlength{\tabcolsep}{3pt}
    \renewcommand{\arraystretch}{1.2}
    
    \newcolumntype{Y}{>{\centering\arraybackslash}X}

    \begin{tabularx}{\textwidth}{ |Y|Y|Y|Y|Y|Y|Y| }
        \hline

        \footnotesize \textbf{CRPol6} &
        \footnotesize \textbf{CRPol7} &
        \footnotesize \textbf{CRPol8} &
        \footnotesize \textbf{CRPol9} &
        \footnotesize \textbf{CRPol10} &
        \footnotesize \textbf{COct6} &
        \footnotesize \textbf{CCub8} \\
        
        \includegraphics[width=0.95\linewidth]{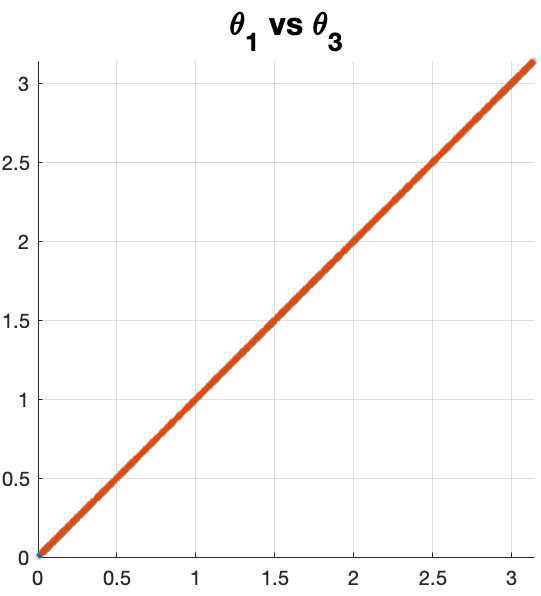} &
        \includegraphics[width=0.95\linewidth]{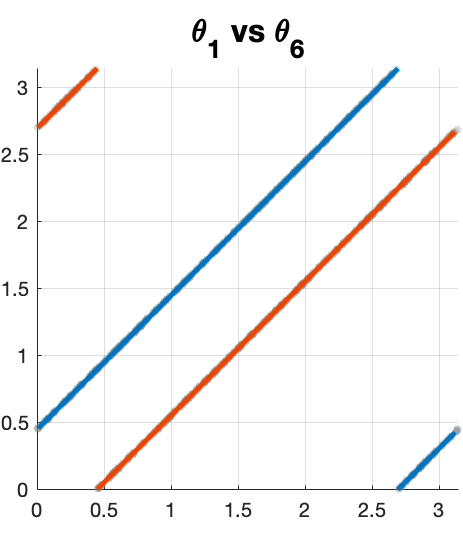} &
        \includegraphics[width=0.95\linewidth]{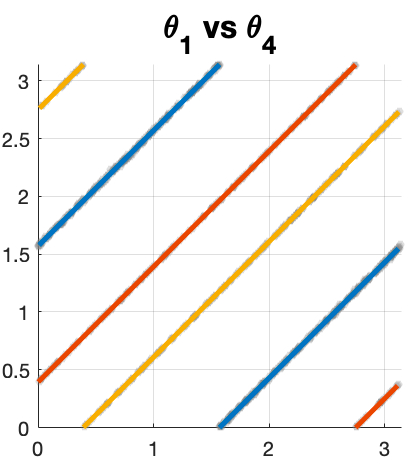} &
        \includegraphics[width=0.95\linewidth]{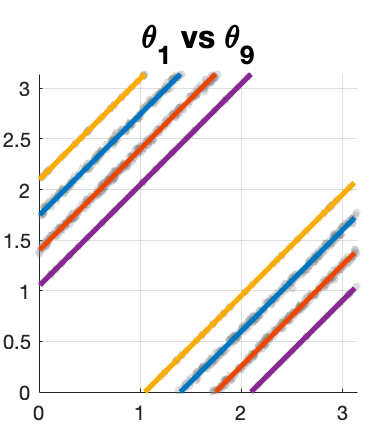} &
        \includegraphics[width=0.95\linewidth]{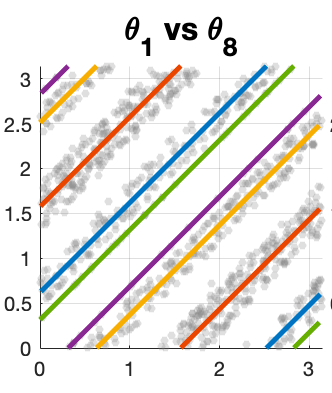} &
        \includegraphics[width=0.95\linewidth]{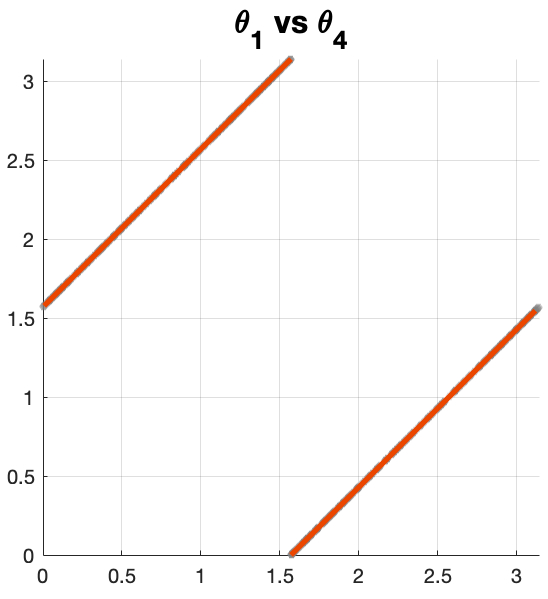} &
        \includegraphics[width=0.95\linewidth]{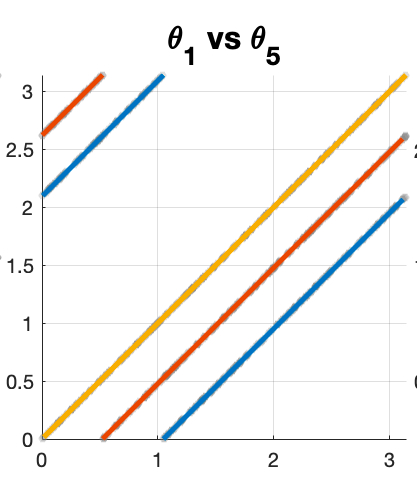} \\[-2pt] 

        \includegraphics[width=0.95\linewidth]{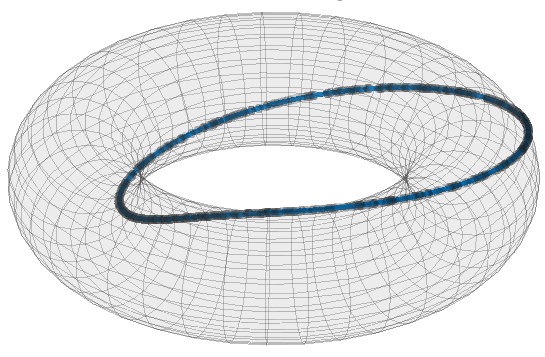} &
        \includegraphics[width=0.95\linewidth]{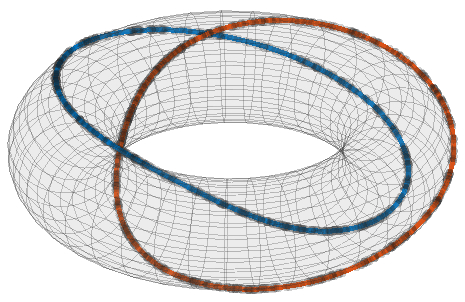} &
        \includegraphics[width=0.95\linewidth]{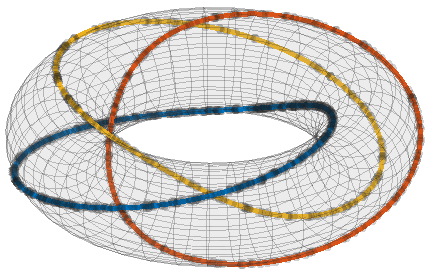} &
        \includegraphics[width=0.95\linewidth]{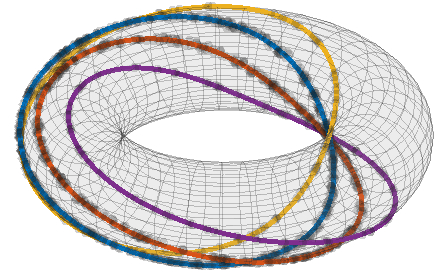} &
        \includegraphics[width=0.95\linewidth]{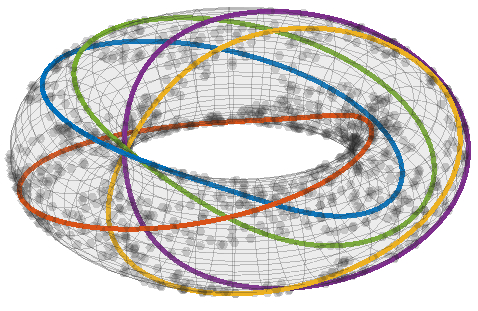} &
        \includegraphics[width=0.95\linewidth]{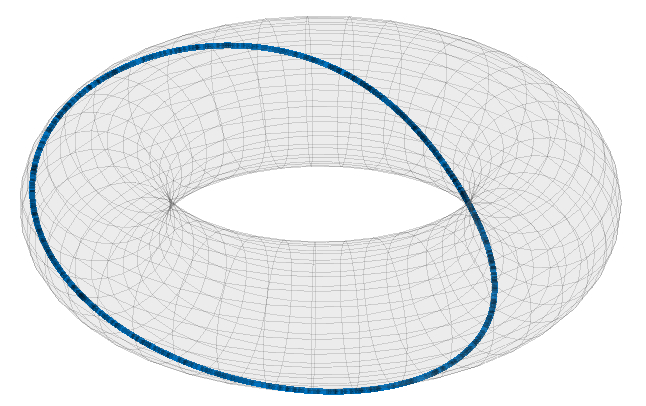} &
        \includegraphics[width=0.95\linewidth]{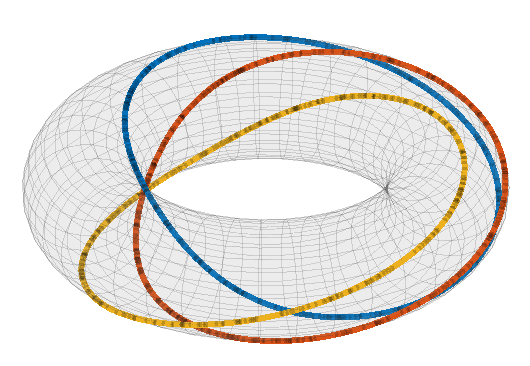} \\[2pt]

        \hline
    \end{tabularx}

    \caption{A partial summary fo the 1D-Manifold extraction results): \textbf{Top row:} Selection of fitted pairwise angular correlations. Each subplot shows a representative projection of the solution manifold for a specific regular chassis. The data points are overlaid with the fitted affine phase-locking curves, colored to distinguish the distinct $K$ topological branches (isomers) identified by the algorithm. \textbf{Bottom row:} Toroidal visualization. The correlation data is plotted on the 2-Torus embedded in $\mathbb{R}^3$. The distinct closed curves wrap around the torus without intersecting, providing a direct visual confirmation of the $K$ isomers.}
    \label{fig:correletion_fit_select}
\end{figure*}

\end{appendices}

\end{document}